\documentclass[trsc,nonblindrev]{informs3} 

\OneAndAHalfSpacedXI 

\usepackage{natbib}
 \bibpunct[, ]{(}{)}{,}{a}{}{,}%
 \def\bibfont{\small}%
\TheoremsNumberedThrough     

\EquationsNumberedThrough    

\usepackage{graphicx} 
\usepackage{caption} 

\usepackage{amsmath}
\usepackage{amsfonts}
\usepackage{amssymb}
\usepackage{verbatim}

\usepackage[dvipsnames,svgnames,x11names]{xcolor}
\usepackage[commandnameprefix=ifneeded, markup=underlined]{changes}

\usepackage{longtable}

\usepackage[linesnumbered,ruled]{algorithm2e}

\usepackage[export]{adjustbox} 
\usepackage{booktabs}
\usepackage{nicematrix}

\usepackage{subcaption}
\usepackage{multirow}

\usepackage[para,online,flushleft]{threeparttablex}

\usepackage{natbib}

\usepackage{hyperref}
\hypersetup{
	colorlinks,
	linkcolor={red!50!black},
	citecolor={blue!50!black},
	urlcolor={blue!80!black}
}

\usepackage{floatrow}
\newfloatcommand{capbtabbox}{table}[][\FBwidth]

\newcommand{\indep}{\perp \!\!\! \perp}
\renewcommand{\endproof}{\strut\hfill\Halmos\endTrivlist\addvspace{0pt}}

\begin{document}


\RUNAUTHOR{Liu et al.}

\RUNTITLE{Managing PUDOs with Causal Inference}

\TITLE{Estimating and Mitigating the Congestion Effect of Curbside Pick-ups and Drop-offs: A Causal Inference Approach}
\ARTICLEAUTHORS{%
\AUTHOR{Xiaohui Liu}
\AFF{Department of Information Systems and Analytics, School of Computing, National University of Singapore, Singapore\\
\EMAIL{xiaohuiliu@u.nus.edu}}

\AUTHOR{Sean Qian}
\AFF{Department of Civil and Environmental Engineering \& H. John Heinz III Heinz College, Carnegie Mellon University, Pittsburgh, PA, USA, 
\EMAIL{seanqian@cmu.edu}}

\AUTHOR{Hock-Hai Teo}
\AFF{Department of Information Systems and Analytics, School of Computing, National University of Singapore, Singapore \\
\EMAIL{teohh@comp.nus.edu.sg}}

\AUTHOR{Wei Ma}
\AFF{Department of Civil and Environmental Engineering, The Hong Kong Polytechnic University, Hong Kong SAR\\
\EMAIL{wei.w.ma@polyu.edu.hk}}

} 

\ABSTRACT{Curb space is one of the busiest areas in urban road networks. Especially in recent years, the rapid increase of ride-hailing trips and commercial deliveries has induced massive pick-ups/drop-offs (PUDOs), which occupy the limited curb space that was designed and built decades ago. These PUDOs could jam curbside utilization and disturb the mainline traffic flow, evidently leading to significant negative societal externalities. However, there is a lack of an analytical framework that rigorously quantifies and mitigates the congestion effect of PUDOs in the system view, particularly with little data support and involvement of confounding effects. To bridge this research gap, this paper develops a rigorous causal inference approach to estimate the congestion effect of PUDOs on general regional networks. A causal graph is set to represent the spatio-temporal relationship between PUDOs and traffic speed, and a double and separated machine learning (DSML) method is proposed to quantify how PUDOs affect traffic congestion. Additionally, a re-routing formulation is developed and solved to encourage passenger walking and traffic flow re-routing to achieve system optimization. Numerical experiments are conducted using real-world data in the Manhattan area. On average, 100 additional units of PUDOs in a region could reduce the traffic speed by 3.70 and 4.54 mph on weekdays and weekends, respectively. Re-routing trips with PUDOs on curb space could respectively reduce the system-wide total travel time by 2.44\% and 2.12\% in Midtown and Central Park on weekdays. Sensitivity analysis is also conducted to demonstrate the effectiveness and robustness of the proposed framework. 


}%


\KEYWORDS{Curbside management; Curbside pick-up and drop-off; Causal inference, Double and separated machine learning; Causal graph; Spatio-temporal data analytics; Machine learning}

\maketitle

%

\section{Introduction}


\subsection{Motivation}

\begin{figure}[h]
    \centering
    \includegraphics[width=\linewidth]{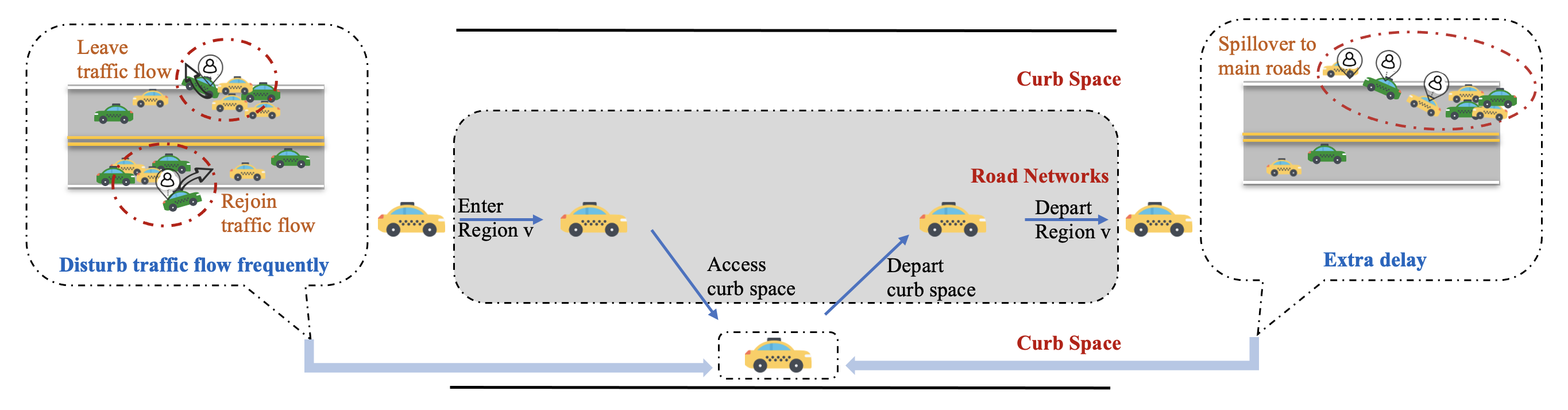}
    \caption{Illustration of congestion effect caused by PUDOs.}
    \label{fig:illustration of PUDO}
\end{figure}

In addition to roads and intersections, curb space is becoming a new conflicting area where multiple traffic flow converges and interacts \citep{ITE2018Curbside}. Curb space serves various traffic modes such as car parking, truck loading/unloading, scooters, and passenger pick-ups/drop-offs \citep{ITE2018Curbside,jaller2021fighting}. In recent years, substantial concerns about the congestion effect caused by curbside passenger pick-ups/drop-offs (PUDOs) have arisen \citep{jaller2021fighting, erhardt2019transportation, golias2001taxi}, and this study focuses on mitigating such concerns. The PUDO refers to the behavior that passengers get on and off the vehicles on curb space. Although the action of the curbside PUDO only takes about $1\sim2$ minutes~\citep{erhardt2019transportation, Ryland2019curb, jaller2021fighting, SFP2019}, it could induce traffic congestion by disturbing traffic flow and occupying curb space, as shown in Figure~\ref{fig:illustration of PUDO}. The reasons are two-fold: 1) PUDOs force vehicles to leave and rejoin the main traffic stream frequently, which disrupts vehicles on main roads~\citep{uwstudy, golias2001taxi, erhardt2019transportation, chai2020automated}; 2) PUDOs can be viewed as temporary parking on curb space \citep{schaller2011parking}. If the curb space is extensively filled with PUDOs \citep{butrina2020municipal}, vehicles will spill over to main roads and induce extra delay.

With the proliferation of ride-hailing services, massive orders generate excessive PUDOs on limited curb space, which further exacerbates the congestion effect caused by PUDOs. In 2019, Uber owns 111 million users and has completed 6.9 billion trip orders \citep{Iqbal2019uber}. Each order always starts with passengers' pick-ups and ends with drop-offs. Some studies~\citep{wijayaratna2015impacts, erhardt2019transportation} model the congestion effect of PUDOs as the average duration of each PUDO blocking or disturbing the traffic flow on curb space. In 2016, when the Transportation Network Companies (TNCs) started to provide services, the average duration of TNC PUDOs is 144.75 seconds on major arterial and 79.49 seconds on the minor arterial in San Francisco \citep{erhardt2019transportation}. However, the PUDO duration is around 60 seconds when TNCs do not exist in 2010~\citep{erhardt2019transportation, Ryland2019curb, jaller2021fighting, SFP2019}. The time delay caused by PUDO activities is ranked third among temporary loss of transportation capacity (TLC) events, right behind vehicular crashes and the presence of work zones~\citep{han2005estimating}. With the further development of Mobility as a Service (MaaS), it is foreseeable that the number of PUDOs will keep increasing in the near future \citep{smith2019data}. Therefore, it is challenging for public agencies to allocate the limited curb space to accommodate the massive PUDOs. In this paper, we focus on the PUDOs of ride-hailing vehicles as they account for the majority of all PUDOs in urban cities.

Although the omnipresent PUDOs play a significant role in traffic congestion~\citep{uwstudy}, to the best of our knowledge, there are few studies aiming to disentangle the congestion effect of curbside PUDOs. From the perspective of the ride-hailing service, both PUDOs and vehicle cruising are its by-products. However, PUDOs have not received as much attention as vehicular cruising \citep{xu2017optimal, xu2021equilibrium, zhang2022mitigating}. Many recent studies have pointed out that ride-hailing services contribute to traffic congestion by occupying more public resources than public transit does. The public resources include not just roads, but also curb space \citep{castiglione2016tncs, castiglione2018tncs, erhardt2019transportation, agarwal2023impact, tirachini2020ride, beojone2021inefficiency, zhang2023ride}. From the perspective of traffic operation and management, curbside parking has been extensively studied in recent years. For example, \citet{arnott2013curbside} consider parkers' heterogeneity in a high or low value of time and propose that adding curbside parking time limits can reduce the number of parkers with long-time parking and reduce wasteful cruising for parking. \citet{liu2021modeling} model the joint equilibrium of destination and parking choices given public curbside and private shared parking. By contrast, the PUDO, as a temporary form of parking, is less well-understood and needs further investigation. An evaluation study on the congestion effect of PUDOs shows that increasing one-unit PUDO could reduce the traffic speed by around $1\%$, and the estimation result is statistically significant~\citep{uwstudy}. The proposed estimation method is for one single region; approaches for estimating the network-wide congestion effects are still lacking.

The current practice to manage PUDOs relies on expert experience and heuristics~\citep{mccormack2019developing}. Ride-hailing PUDOs have not emerged as a major problem until 2012~\citep{zalewski2012regulating, butrina2020municipal}. Hence governments, TNCs, and researchers have not begun to turn their attention to the management of curb space to combat the chaos caused by PUDO \citep{smith2019data, zhang2022mitigating, castiglione2018tncs, uwstudy, schaller2011parking, Anurag2019curb, Ryland2019curb}. For example, airports like JFK have prepared a specific area for the PUDOs of ride-hailing vehicles (RVs), while some other airports ({\em e.g.}, LAX) have directly banned the curbside PUDOs by RVs. 
In general, various operation and management strategies can be adopted to mitigate the PUDOs' congestion effects, including traffic flow re-routing, curbside pricing \citep{liu2023optimal}, curb space allocation \citep{uwstudy}, and curb space re-planning \citep{mccormack2019developing}. \citet{jaller2021fighting} also propose to utilize curb space as a flex zone where multiple vehicles can occupy a different proportion of curb space at different time periods and locations. However, how to incorporate the precise estimation of the congestion effect of PUDOs into the management framework deserves further investigation. In this paper, we explore the possibility of using a traffic flow re-routing strategy to mitigate the overall congestion caused by PUDOs. The key idea is to shift the PUDOs from the areas with severe congestion effects to the areas with less congestion effects, thereby reducing the city-wide total travel time. 

In summary, this paper aims to estimate and mitigate the congestion effect of PUDOs, and the following two research questions will be addressed:
\begin{itemize}
    \item How do we estimate the congestion effect caused by PUDOs from observed traffic data?
    \item How do we manage PUDOs to minimize the city-wide total travel time based on the differences in heterogeneous congestion effects among regions?
\end{itemize} 

\subsection{Challenges and opportunities}
\label{subsec:Challenge}
This section introduces the challenges and difficulties in estimating the congestion effect of PUDOs. Only with an accurate estimation of the congestion effect can the corresponding management strategies be developed effectively using network modeling approaches. First, we define the number of PUDOs (NoPUDO) as the total number of pick-ups and drop-offs in a region within a fixed time interval. Without loss of generalization, this paper mainly focuses on estimating the average congestion effect of PUDOs, and the proposed framework can be utilized for PU and DO separately. Secondly, we use the average traffic speed in a region to represent its congestion levels. Specifically, lower traffic speed indicates a more severe congestion level. Therefore, the congestion effect of PUDOs can be quantitatively measured as the change of speed induced by the change of NoPUDO.

\begin{figure}[h]
    \centering
    \includegraphics[width=\linewidth]{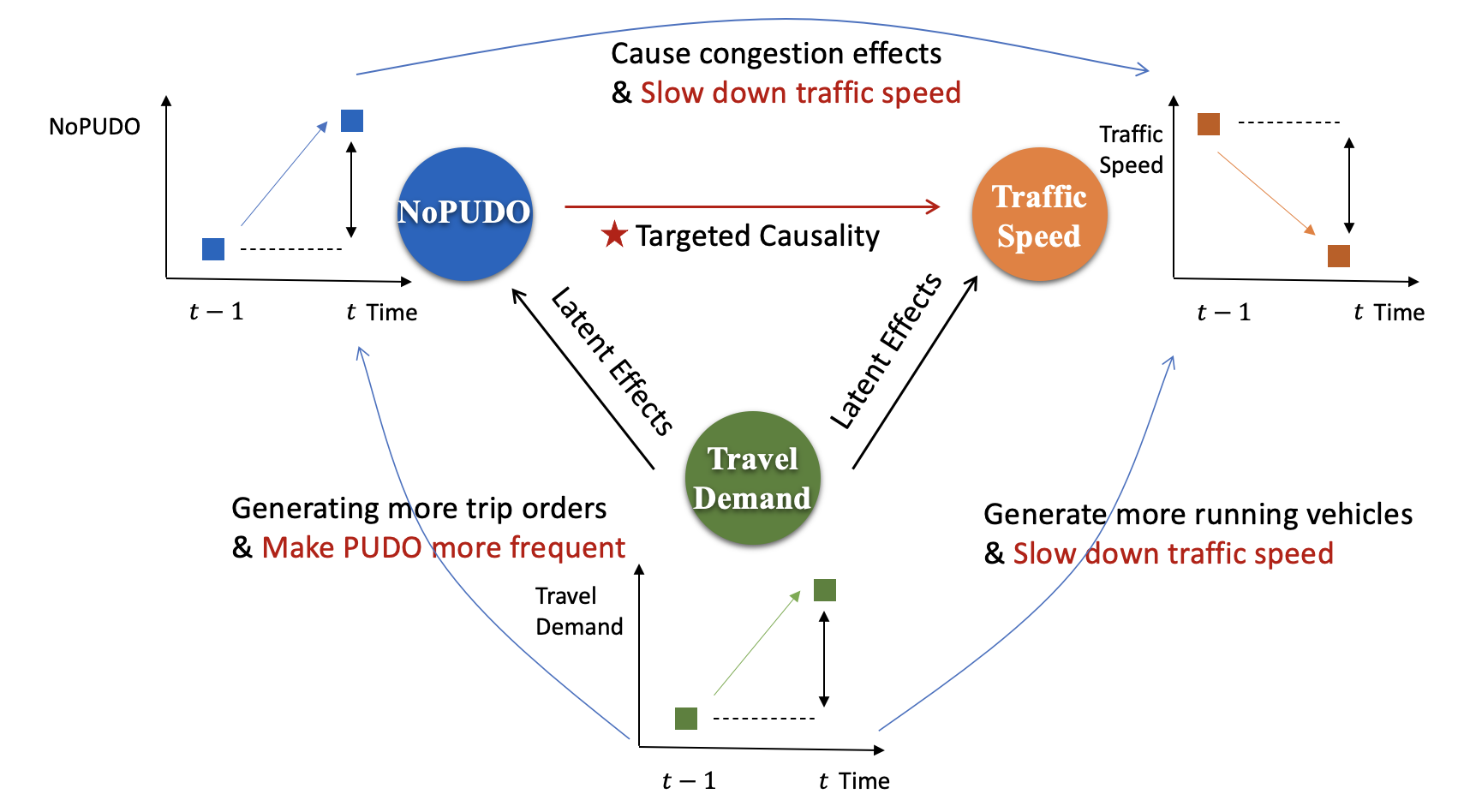}
    \caption{Relationship among travel demands, NoPUDO and traffic speed.}
    \label{fig:illustration of travel demand}
\end{figure}

However, it is challenging to capture and identify such congestion effects because both speed and NoPUDO are mutually affected by other latent factors, such as travel demands. An illustration of the relationship among travel demands, NoPUDO, and traffic speed is shown in Figure~\ref{fig:illustration of travel demand}. In general, the PUDO has a negative effect on traffic speed, which is our estimation target. However, the growing travel demands might spur more ride-hailing requests, making PUDOs happen more frequently. Simultaneously, the increasing travel demands also slow down traffic speed because more vehicles will occupy roads \citep{yuan2015capacity, retallack2019current}.

If the latent effect of traffic demand is overlooked and we directly estimate the relationship between NoPUDO and traffic speed, we are likely to overestimate the congestion effect. We use Example~\ref{ex:over} to illustrate how the overestimation arises.
 
\begin{example}
\label{ex:over}
For a specific time interval $t$, suppose 100 additional travelers arrive in a specific region, 20 of them take RVs and the rest 80 travelers drive by themselves. The 20 travelers will be dropped off on curb space, while the 80 travelers directly park in the garage. Due to the increasing traffic demand, the average speed in the region decreases by 2 miles/hour (mph). The sources inducing speed reduction are actually two-fold: 1) the congestion induced by the 100 vehicles (both RVs and conventional cars) on the roads; 2) the congestion effect induced by the 20 PUDOs. For the speed reduction, we suppose that the former accounts for 1.5 mph, and the latter accounts for 0.5 mph. Then the congestion effect of a PUDO can be calculated as $0.5 / 20 = 0.025$ mph/PUDO. However, if we directly evaluate the relationship between NoPUDO and traffic speed without identifying the causal relationship, the congestion effect of a PUDO will be {\em  wrongly} calculated as $2 / 20 = 0.1$ mph/PUDO, and hence the congestion effect is over-estimated.

\end{example}

Essentially, what Example~\ref{ex:over} demonstrates is the difference between correlation and causality between NoPUDO and traffic speed. The problem of estimating the congestion effect of PUDOs is indeed a causal effect estimation problem. Theoretically, it can be formulated as the problem of quantifying how the changes of NoPUDO affect the changes in traffic speed after eliminating other latent confounders~\citep{greenwood2017show, burtch2018can, babar2020examining}. One intuitive solution to measure the congestion effect is to conduct field experiments in the real world, but this is practically demanding and costly. In recent years, the emergence of advanced machine learning (ML) models has empowered us to infer the causal effect from observational data \citep{pearl2019seven}. 

Causal inference consists of two major tasks: 1) Causal discovery; and 2) Causal estimation. For a comprehensive review of using ML models for causal inference, readers can refer to \citet{yao2021survey}. This paper focuses on estimating the causal effects, and we assume that the causal relationship has been identified. Some representative models for causality estimation include inverse-propensity scoring (IPS) methods, meta-learners, deep representation-based methods, and double machine learning (DML). The IPS methods require estimating the probability of occurrence of each data point, which can be challenging in traffic applications. The meta-learner methods including T-learner, S-learner, X-learner, etc, aim at addressing the estimation bias caused by an unbalanced group size between treatment and control groups and they are more suitable for scenarios with binary treatments~\citep{kunzel2019metalearners}. The deep representation-based methods lack theoretical guarantees, making them less reliable for engineering applications \citep{yao2019estimation}. The closest work to this paper is DML, which can estimate the causal effects by training two ML models~\citep{wager2018estimation,oprescu2019orthogonal}. This method has rigorous theoretical guarantees on the estimation quality, and hence it is suitable for engineering applications \citep{chernozhukov2017double, chernozhukov2018double}. 

In recent years, the DML method has been extensively utilized to estimate treatment effects in diverse contexts, such as marketing promotion, social media platforms, labor markets, and so on. Some previous literature demonstrates and justifies the effectiveness of the DML method from a statistical perspective. One study investigates how effectively non-experimental approaches, DML and stratified propensity score (SPS), can recover the causal effect estimation from randomized controlled trials (RCT) in advertising measurement on the Facebook platform, and concludes that DML can perform better than the traditional econometric method SPS~\citep{gordon2023close}. When it is challenging to explicitly model the distribution of the covariate-conditional average treatment effect function, the DML method can enable exploring and ascertaining the precise upper and lower bounds of the conditional value at risk (CVaR) of the individual treatment effect (ITE) distribution \citep{kallus2023treatment}. Existing pioneering studies empirically validate the feasibility and efficiency of applying the DML method to estimate causality in real-world applications. \citet{ellickson2023estimating} employ the DML method to estimate the distinct effects of email components, subject lines and content, on consumers' decisions in email opening and goods purchasing in the digital marketing promotion. \cite{xu2023mobile} adopt the DML method to address the endogeneity issues in estimating the effect of mobile payment adoption on bank customers' credit card activities and their loyalty. When studying the causal effect of users' reputation on persuasion in a deliberation online platform, \cite{manzoor2023influence} utilize the DML method to control the unstructured confounders. \cite{covert2023relinquishing} apply the DML method to non-parametrically control confounders including location and time, and quantify how auctions and informal negotiations affect upfront payment and output of oil and gas leasing differently in Texas. \cite{dube2020monopsony} separate potential exogenous variation by including a large number of covariates in the DML method and investigate the monopsony power of one of the largest on-demand labor online platforms, Amazon Mechanical Turk.

Despite the recent trends of applying DML for causal analysis, it is challenging for the standard DML to model the interwoven causal relationship between NoPUDO and traffic speed, especially when such a relationship is convoluted with both time and space. A novel method needs to be developed to consider the spatio-temporal patterns of both NoPUDO and traffic speed when estimating the network-wide congestion effects of PUDOs.

\subsection{Contributions}

Overall, the literature lacks a quantitative method to estimate the congestion effect of PUDOs on traffic speed using the observed traffic data, and how the estimated congestion effect can be utilized for traffic management has not been examined. To bridge the research gaps, this paper proposes a data-driven framework to evaluate and manage the spatio-temporal congestion effects of PUDOs using multi-source traffic data. This paper first rigorously analyzes the causal relationship between NoPUDO and traffic speed. Next, we develop the Double and Separated Machine Learning (DSML) method to estimate the congestion effect of PUDO. A re-routing strategy is further formulated and solved by re-distributing the PUDO from busy regions to less busy regions, thereby mitigating the overall traffic congestion. Lastly, the proposed framework is examined with real-world data in the large-scale network in New York City to demonstrate its effectiveness.

The contributions of this paper can be summarized as follows:
\begin{itemize}
    \item To the best of our knowledge, this is the first study to use the causal inference approach to estimate the congestion effect of PUDOs from a data-driven perspective.
    
    \item This study rigorously formulates a causal graph to articulate the spatio-temporal relationship between the NoPUDO and traffic speed. A novel DSML method is developed and theoretically analyzed for estimating the congestion effect of PUDOs based on the causal graph.
    
    \item To demonstrate the usefulness of the estimated congestion effects of PUDOs, we develop a re-routing formulation to re-distribute PUDOs to minimize the network-wide total travel time, and a customized solution algorithm is developed to effectively solve the formulation.
    
    \item The effectiveness of the DSML method and re-routing strategy are validated with real-world data in a large-scale network in the Manhattan area. The estimation results obtained by the DSML method are consistent with the actual traffic conditions from both the time and space perspectives. Besides, the re-routing strategy effectively reduces the network-wide total travel time in Manhattan, eventually verifying the potential usefulness of the estimated congestion effect of PUDOs.
 
\end{itemize}

The remainder of this paper is organized as follows. Section~\ref{sec:methodology} discusses the causal estimation framework, which consists of the causal graph, DSML, and the re-routing formulation. Section~\ref{sec:solution algorithm} presents the solution algorithms, and Section~\ref{sec:numberical experiment} exhibits the numerical experiments on the Manhattan area. Finally, conclusions are drawn and discussed in Section~\ref{sec:conclusion}.

\section{Model}
\label{sec:methodology}

In this section, we first develop a causal graph to model the spatio-temporal causality between NoPUDO and traffic speed and mathematically formulate the structural equation models. Secondly, the DSML method is developed and analyzed for the causal graph. Thirdly, a system-optimal problem is mathematically formulated and solved to minimize the total travel time by re-routing PUDOs across regions. Notations used in this paper are summarized in Appendix~\ref{sec:notations} Table~\ref{tb:dsmldf}, and each notation will also be introduced when it first appears.

\subsection{Causal relationship between NoPUDO and Traffic Speed}
A causal model consists of a causal graph and the corresponding statistical model~\citep{spirtes2010introduction}. In this section, we first develop and analyze the causal graph that describes the causal relationship between the NoPUDO and traffic speed. Next, we adopt a structural equation model to mathematically formulate the above causal relationship by specifying the functions of involved variables.

\subsubsection{Structural causal modeling}
\label{sec:causal}

We translate the transportation domain knowledge about the causal relationship between the NoPUDO and traffic speed into the causal graph. Causal graphs have been extensively utilized to infer and estimate treatment effects in numerous studies. The causal graph can articulate the complicated causal relationship among involved variables in an intuitive and concise way based on the prior domain knowledge~\citep{pearl2000models, spirtes2010introduction, pearl2022external, morgan2015counterfactuals}.~\cite{tafti2020beyond, park2023transporting} argue that causality estimation in a new context can be derived from analogous causal scenarios based on observable causal diagrams.~\cite{manzoor2023influence} adopt the causal graph to control confounding variables in estimating the effect of reputation on persuasion on the deliberation online platform. Leveraging the causal graph lays a robust theoretical foundation, enabling us to employ data-driven approaches to conduct causality estimation.

To present the developed causal graph, we first discuss the involved notations. Traffic states of a city are modeled by a spatial random variable that evolves over each time interval, $\{y_v^t \in \mathbb{R}^+, t \in \mathbb{T}\}$, where $v$ is a region in $\mathcal{G}$, $v\in \mathcal{V}$, and $\mathcal{G}$ denotes the multi-community city consisting of the set of regions $\mathcal{V}$~\citep{liu2021modeling}. $\mathbb{T}$ is the set of time intervals, and $t$ represents a specific time interval. We use time intervals instead of time points as the unit of analysis due to the limitation of data granularity in our study. Specifically, the data granularity of NoPUDO can be traced down to the level of seconds, while that of traffic speed is only every 5 minutes. Therefore, the entire analysis is restricted by the coarser granularity, {\em i.e.}, 5 minutes. $y_v^t$ is the quantitative measures of traffic states ({\em e.g.}, speed or flow) in the region $v$ at the time interval $t$~\citep{he2016traffic}. Besides, we use $d_v^t$ to denote the NoPUDO in the region $v$ at the time interval $t$. Without loss of generality, this paper models the average effect of the total NoPUDO on traffic speed in a given region $v$. We can further extend the proposed framework to consider other traffic states, such as flow, density, travel time, etc. 

As discussed in Section~\ref{subsec:Challenge}, the relationship between NoPUDO and traffic speed is convoluted with latent factors such as travel demands. In addition, the estimation of the congestion effect should also consider the temporal and spatial features of traffic scenes. Given a time interval $t$ and a region $v$, we elaborate the interwoven causal relationship between the NoPUDO and traffic states as follows:

\begin{itemize}
    \item Firstly, the traffic speed $y_v^t$ is affected by its historical traffic speed records $\mathbf{Y}_v^{t-I:t-1}$ from time interval $t-I$ to $t-1$. Because the traffic speed changes gradually throughout the day, the historical traffic speed $\mathbf{Y}_v^{t-I:t-1}$ can reflect the congestion levels, and passengers may refer to the past speed records to avoid picking up or dropping off in the congested regions. Hence $\mathbf{Y}_v^{t-I:t-1}$ is a critical factor for predicting the traffic speed.

    \item Secondly, the traffic speed $y_v^t$ in the region $v$ is also affected by the traffic speed of its surrounding regions during the past time window $\mathbf{Y}_{\mathcal{N}(v)}^{t-I:t-1}$, due to traffic flow exchanges. For example, if the neighboring regions $\mathcal{N}(v)$ of the region $v$ are congested by traffic accidents, the accumulated vehicles will spillover to the region $v$. The consideration of the surrounding traffic state actually manifests the importance of spatial correlation in causality estimation.

    \item Thirdly, the NoPUDO $d_v^t$ is affected by its historical NoPUDO $\mathbf{D}_v^{t-I:t-1}$ in region $v$ from the time interval $t-I$ to the time interval $t-1$. Similar to the traffic speed prediction, the historical NoPUDO $\mathbf{D}_v^{t-I:t-1}$ reflects the demand levels, and hence it is critical for predicting $d_v^t$. 

    \item Fourthly, external control variables $\mathbf{W}_{v}^t$, such as weather, holidays, peak hours, and so on, also affect the traffic speed and NoPUDO. For instance, rain and snow may limit drivers' sight, therefore making traffic speed slower and travel time longer~\citep{ahmed2018impacts}. Besides, holidays may spur more trip orders around places of interest than usual~\citep{rong2017taxi}, which accumulates more NoPUDO. Therefore, these external control variables should be included to eliminate potential biases in causality estimation.
\end{itemize}

Additionally, we assume Assumption~\ref{ap:uncorrelated y and d_history} holds as the congestion effect of PUDOs is immediate and the effect duration is short. 

\begin{assumption}
\label{ap:uncorrelated y and d_history}
For region $v$ in the network $\mathcal{G}$, the average traffic speed $y_v^t$ in the time interval $t$ is not causally affected by the historical records of the NoPUDO $\mathbf{D}_v^{t-I:t-1}$.
\end{assumption}

In short, the continuity of time, interactivity in space, and extra influence caused by external variables make the causality estimation between NoPUDO and traffic speed more dynamic and intricate. To prepare for the modeling of the causal graph between NoPUDO and traffic speed, we will make rigorous assumptions and present the causal graph in the following section.



\subsubsection{Structural equation models}
\label{sec:structural equation models}

In this section, we rigorously formulate the congestion effect caused by PUDOs in a structural equation model. We define $\theta_v$ as the changes in traffic speed $y_v$ caused by a one-unit change of NoPUDO in the region $v$. In this study, we use traffic speed to represent the traffic conditions and a smaller speed indicates that the current traffic is more likely to become congested. Mathematically, $\theta_v$ is defined based on Assumption~\ref{ap:01}.

\begin{assumption}[Linear effects]
\label{ap:01}
For a specific region $v$, given fixed $\mathbf{Y}_v^{t-I: t-1}, \mathbf{Y}_{\mathcal{N}(v)}^{t-I:t-1}, \mathbf{W}_{v}^t$, the congestion effect $\theta_v$ is defined in Equation~\eqref{eq:theta}.

\begin{equation}
\label{eq:theta}
y_v^t |_{\texttt{do}(d_v^t = d_1)} - y_v^t |_{\texttt{do}(d_v^t = d_2)} = \theta_v (d_1 - d_2) 
\end{equation}

where $\texttt{do}(\cdot)$ is the do-operation defined in \citet{pearl2009causality}, and $d_1$ and $d_2$ are two arbitrarily positive integers representing the NoPUDO.
\end{assumption}

One can read from Equation~\eqref{eq:theta} that given a region $v$, the congestion effect of PUDOs on traffic speed is linear. The linear relationship refers to that adding one additional unit of PUDO will make traffic speed decrease by $|\theta_v|$ in the region $v$. Additionally, we expect that $\theta_v \leq 0$ because the increase of the NoPUDO could induce more congestion. 

Generally, different regions in a city have different population densities, economic statuses, and traffic conditions. These factors will all contribute to the fluctuation of the estimated congestion effect caused by PUDOs in different regions. We assume the homogeneity within each region, and it means that the congestion effect caused by PUDOs ({\em i.e.}, $\theta_v$) is constant within a region. Therefore, we conduct the causal analysis based on the regional level.

At the same time, it is noted that the linear assumption of the congestion effect is strong for the actual complex and ever-changing traffic scenarios. To generalize the DSML method, we relax the assumption of linear effects into the non-linear effects and estimate the congestion effect based on different conditions. The extended estimation framework presented in this paper to estimate the conditional average treatment effect (CATE)~\citep{abrevaya2015estimating} is shown in Appendix~\ref{subappendix:CATE}. Given different traffic conditions, the estimated value of $\theta_v^t$ changes with respect to the dynamic traffic over time and region. The time and space-varying change of $\theta_v^t$ indicates the non-linear effects between the NoPUDO and traffic speed under different traffic conditions.

To the best of our knowledge, this is the first study to formulate and estimate the congestion effect of NoPUDO on the curb space from the causal inference perspective. It is indispensable for us to start the congestion effect estimation from the average treatment effect (ATE) as shown in the linear assumption (Assumption~\ref{ap:01}). Therefore, we mainly present the following causal analysis based on the Assumption~\ref{ap:01}. 

To better understand Equation~\eqref{eq:theta}, we note that the following three remarks hold for $\theta_v$ as a result of Assumption~\ref{ap:01}. Remark \ref{re:time} focuses on the homogeneous congestion effect across time, Remark \ref{re:hetero} emphasizes the heterogeneous congestion effect across different regions, and Remark~\ref{re:ind} highlights the independence of the congestion effect for each region.

\begin{remark}[Constant effects within a region]
\label{re:time}
The congestion effect is constant within each region and across different time intervals. In other words, for each region $v$, $\theta_v$ does not depend on the time intervals in which the PUDO occurs.
\end{remark}

Given a region $v$, Remark~\ref{re:time} simplifies the research question regarding dynamic congestion effect estimation to a static estimation problem by not considering its time variation. In this paper, we mainly focus on estimating and discussing the $\theta_v$ for region $v$ on weekdays and weekends respectively.

Besides, to enhance the generalization of our method, we also propose that we can estimate the time-varying and dynamic congestion effect based on the static estimation principle of Remark~\ref{re:time}. Specifically, given region $v$, we can first separate the whole observed traffic records into small pieces according to different time intervals. Secondly, we can run the proposed framework based on the above-split datasets in each time interval separately to obtain the time-varying congestion effect estimation. We can also conduct the congestion effect estimation given different traffic conditions.

\begin{remark}[Heterogeneous effects across different regions]
\label{re:hetero}
The congestion effect $\theta_v$ is a constant property of region $v$, and it can be different across different regions. 

\end{remark}

The congestion effects of NoPUDO are dependent on different built environments of different regions, and hence $\theta_v$ is heterogeneous across different regions. Specifically, different traffic and external conditions, such as different numbers of roads, economic development, and so on in a region will also affect the congestion effect of NoPUDO.
If we only run an aggregated estimation model based on the entire trip records data, the estimated congestion effects would be inaccurate and biased because it does not control the region-fixed effect. Instead, if we run separate estimation models for each region only based on the trip records of the current region, it will implicitly control region-based variables of different regions, and mitigate the interference of other variables.


\begin{remark}[Independent effects across different regions]
\label{re:ind}
The congestion of region $v$ is not related to the congestion effects of other regions $\theta_{v'}, \forall v' \neq v$.
\end{remark}

In other words, the numerical value of the congestion levels in the region $v$ will not be directly affected by the congestion effect of other regions $\theta_{v'}$. Specifically, we can include historical records of neighboring traffic speed as features in predicting the congestion level $y_v^t$ in region $v$, while $\theta_{v'}$, the congestion effect in other regions, should not be included to affect traffic speed directly.

Combining Remark \ref{re:time}, \ref{re:hetero}, and \ref{re:ind}, one can see that Assumption \ref{ap:01} ensures that the estimation of $\theta_v$ can be conducted for each region $v$ separately. The region-based estimation is also effective in controlling confounding variables when we estimate the congestion effect within each region. 

\begin{figure}[h]
    \centering
    \includegraphics[width=\linewidth]{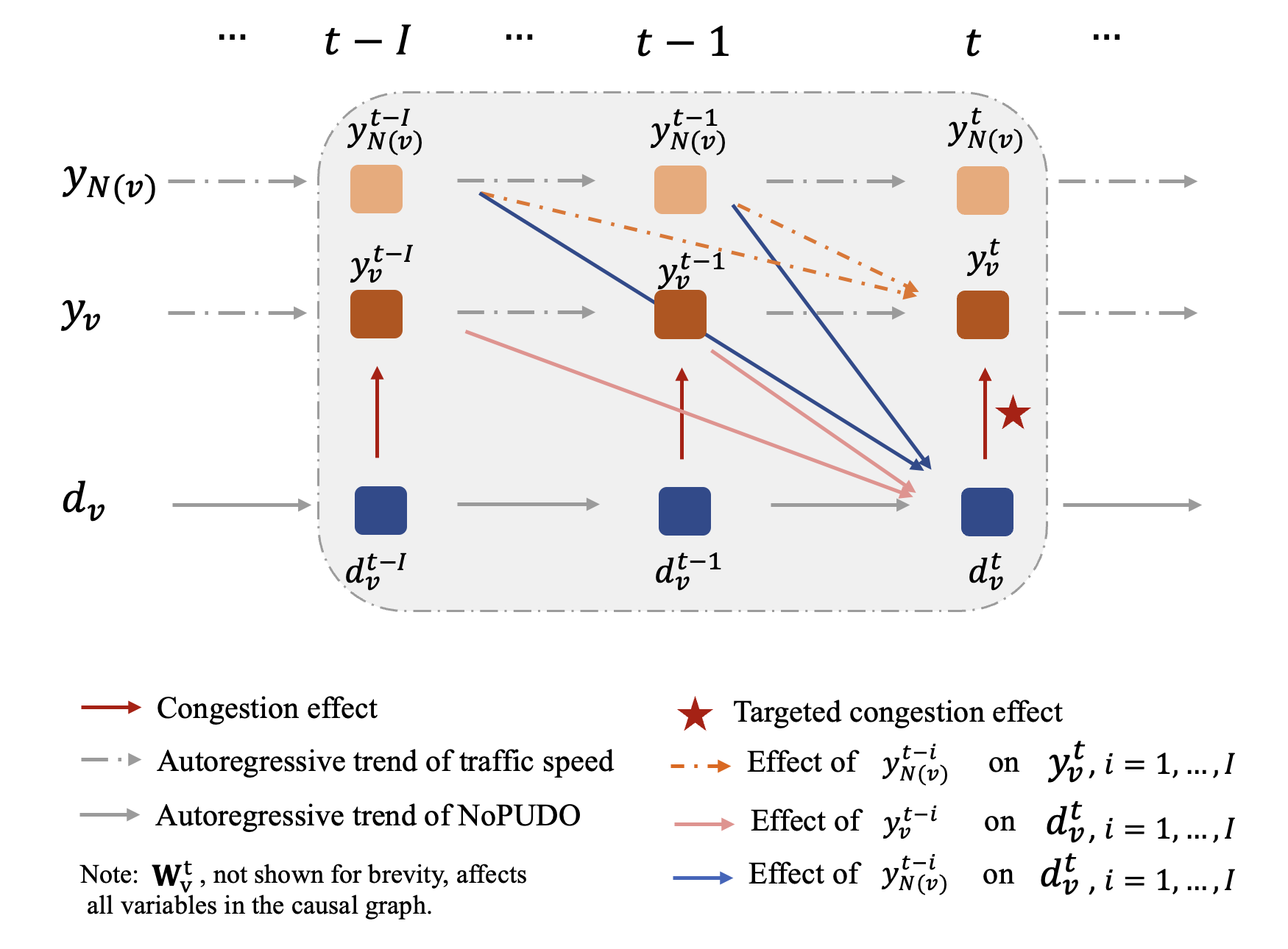}
    \caption{The causal graph of the NoPUDO and traffic speed.}
    \label{fig:dsml_casual_graph}
\end{figure}

Combining Assumption~\ref{ap:uncorrelated y and d_history} and Assumption~\ref{ap:01}, we develop the causal graph, as shown in Figure~\ref{fig:dsml_casual_graph}, to depict the causal relationship of PUDOs and traffic speed in both time and space dimensions. It is worth noting that, for region $v$, the NoPUDO in $\mathcal{N}(v)$ does not causally affect $d_v^t, y_v^t, \forall t$. We believe conditioning on $\mathbf{Y}_v^{t-I:t-1}$, $\mathbf{Y}_{\mathcal{N}(v)}^{t-I:t-1}$, $\mathbf{D}_v^{t-I:t-1}$, and $\mathbf{W}_{v}^t$, $\mathbf{D}_{\mathcal{N}(v)}^{t-I:t-1}$ is independent of $d_v^t$ and $y_v^t$.

The proposed causal graph contains two random variables $y_v^t$ and $d_v^t$, as we omit $\mathbf{W}_v^t$ for simplicity. 
To show the causal relationship more clearly in Figure~\ref{fig:dsml_casual_graph}, 
we have the red solid line with a red star from $d_v^{t}$ to $y_v^{t}$ indicating the causal effect of PUDOs on traffic speed, which is the estimation target. Specifically, the effect $\theta_v$ is represented by the change of current speed $y_v^t$ induced by increasing one additional unit of PUDO in the region $v$, given other variables are unchanged. 

First, we use the grey dotted line from $y_v^{t-1}$ to $y_v^t$ to describe the autoregressive trend of the traffic speed from the past time windows $t-I:t-1$ to the time interval $t$. The orange dotted line from $y_{\mathcal{N}(v)}^{t-1}$ to $y_v^t$ represents the effect of the speed in surrounding regions $\mathcal{N}(v)$ on the current region $v$. The reason for both two dotted lines here is that traffic state $y_v^{t}$ is affected by both historical traffic speed from the time interval $t-I$ to $t-1$ in the region $v$ ($\mathbf{Y}_v^{t-I:t-1}$) and traffic speed from the time interval $t-I$ to $t-1$ in neighboring regions $\mathcal{N}(v)$ ($\mathbf{Y}_{\mathcal{N}(v)}^{t-I:t-1}$). 
Secondly, we use the grey solid line from $d_v^{t-1}$ to $d_v^t$ to indicate the autoregressive trend of the NoPUDO from the past time windows $t-I: t-1$ to the time interval $t$. Because the NoPUDO in the time interval $t$ and region $v$ ($d_v^{t}$) is affected by its historical trends from the time interval $t-I$ to the time interval $t-1$ in the region $v$, denoted by $\mathbf{D}_v^{t-I:t-1}$. Additionally, the pink and blue solid lines represent the effects of historical traffic speed ($\mathbf{Y}_{v}^{t-I:t-1}, \mathbf{Y}_{\mathcal{N}(v)}^{t-I:t-1}$) on the NoPUDO. 

We claim that the historical NoPUDO $\mathbf{D}_v^{t-I:t-1}$ will not affect the traffic speed $y_v^{t}$ at time interval $t$. This study conducted the congestion effect estimation based on 5-minute intervals. The time gap between the $d_v^{t-1}$ and $y_v^{t}$ is 5 minutes. Because the traffic states, especially the travel demands, change rapidly in every 5 minutes~\citep{ma2018estimating}. Therefore, the change in the NoPUDO at time interval $t-1$ will affect the speed at the time interval $t-1$ but will not affect the speed at the time interval $t$.

Given the causal graph and Assumption~\ref{ap:01}, we are now ready to formulate the causal relationship between NoPUDO and traffic speed in Equation~\eqref{eq:y} and Equation~\eqref{eq:d}.

\begin{eqnarray}
\label{eq:model}
{y}_{v}^t &=& \varphi_v(\mathbf{Y}_v^{t-I: t-1};\mathbf{Y}_{\mathcal{N}(v)}^{t-I:t-1}; \mathbf{W}_{v}^t) + \theta_v \cdot d_v^t + e_{v}^{t} \label{eq:y}\\
{d}_{v}^t  &=&  \psi_v \left( \mathbf{D}_v^{t-I: t-1}, \mathbf{Y}_v^{t-I: t-1}, \mathbf{Y}_{\mathcal{N}(v)}^{t-I:t-1}; \mathbf{W}_{v}^t \right)+ \xi_{v}^t \label{eq:d}
\end{eqnarray}
where $\varphi_v$ predicts the traffic speed $y_v^t$ using historical traffic speed records, and $\psi_v$ predicts the NoPUDO $d_v^t$ using historical traffic speed as well as the historical NoPUDO. Both $e_{v}^{t}$ and $\xi_{v}^t$ are zero-mean noise, which are defined in Equation~\eqref{eq:noise1} and~\eqref{eq:noise2}.

Equation~\eqref{eq:y} and \eqref{eq:d} can be viewed as a Structural Equation Model (SEM): traffic speed $y_v^{t}$ is the outcome variable, the NoPUDO $d_v^t$ is the treatment variable, and $\mathbf{D}_v^{t-I: t-1}, \mathbf{Y}_v^{t-I: t-1}, \mathbf{Y}_{\mathcal{N}(v)}^{t-I:t-1}, \mathbf{W}_{v}^t$ are control variables. Specifically, $\theta_v$ is the treatment effect that shows the effect of the NoPUDO $d_v^t$ on traffic speed $y_v^t$. The inclusion of control variables can help to eliminate the biased influence of other factors on the estimation results. 

It is worth noting that, the congestion effect $\theta_v$ is estimated by the traffic speed and NoPUDO at the same time interval of 5 minutes $t$ in Equation~\eqref{eq:y} for two main reasons. First, the finest granularity of the available traffic speed dataset is 5 minutes. Although the NoPUDO dataset is measured at the level of seconds, they are aggregated at the same 5-minute interval in order for the DSML method to run~\citep{xu2023mobile}. The details of the datasets are introduced in Section \ref{data_description}. Second, empirical evidence shows that each PUDO lasts for about 79 - 145 seconds, which is much shorter than the 5-minute time interval. Hence, restricted by the coarser granularity of the speed dataset and the short-lived nature of the congestion effect, we chose 5 minutes as the finest available unit of the time interval~\citep{ma2018estimating, erhardt2019transportation}. We also adopt a distributed lag model based on the datasets to quantify the correlation between traffic speed and NoPUDO over time. Our empirical result shows 1) a general negative temporal trend between traffic speed and NoPUDO; and 2) the most negative correlation occurs at the same time interval $t$ compared to that at other time intervals $t-1$, $t-2$... and $t-I$ (see Appendix~\ref{sec:distributed_lag_model}). Therefore, it is reasonable to consider that the congestion effects happen within the same time interval along with the NoPUDO and traffic speed given the time granularity of the current datasets. In other words, we consider a region $v$ as one subject, where the aggregated numbers of PUDO within the time interval $t$ can be viewed as the observational value of the treatment for the region, and the average traffic speed within the same time interval $t$ can be regarded as an observational value of the dependent variable~\citep{zhang2023ridesharing, tan2021good}. If the data granularity allows, this structural equation model can also be extended to use different time indices between the dependent variable and independent variables for more detailed analysis. For instance, if we have both NoPUDO and traffic speed data in the second level, we can make a more precise estimation by using the traffic speed at the time point $t$ and NoPUDO at the time point $t-1$, and $t$ represents each second.

The above formulation, Equation~\eqref{eq:y} and Equation~\eqref{eq:d}, takes the spatial autocorrelation across regions into account. Although we estimate the congestion effect of the NoPUDO in each region by a separate model, we have considered the spatial interaction effect by adding variables representing the spatial correlation. Specifically, we add the average speed of neighboring regions $\mathbf{Y}_{\mathcal{N}(v)}^{t-I:t-1}$ as one of the factors affecting the speed ${y}_{v}^t$ and $d_v^t$. Additionally, it is worth noting that the size of the region might also affect the effects of NoPUDO, and sensitivity analysis will be conducted in the numerical experiments.

In conclusion, Equation~\eqref{eq:y} and \eqref{eq:d} characterize the causal relationship between NoPUDO and traffic speed in a spatio-temporal manner, and the above equations are consistent with the causal graph discussed in Section~\ref{sec:structural equation models}. We further assume that the random errors $e_{v}^t$ and $\xi_{v}^t$ follow Assumption~\ref{as:noise}.

\begin{assumption}[Independent Noise]
\label{as:noise}
For any time interval $t$ and region $v$, we have the following equations hold.
\begin{eqnarray}
&&\mathbb{E}[e_{v}^t | \mathbf{Y}_v^{t-I: t-1}; \mathbf{Y}_{\mathcal{N}(v)}^{t-I:t-1}; d_v^t; \mathbf{W}_{v}^t] 
= 0 \label{eq:noise1}\\
&&\mathbb{E}[ \xi_{v}^t | \mathbf{D}_v^{t-I: t-1}; \mathbf{Y}_v^{t-I: t-1}; \mathbf{Y}_{ \mathcal{N}(v)}^{t-I:t-1}; \mathbf{W}_{v}^t)] = 0 \quad \label{eq:noise2}\\
&&e_{v}^t \sim U^e \left(\mathbf{Y}_v^{t-I: t-1}; \mathbf{Y}_{\mathcal{N}(v)}^{t-I:t-1}; d_v^t; \mathbf{W}_{v}^t \right)  \label{eq:noise12}\\
&&\xi_{v}^t \sim U^\xi \left( \mathbf{D}_v^{t-I: t-1}; \mathbf{Y}_v^{t-I: t-1}; \mathbf{Y}_{ \mathcal{N}(v)}^{t-I:t-1}; \mathbf{W}_{v}^t\right) \label{eq:noise22}
\end{eqnarray}

$U^e$ and $U^\xi$ are unknown and parameterized zero-mean distributions.
\end{assumption}

Intuitively, Assumption~\ref{as:noise} indicates that unknown random errors in predicting $y_v^t$ and $d_v^t$ are zero-mean and independent. Specifically, Equation~\eqref{eq:noise1} and Equation~\eqref{eq:noise2} assume that the error terms $e_{v}^t$ and $\xi_{v}^t$ are white noise. It aims to guarantee that we can obtain an unbiased estimation of Equation~\eqref{eq:y} and~\eqref{eq:d}. Besides, Equation~\eqref{eq:noise12} and~\eqref{eq:noise22} assume that the error terms follow an unknown and parameterized zero-mean distribution. Hence, Assumption~\ref{as:noise} indicates that the distribution $U^e$ and $U^\xi$ are unknown and can be learned from data using machine learning models.

 
Based on the above formulation, we prove that when the traffic speed $y_v^t$, NoPUDO $d_v^t$, and external control variables $\mathbf{W}_v^t$ are observable, it is theoretically sufficient to estimate $\theta_v$, as presented in Proposition~\ref{prop:id}.

\begin{proposition}[Identifiable]
\label{prop:id}
Suppose that Equation~\eqref{eq:y}, \eqref{eq:d}, \eqref{eq:noise1}, and \eqref{eq:noise2} hold and $y_v^t$, $d_v^t$, and $\mathbf{W}_v^t$ are observable for all $v, t$, then $\theta_v$ is identifiable, {\em i.e.}, $\theta_v$ can be uniquely estimated from $y_v^t, d_v^t, \mathbf{W}_v^t, \forall v, t$.
\end{proposition}

\proof{Proof.}

First, given $y_v^t, d_v^t, \mathbf{W}_v^t, \forall v, t$ are observable, we have $\mathbf{Y}_v^{t-I: t-1}, \mathbf{Y}_{\mathcal{N}(v)}^{t-I:t-1}, \mathbf{D}_v^{t-I: t-1}$ are also observable. 
Second, in the time interval $t$ and for any region $v$, we consider the ordered pair of variables $(d_v^t, y_v^t)$, and we define $\mathcal{Z} = \{\mathbf{Y}_v^{t-I: t-1}, \mathbf{Y}_{\mathcal{N}(v)}^{t-I:t-1}, \mathbf{D}_v^{t-I: t-1}, \mathbf{W}_v^t \}$. We claim that $\mathcal{Z}$ satisfies the back-door criterion relative to $(d_v^t, y_v^t)$. The reason is: in the causal graph presented in Figure~\ref{fig:dsml_casual_graph}:
\begin{itemize}
    \item No node in $\mathcal{Z}$ is a descendant of $d_v^t$;
    \item $\mathcal{Z}$ blocks every path between $d_v^t$ and $y_v^t$ that contains an arrow into $y_v^t$.
\end{itemize}
Based on Theorem 3.3.2 in \citet{pearl2009causality}, the congestion effect $\theta_v$ is identifiable, and hence $\theta_v$ can be uniquely estimated based on the Definition 3.2.3 in \citet{pearl2009causality}.
\endproof

\subsection{Double and Separated Machine Learning}

In this section, we propose a novel method to estimate the congestion effect of PUDOs $\theta_v$ based on Equation~\eqref{eq:y} and \eqref{eq:d}. 
As we discussed in \ref{sec:causal}, the challenge in estimating $\theta_v$ lies in the complex spatio-temporal relationship between traffic speed and NoPUDO, as shown in Equation~\eqref{eq:y} and \eqref{eq:d}. To accurately model such a spatio-temporal relationship, both $\varphi_v$ and $\psi_v$ need to be generalized as non-linear functions that can model the arbitrary relationship between the traffic speed and NoPUDO. ML models shed light on modeling the non-linear relationship among variables with simple model specifications, and hence we propose to employ ML methods to learn both $\varphi_v$ and $\psi_v$ using massive data. 

When both $\varphi_v$ and $\psi_v$ are modeled as non-linear ML models, directly estimating $\theta_v$ becomes challenging. The main reason is that most ML models are biased due to model regularization \citep{hastie2009elements}. With the biased estimation of $\varphi_v$ and $\psi_v$, we need to estimate $\theta_v$ in an unbiased manner, and this presents challenges for the model formulation. To this end, we propose the DSML method with consideration of the potential biases in the ML models for $\varphi_v$ and $\psi_v$. The proposed DSML method consists of three sub-models: 1) Model $\mathtt{Y}$ learns $\varphi_v$ and predicts the traffic speed $y_v^t$; 2) Model $\mathtt{D}$ learns $\psi_v$ and predicts the NoPUDO $d_v^t$; and 3) Model $\mathtt{Z}$ estimates the congestion effect of PUDOs on traffic speed.

\begin{figure}[h]
    \centering
    \includegraphics[width=\linewidth]{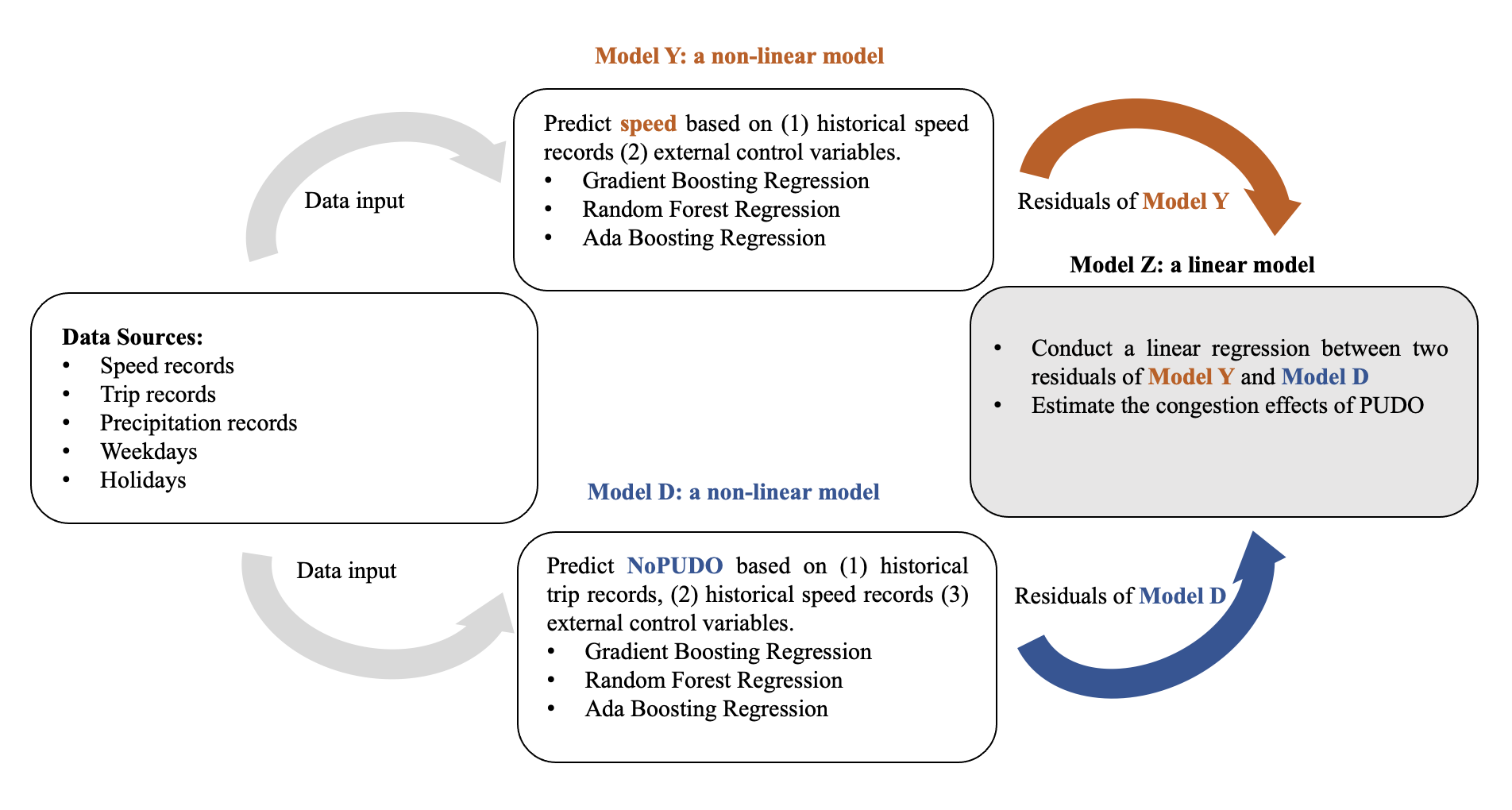}
    \caption{The framework of the DSML method.}
    \label{fig:dsml_framework}
\end{figure}

The relationship among the three sub-models is presented in Figure~\ref{fig:dsml_framework}. To be specific, we present each model as follows.
\begin{itemize}
\item Model $\mathtt{Y}$, which is denoted as $\hat{\varphi}_v$, predicts speed $y_v^t$ based on historical speed record $\mathbf{Y}_v^{t-I:t-1}$ in current region $v$, $\mathbf{Y}_{\mathcal{N}(v)}^{t-I:t-1}$ in the neighboring regions $\mathcal{N}(v)$, and external control variables $\mathbf{W}_v^t$, without considering the congestion effect of NoPUDO.
\item Model $\mathtt{D}$, which is denoted as $\hat{\psi}_v$, predicts the NoPUDO $d_{v}^t$ based on historical record of NoPUDO $\mathbf{D}_v^{t-I:t-1}$, speed record $\mathbf{Y}_v^{t-I:t-1}$, $\mathbf{Y}_{\mathcal{N}(v)}^{t-I:t-1}$, and external control variables $\mathbf{W}_v^t$. 
\item Model $\mathtt{Z}$ fits a linear regression model from the residuals of Model $\mathtt{D}$ to the residuals of Model $\mathtt{Y}$, and the slope is the estimation of $\theta_v$. Proof and intuitive explanations will be provided in the following sections.
\end{itemize}

In each sub-model, both spatial and temporal dependencies of variables are considered. One can see that in the DSML method, the task of estimating $\theta_v$ is decomposed into Model \texttt{Y}, \texttt{D}, and \texttt{Z} respectively.
We note that the DSML method is an extension of the generalized DML method~\citep{chernozhukov2018double}, and the DSML method is specifically designed for the congestion effect estimation using the causal graph in Figure~\ref{fig:dsml_casual_graph}. We further elaborate on the underlying mechanism regarding how the DSML method learns causality rather than correlation in the Appendix~\ref{Causality analysis of DSML}. In the following sections, we will present each sub-model in detail.

\subsubsection{ Model $\mathtt{Y}$ }

Model $\mathtt{Y}$ predicts the traffic speed using historical speed data without considering the congestion effect caused by PUDOs, as formulated in Equation~\eqref{eq:model y}.

\begin{equation}
\label{eq:model y}
    \hat{y}_{v}^t = \hat{\varphi}_v(\mathbf{Y}_v^{t-I: t-1};\mathbf{Y}_{\mathcal{N}(v)}^{t-I:t-1}; \mathbf{W}_v^t)
\end{equation}
where, $\hat{y}_v^t$ is the predicted speed in the time interval $t$ and region $v$. Three input variables include a vector of history speed record $\mathbf{Y}_v^{t-I:t-1}$ from the time interval $t-I$ to the time interval $t-1$ in the region $v$, historical average speed record $\mathbf{Y}_{\mathcal{N}(v)}^{t-I:t-1}$ from the time interval $t-I$ to the time interval $t-1$ at neighboring regions $\mathcal{N}(v)$, and the external control variables $\mathbf{W}_v^t$.
$\hat{\varphi}_v$ is the function that maps these input variables to the speed $y_v^t$, which can be learned by ML models using massive observed data. 

The residual of Model $\mathtt{Y}$, $\hat{\epsilon}_v^t$ is the difference of the predicted value $\hat{y}_v^t$ and the true value $y_v^t$, as shown in Equation~\eqref{eq:residual of model y}.

\begin{equation}
\label{eq:residual of model y}
\hat{\epsilon}_v^t = y_v^t - \hat{y}_v^t\\
\end{equation}

The residual $\hat{\epsilon}_v^t$ deserves more attention, as it is a random variable that consists of two sources of variation: 1) the changes of $y_v^t$ due to the NoPUDO, and 2) the other random noise. Intuitively, $\hat{\epsilon}_v^t =  \theta_v d_v^t + \left[ \varphi_v(\cdots) - \hat{\varphi}_v(\cdots) \right] + e_{v}^{t} \approx \theta_v d_v^t + e_v^t$. To extract the $\theta_v$ from $\hat{\epsilon}_v^t$, we make use of the Model \texttt{D} to capture the correlation between $\hat{\epsilon}_v^t$ and $d_v^t$. 

\subsubsection{Model $\mathtt{D}$ }

Model $\mathtt{D}$ aims to predict the NoPUDO using the historical traffic speed and NoPUDO, and the formulation is presented in Equation~\eqref{eq:model d}. 

\begin{equation}
\label{eq:model d}
\hat{d}_{v}^t  =  \hat{\psi}_v \left( \mathbf{D}_v^{t-I: t-1}, \mathbf{Y}_v^{t-I: t-1}, \mathbf{Y}_{\mathcal{N}(v)}^{t-I:t-1}, \mathbf{W}_{v}^t \right)\\
\end{equation}

where $\hat{d}_v^t$ is the predicted value of NoPUDO in the time interval $t$ and region $v$. 
 
Based on the causal graph in Figure~\ref{fig:dsml_casual_graph}, $\hat{d}_v^t$ is not only affected by the historical traffic speed ($\mathbf{Y}_v^{t-I:t-1}, \mathbf{Y}_{\mathcal{N}(v)}^{t-I:t-1}$), but also affected by the historical NoPUDO ($\mathbf{D}_v^{t-I: t-1}$).

Similarly, the function, $\hat{\psi}_v$, is obtained by training historical data. One important note is that the training data for Model \texttt{D} should be different from that is used for training Model \texttt{Y}, so that the learned $\hat{\varphi}_v$ and $\hat{\psi}_v$ are independent of each other~\citep{chernozhukov2018double}. To this end, the data splitting technique is adopted, and details will be introduced in Section~\ref{sec:dsmlsol}.

The residual $\hat{\xi}_v^t$ of Model $\mathtt{D}$ can be computed as the subtraction of true value $d_v^t$ and predicted value $\hat{d}_v^t$ of the NoPUDO, as shown in Equation~\eqref{eq:residual of model d}. 

\begin{equation}
\label{eq:residual of model d}
\hat{\xi}_v^t = d_v^t - \hat{d}_v^t\\
\end{equation}

The residual $\hat{\xi}_v^t$ is a random variable, and it contains the proportion of $d_v^t$ that is not affected by the historical traffic speed. Intuitively, $\hat{\xi}_v^t$ and $\hat{\epsilon}_v^t$ are correlated because of the congestion effect of PUDOs, and we have Proposition~\ref{prop:ind} holds.

\begin{proposition}
\label{prop:ind}
Given a region $v$, suppose Equation~\eqref{eq:y}, \eqref{eq:d}, and Assumption \ref{as:noise} hold, when $\mathbf{D}_v^{t-I: t-1}$, $\mathbf{Y}_v^{t-I: t-1}$, $\mathbf{Y}_{\mathcal{N}(v)}^{t-I:t-1}$, and $\mathbf{W}_{v}^t$ are observed for any $t$, we have
\begin{equation}
\theta_v = 0    \iff  \hat{\xi}_v^t \indep \hat{\epsilon}_v^t,
\end{equation}
where $\indep$ means independence. 

\end{proposition}

\proof{Proof.}
Based on Equation~\eqref{eq:y} and \eqref{eq:residual of model d}, we have $\hat{\epsilon}_v^t =  \theta_v d_v^t + \left[ \varphi_v(\cdots) - \hat{\varphi}_v(\cdots) \right] + e_{v}^{t} $ and $\hat{\xi}_v^t = \left[ \psi_v \left( \cdots \right) - \hat{\psi}_v \left( \cdots \right) \right]+ \xi_{v}^t$, where we use $\cdots$ to represent the input variables. We show the proposition from two directions:
\begin{itemize}
    \item When $\theta_v = 0$, we have $\hat{\epsilon}_v^t =  \left[ \varphi_v(\cdots) - \hat{\varphi}_v(\cdots) \right] + e_{v}^{t}$. Additionally, $\left[ \varphi_v(\cdots) - \hat{\varphi}_v(\cdots) \right] \indep e_{v}^{t}$, which is because $e_v^t$ is the random noise. Then we have $\hat{\xi}_v^t \indep \left[ \varphi_v(\cdots) - \hat{\varphi}_v(\cdots) \right]$ due to the data splitting technique, and $\hat{\xi}_v^t \indep e_{v}^{t}$ due to Equation~\eqref{eq:noise12} and \eqref{eq:noise22}. Therefore, $\hat{\xi}_v^t \indep \hat{\epsilon}_v^t$.
    \item When $\hat{\xi}_v^t \indep \hat{\epsilon}_v^t$, we know $\hat{\xi}_v^t \indep \theta_v d_v^t + \left[ \varphi_v(\cdots) - \hat{\varphi}_v(\cdots) \right] + e_{v}^{t}$. Again $\hat{\xi}_v^t \indep e_{v}^{t}$ and $\hat{\xi}_v^t \indep \left[ \varphi_v(\cdots) - \hat{\varphi}_v(\cdots) \right]$ hold, so $\hat{\xi}_v^t \indep \theta_v d_v^t$. Because $\hat{\xi}_v^t$ and $d_v^t$ are correlated due to Equation~\eqref{eq:d}, then $\theta_v$ has to be zero.
\end{itemize}
Combining the above two directions, we have the proof completed. 
\endproof

One can see from Proposition~\ref{prop:ind}, the correlation of $\hat{\xi}_v^t$ and $\hat{\epsilon}_v^t$ is closely associated with the value of $\theta_v$. Indeed, $\theta_v$ can be estimated from the two residuals, as presented in the next section.

\subsubsection{ Model $\mathtt{Z}$ }

Based on the discussions in the previous two sections, both Model $\mathtt{Y}$ and Model $\mathtt{D}$ depict the trends of traffic speed and NoPUDO using the spatio-temporal historical data, respectively. Importantly, all the edges were modeled in the causal graph in Figure~\ref{fig:dsml_casual_graph}, except for the congestion effect of PUDOs, which is marked with $\star$. To estimate the congestion effect $\theta_v$, we develop Model $\mathtt{Z}$ that fits a linear regression model from the residual $\hat{\xi}_v^t$ of Model $\mathtt{D}$ to the residuals $\hat{\epsilon}_v^t$ of Model $\mathtt{Y}$, as represented by Equation~\eqref{eq:lr}.

\begin{equation}
\label{eq:lr}
   \hat{\epsilon}_v^t = \theta_v \hat{\xi}_v^t + \hat{e}_v^t
\end{equation}
where $\hat{e}_v^t$ represents the random error of the linear regression model. We note that $\theta_v$ can be estimated using the Ordinary Least Square (OLS), as presented in Equation~\eqref{eq:ols}.

\begin{equation}
\label{eq:ols}
\hat{\theta}_v = \argmin_{\theta} \mathbb{E}\left[ \sum_{t\in \mathbb{T}} \left( \hat{\epsilon}_v^t - \theta \hat{\xi}_v^t \right)^2  \right]
\end{equation}
We claim that $\hat{\theta}_v$ is an unbiased estimator of $\theta_v$. Before the rigorous proof, we intuitively explain why this claim is true. To this end, the variable $\hat{e}_v^t$ can be derived in Equation~\eqref{eq:ev}.

\begin{equation}
\label{eq:ev}
\begin{array}{lllll}
\hat{e}_v^t = \hat{\epsilon}_v^t - \theta_v \hat{\xi}_v^t &=& \theta_v d_v^t + \left[ \varphi_v(\cdots) - \hat{\varphi}_v(\cdots) \right] + e_{v}^{t} -   \theta_v \left( \left[ \psi_v \left( \cdots \right) - \hat{\psi}_v \left( \cdots \right) \right]+ \xi_{v}^t \right)\\
&=& \left( \theta_v d_v^t - \theta_v \left( \psi_v \left( \cdots \right) + \xi_{v}^t\right) \right) + \left[ \varphi_v(\cdots) - \hat{\varphi}_v(\cdots) \right] + \theta_v \hat{\psi}_v \left( \cdots \right) + e_{v}^t\\
&=& \left[ \varphi_v(\cdots) - \hat{\varphi}_v(\cdots) \right] + \left[ \theta_v \hat{\psi}_v \left( \cdots \right)\right] + e_{v}^t\\
&\approx& \left[ -\theta_v d_v^t\right] + \left[\theta_v d_v^t \right] + e_v^t\\
&=&e_v^t
\end{array}
\end{equation}
where $\varphi_v(\cdots) - \hat{\varphi}_v(\cdots) = -\theta_v d_v^t$ because $\hat{\varphi}_v(\cdots)$ is a ML model to predict $y_v^t$, and $\theta_v \hat{\psi}_v \left( \cdots \right) = \theta_v d_v^t$ because $\hat{\psi}_v$ is a ML model to predict $d_v^t$. Therefore, $\hat{e}_v^t$ is zero-mean, and hence $\theta_v$ can be estimated using linear regression from $\hat{\xi}_v^t$ to $\hat{\epsilon}_v^t$.

Now we are ready to present Proposition~\ref{prop:linear}, which proves that $\hat{\theta}_v$ is an unbiased estimator of $\theta_v$ when $\varphi_v$ and $\psi_v$ are linear models.

\begin{proposition}[FWL Theorem]
\label{prop:linear}
For any region $v$, we suppose Equation~\eqref{eq:y}, \eqref{eq:d}, and Assumption \ref{as:noise} hold. When $\varphi_v$ and $\psi_v$ are linear models, $\hat{\theta}_v$ obtained from Equation~\eqref{eq:ols} is an unbiased estimator of $\theta_v$. Mathematically, we have $\hat{\theta}_v = \theta_v$.

\end{proposition}

\proof{Proof.}
See Appendix~\ref{ap:linear}. 
\endproof

We further extend to consider both $\varphi_v$ and $\psi_v$ are non-linear functions and can be learned by ML models, as presented in Proposition~\ref{prop:nonlinear}.

\begin{proposition}
\label{prop:nonlinear}
For any region $v$, we suppose Equation~\eqref{eq:y}, \eqref{eq:d}, and Assumption \ref{as:noise} hold. Given both $\varphi_v$ and $\psi_v$ are learnable by the ML models, we have Equation~\eqref{eq:converge} holds.

\begin{equation}
\begin{array}{lllll}
\frac{1}{|\mathbb{T}|} \sum_{t \in \mathbb{T}} \left( \hat{\varphi}_v - \varphi_v  \right)^2 &\overset{P}{\to}&0 \\
\frac{1}{|\mathbb{T}|} \sum_{t \in \mathbb{T}} \left( \hat{\psi}_v - \psi_v  \right)^2 &\overset{P}{\to}&  0    
\end{array}
\label{eq:converge}
\end{equation}
where $\overset{P}{\to}$ represents the convergence in probability. If  $\hat{\varphi}_v$ and $\hat{\psi}_v$ are learned with data splitting technique, then $\hat{\theta}_v$ obtained from Equation~\eqref{eq:ols} follows Equation~\eqref{eq:conv2}.
\begin{equation}
\label{eq:conv2}
\hat{\theta}_v - \theta_v \sim \mathbf{N} \left(0, \frac{1}{|\mathbb{T}|} \right)
\end{equation}
where $\mathbf{N} \left(0, \frac{1}{|\mathbb{T}|} \right)$ denotes the normal distribution with mean zero and variance $\frac{1}{|\mathbb{T}|}$.
\end{proposition}

\proof{Proof.}
See Appendix~\ref{ap:nonlinear}.
\endproof

Both Proposition~\ref{prop:linear} and \ref{prop:nonlinear} support the claim that the DSML method can estimate $\theta_v$ in an unbiased manner. Proposition~\ref{prop:linear} is actually a special case of Proposition~\ref{prop:nonlinear} with more intuitive explanations, which could help readers better understand the essential idea of the proposed DSML method. 

\subsection{Re-routing traffic flow with PUDOs to reduce total travel time}

In this section, we present to re-route traffic flow with PUDOs to minimize the network-wide total travel time. Currently, PUDOs are mainly concentrated in bustling regions such as office buildings, shopping malls, and residential areas. The uneven spatial distribution of PUDOs cause varying congestion in specific regions~\citep{zhang2023ride, dong2022strategic}. Consequently, one unit of the PUDO induces a more significant congestion effect in those busy regions, which further exacerbates the congestion. Taking the Manhattan area as an example, the $|\theta_v|$ in Midtown is typically higher than that in Upper West Side, and hence the congestion caused by PUDOs in Midtown is more severe. 

To reduce total travel time on the entire city road network, this paper aims to re-route parts of the traffic flow with PUDOs to the neighboring regions based on the differences of congestion effects in different regions. To be specific, part of travelers are required to 1) walk from their origin regions to the nearby regions and get picked up, and/or 2) get dropped off in nearby regions, and then walk to their destination regions. The underlying idea behind the above re-routing strategy is to re-distribute PUDOs from regions with severe congestion effect to neighboring regions with less congestion effect. Example~\ref{ex:reroute} further illustrates how the re-routing strategy reduces the total travel time.

\begin{figure}[h]
    \centering
    \includegraphics[width=0.7\linewidth]{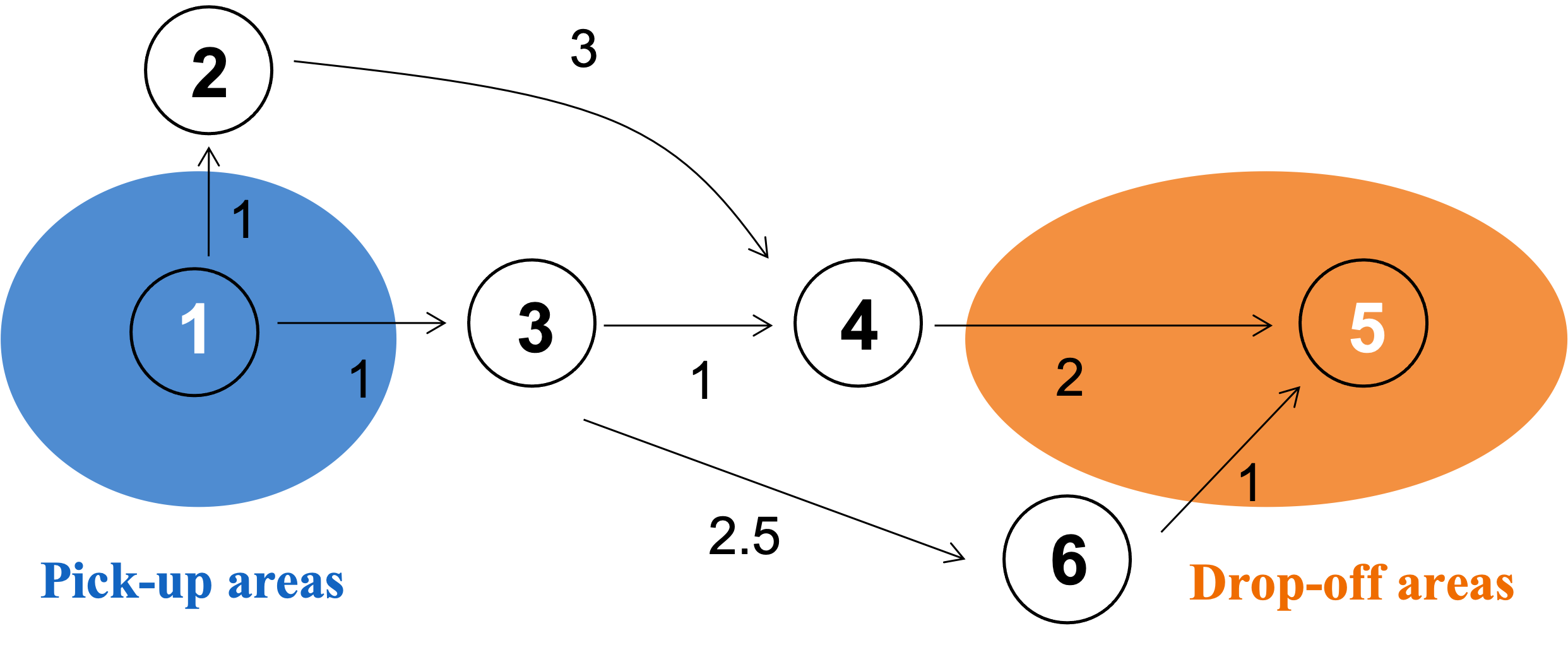}
    \caption{An example of the traffic flow re-routing with PUDOs.}
    \label{fig:illustration of rerouting}
\end{figure}

\begin{example}
\label{ex:reroute}
Consider a transportation network with 6 regions, which are represented by 6 nodes in Figure~\ref{fig:illustration of rerouting}. Weights on each link represent the time cost to drive from the tail to the head of the link. Region 5 has a severe congestion effect of PUDOs, while Region 4 and 6 are adjacent regions with less congestion effect. Therefore we assume the absolute congestion effect of Region 5, $|\theta_5|$, is larger than that of Region 4 $|\theta_4|$ and Region 6 $|\theta_6|$. An additional passenger departs from Region 1 to Region 5: if the passenger arrives at Region 5 by vehicle directly, the average speed in Region 5 will decrease, and hence the travel time in this region will increase accordingly. Instead, if the passenger is required to get dropped off in Region 4 or 6 and continue to walk to Region 5, traffic speed in Region 5 will increase. Finally, although the traffic speed in Region 4 and 6 will be reduced, the sacrificed traffic condition of the above regions contributes to the improved traffic condition in the busy Region 5, thus reducing the whole travel time at the city level.
\end{example}

One can see that this example utilizes the uneven geographical distribution of PUDOs, which is attributed to the common phenomenon of uneven travel demands \citep{zhang2023ride, dong2022strategic}. The differences in congestion effects among different regions can be exploited to re-distribute PUDOs, finally resulting in a decrease of the overall travel time. Specifically, when a great number of passengers flock to the same Central Business District (CBD), even a subtle improvement in travel time for each passenger will bring significant improvements to the entire network.

\begin{figure}[h]
    \centering
    \includegraphics[width=0.6\linewidth]{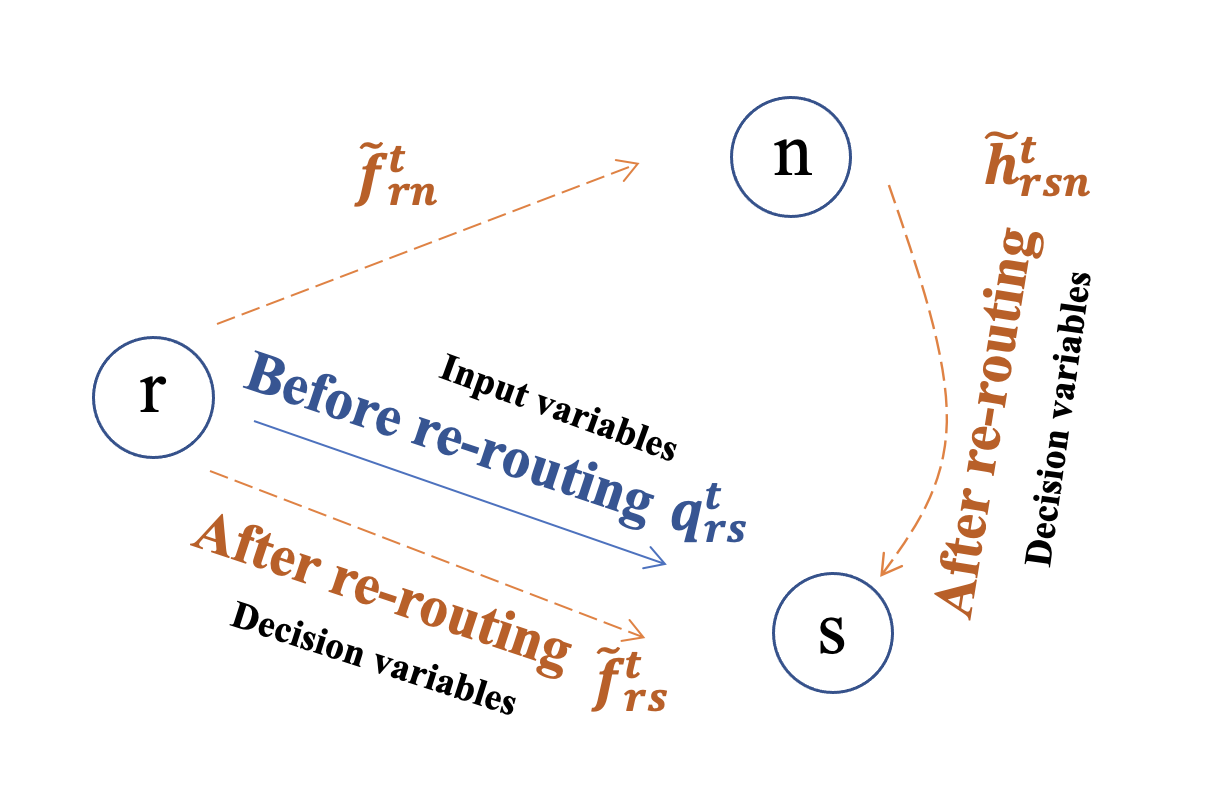}
    \caption{ Illustration of variables related to the re-routing formulation.}
    \label{fig:illustration of rerouting variables}
\end{figure}

We consider travelers from region $r$ to region $s$ in the time interval $t$, and their quantity is denoted as $q_{rs}^t$. These travelers are divided into two groups depending on whether they are re-routed or not. As shown in Figure \ref{fig:illustration of rerouting variables}, $\tilde{h}_{rsn}^t$ indicates the number of travelers whose original path is from region $r$ to region $s$, will be re-routed to drop off in region $n$, and these travelers need to walk from region $n$ to their final destination $s$. Other travelers, which is denoted as $\tilde{f}_{rs}^{t}$, will keep their original routes by vehicles directly. After re-routing, the NoPUDO in each region will be updated, and hence the travel time in each region will be adjusted with the congestion effect $\theta_v$. Ultimately, the re-routing of the traffic flow is expected to reduce the total travel time (TTT) on the network.

To ensure the feasibility of the re-routing strategy in actual traffic scenarios, we limit the walking distance and assume that the drop-off region $n$ belongs to the destination region's neighboring regions, {\em i.e.}, $n \in \mathcal{N}(s)$, where $\mathcal{N}(s)$ represents the set of neighboring regions of the region $s$. The mathematical formulation for re-routing the traffic flow with PUDOs in the time interval $t$ is as presented in Formulation~\eqref{eq:rerouting}. 

\begin{equation}
\label{eq:rerouting}
\begin{array}{rrcllll}
\vspace{5pt}
\displaystyle \min_{ \{\tilde{f}_{rs}^t\}_{rst} , \{\tilde{h}_{rsn}^t\}_{rsnt} } & \multicolumn{4}{l}{\displaystyle  \sum_{r \in \mathcal{R}} \sum_{s \in \mathcal{R}} \tilde{f}_{rs}^t \tilde{m}_{rs}^t + \sum_{r \in \mathcal{R}} \sum_{s \in \mathcal{R}} \sum_{n \in \mathcal{N}(s)} \tilde{h}_{rsn}^t  \tilde{c}_{rsn}^t}\\
\text{s.t.} & \displaystyle \tilde{f}_{rs}^t + \sum_{n \in \mathcal{N}(s)} \tilde{h}_{rsn}^t & = & q_{rs}^t & \forall r, s \\ 
~ & \tilde{f}_{rs}^t & \geq & 0 & \forall r, s \\
~ & \tilde{h}_{rsn}^t & \geq & 0 & \forall r, s, n \\
~ & \left( \{\tilde{m}_{rs}^t \}_{rst}, \{\tilde{c}_{rsn}^t\}_{rsnt}, \{\tilde{d}_s^t \}_{st} \right) &=& \Psi \left(\{\tilde{f}_{rs}^t\}_{rst},  \{\tilde{h}_{rsn}^t\}_{rsnt} \right) 
\end{array}
\end{equation}

The objective function of the formulation~\eqref{eq:rerouting} is to minimize the total travel time (TTT) consisting of two branches of traffic flow $\tilde{f}_{rs}^t$ and $\tilde{h}_{rsn}^t$ in the time interval $t$, which are the decision variables. $\tilde{f}_{rs}^t$ represents the traffic flow that remains on the original routes and $\tilde{h}_{rsn}^t$ presents the traffic flow whose final destination is region $s$ and the drop-off location is region $n$, $n \in \mathcal{N}(s)$. Specifically, The traffic flow $\tilde{f}_{rs}^t$ means the number of vehicles who stay on the original routes from origin region $r$ to the destination region $s$ in the time interval $t$. The traffic flow $\tilde{h}_{rsn}^t$ means the number of vehicles who depart from region $r$ to one temporary destination $n$, $n \in \mathcal{N}(s)$, requiring passengers to continue to walk from region $n$ to the final destination $s$. $q_{rs}^t$ represents the total traffic flow from region $r$ to region $s$ before re-routing in the time interval $t$. It can be calculated by summing the number of vehicles departing from region $r$ to the region $s$ at time interval $t$ given trip records data. $\Psi$ translates the two branches of traffic flow ($\tilde{f}_{rs}^t, \tilde{h}_{rsn}^t$) into $(\{\tilde{m}_{rs}^t \}_{rst}, \{\tilde{c}_{rsn}^t\}_{rsnt}, \{\tilde{d}_s^t \}_{st})$, where $\tilde{m}_{rs}^t$ is the travel time of $\tilde{f}_{rs}^t$, $\tilde{c}_{rsn}^t$ is the travel time of $\tilde{h}_{rsn}^t$, and $\{\tilde{d}_s^t \}_{st}$ is the NoPUDO in region $s$ and time interval $t$. To understand the objective function more accurately, we decompose it into three parts, as discussed in Proposition~\ref{prop:decompose}.

\begin{proposition}[Total travel time decomposition]
\label{prop:decompose}
The change of total travel time (TTT) after the re-routing using Formulation~\eqref{eq:rerouting} can be decomposed into three parts, as presented in Equation~\eqref{eq:decompose}.
\begin{equation}
\label{eq:decompose}
\Delta TTT = \Delta_\text{Counterfactual} + \Delta_\text{PUDO, Remain} + \Delta_\text{PUDO, Detour} 
\end{equation}
where $\Delta TTT$ denotes the change of TTT after the re-routing (after minus before), $\Delta_\text{Counterfactual}$ represents the change of the TTT after re-routing if the congestion effect of PUDOs is zero, $\Delta_\text{Remain}$ represents the change of the TTT after re-routing for the travelers staying on their original routes, and $\Delta_\text{Detour}$ represents the change of TTT after re-routing for the travelers taking the detours. To be specific, we have Equation~\eqref{eq:4parts} holds.

\begin{equation}
\label{eq:4parts}
\begin{array}{llll}
\Delta_\text{Counterfactual} &=& \sum_{r\in \mathcal{R}} \sum_{s \in \mathcal{R}} \left( \tilde{f}_{rs}^t m_{rs}^t + \sum_{n \in \mathcal{N}(s)} \tilde{h}_{rsn}^t\left(m_{rn}^t  + m_{ns}^t \right) - q_{rs}^t m_{rs}^t\right)\\
\Delta_\text{PUDO, Remain} &=& \sum_{r \in \mathcal{R}}\sum_{s \in \mathcal{R}} \tilde{f}_{rs}^t\left(   \tilde{m}_{rs}^t -m_{rs}^t\right)\\
\Delta_\text{PUDO, Detour} &=&\sum_{r \in \mathcal{R}}  \sum_{s \in \mathcal{R}} \sum_{n \in \mathcal{N}(s)} \tilde{h}_{rsn}^t \left(\tilde{m}_{rn}^t  - m_{rn}^t + w_{ns} - m_{ns}^t \right)  
\end{array}
\end{equation}
\end{proposition}

\proof{Proof.}
See Appendix~\ref{sec:proofde}.
\endproof

We expect $\Delta TTT < 0$, which means that the TTT after the re-routing is smaller than the current situation without re-routing. In general, $\Delta_\text{Counterfactual} > 0$ because rational travelers prefer selecting the shortest paths. Additionally, $\Delta_\text{PUDO, Detour} > 0$ because the traffic flow increases on the detour routes, and walking usually takes longer time than driving. To make $\Delta TTT < 0$, we need to make $\Delta_\text{PUDO, Remain}  <  - |\Delta_\text{Counterfactual} + \Delta_\text{PUDO, Detour} | < 0$. That means, the re-routing should reduce the travel time for travelers staying on their original routes. The reduced TTT for the travelers staying on the original routes should be larger than the increased TTT for the travelers taking the detours. 

We further discuss $\Psi$, which can be formulated as a series of constraints, as shown in Equation~\eqref{eq:psi}.

\begin{subequations}
\begin{eqnarray}
\label{eq:o_s}
d_s^t & = & \sum_{r \in \mathcal{R}} q_{rs}^t\\
\label{eq:tilde_o_s}
\tilde{d}_{s}^t & = & {\displaystyle \sum_{r \in \mathcal{R}}} \tilde{f}_{rs}^t + {\displaystyle \sum_{r \in \mathcal{R}} \sum_{n \in \mathcal{N}(s)}}  \tilde{h}_{rns}^t\\
\beta d_s^t &\leq& \tilde{d}_s^t  \leq  \gamma {d}_s^t   \label{eq:delta_con} \\
\label{eq:delta_o_s}\Delta_s^t & = & \tilde{d}_s^t - d_s^t\\
\label{eq:v_s}\tilde{y}_s^t & = & y_s^t + \hat{\theta}_s \Delta_s^t\\
\label{eq:tilde_t_rs}\tilde{m}_{rs}^t & = & {\displaystyle \sum_{v \in \mathcal{L}_{rs}} l_v} / \tilde{y}_v^t \\
\label{eq:w_as}{u}_{ns} & = & {\displaystyle \sum_{v \in \mathcal{L}_{ns}}} l_v / k\\
\label{eq:c_rsn}\tilde{c}_{rsn}^t &=&  \tilde{m}_{rn}^t + u_{ns}
\end{eqnarray}
\label{eq:psi}
\end{subequations}
where $\mathcal{L}_{rs}$ is the set of regions listed in the shortest path from origin $r$ to destination $s$, indexed by $v$. $\Psi$ consists of two parts: 1) limiting the change of NoPUDO; 2) calculating the travel time after re-routing, as discussed below:
\begin{itemize}
    \item Firstly, we count the region-based NoPUDO before and after the re-routing strategy separately, as shown in Equation~\eqref{eq:o_s} and Equation~\eqref{eq:tilde_o_s}. Before re-routing, the NoPUDO $d_s^t$ is equal to the summation of all traffic flow whose destination is in region $s$, as shown in Equation~\eqref{eq:o_s}. After re-routing, the NoPUDO consists of two branches of traffic flows. The traffic flows in the first branch keep their original path, and another branch of traffic flows is scheduled to drop off at one of the neighboring regions of the destination region firstly and require passengers to walk to their final destination. The updated NoPUDO $\tilde{d}_s^t$ sums two branches of traffic flow, {\em i.e.}, $\tilde{f}_{rs}^t$ and $\tilde{h}_{rns}^{t}$, whose PUDOs location is region $s$, as shown in Equation~\eqref{eq:tilde_o_s}. 
    
    \item Secondly, a constraint on the quantity of traffic flows that can be re-routed is added. The change of NoPUDO in the region $s$ at the time interval $t$ before and after re-routing has been captured in Equation~\eqref{eq:delta_o_s}. The updated NoPUDO $\tilde{d}_s^t$ is within $[\beta d_s^t, \gamma d_s^t]$, as presented by Equation~\eqref{eq:delta_con}, where $\beta \leq 1, \gamma \geq 1$ are hyper-parameters to limit the change of NoPUDO.

    \item Thirdly, the redistribution of NoPUDO before and after the re-routing strategy determines the reshuffle of the speed in each region. Given a region, the change of NoPUDO at time interval $t$ will affect the speed at time interval $t$ by the congestion effect obtained by the DSML method. It means the speed after re-routing $\tilde{y}_s^t$ will increase by the product of the change of PUDO $\Delta_s^t$ unit and congestion effect $\hat{\theta}_s$ on the basis of the speed before re-routing $y_s^t$. Therefore, we formulate how the region-based speed changes after the re-routing strategy in Equation~\eqref{eq:v_s}.

    \item Fourthly, given the fixed travel distance from the origin to the destination region, the travel time of each trip record is reshaped by the change in speed accordingly. Because the traffic flows $\tilde{f}_{rs}^{t}$ will not be re-routed but the traffic flows $\tilde{h}_{rsn}^t$ are involved with the re-routing, we distinguish how to calculate the travel time for two different traffic flows by whether or not to take the walking time into account. As for the traffic flows $\tilde{f}_{rs}^{t}$, the travel time from region $r$ to region $s$ after re-routing is equal to the travel distance divided by the updated speed as shown in Equation~\eqref{eq:tilde_t_rs}. $l_v$ is the average trip distance in the region $v$. Regarding the traffic flows $\tilde{h}_{rsn}^t$, the total travel time includes the travel time on vehicle and walking. Because the re-routing strategy requires certain passengers to continue walking from the instant drop-off location to their destination. Therefore, the additional walking duration $u_{ns}$ from region $n$ to region $s$ is calculated in Equation~\eqref{eq:w_as}, where $k$ is the average walking speed. Lastly, the total travel time for the re-routed flow $\tilde{h}_{rsn}^t$ is calculated as the summation of travel time from $r$ to $n$ and from $n$ to $s$, as shown in Equation~\eqref{eq:c_rsn}.
\end{itemize}

Overall, Formulation~\eqref{eq:rerouting} is non-linear programming as the objective function contains the product of $\tilde{f}_{rs}^t$ and $\tilde{m}_{rs}^t$, as well as the product of $\tilde{h}_{rsn}^t$ and $\tilde{c}_{rsa}^t$. The travel time $\tilde{m}_{rs}^t$ is also proportional to the reciprocal of $\tilde{y}_v^t$, as shown in Equation~\eqref{eq:tilde_t_rs}. Given a large-scale network, the number of decision variables $\{\tilde{f}_{rs}^t\}_{rst}, \{\tilde{h}_{rsn}^t\}_{rst}$ can be large, making it difficult to solve by applying standard non-linear programming solvers. In the following sections, we will present a customized solution algorithm to solve Formulation~\eqref{eq:rerouting} effectively.

\section{Solution algorithms}
\label{sec:solution algorithm}

This section presents two solution algorithms. First, we design and implement the solution algorithm of the DSML method based on its theoretical structures. Secondly we develop a new algorithm to solve the re-routing formulation, which splits the solving process into two sub-processes and solves both sub-processes iteratively.

\subsection{Solving the DSML method}
\label{sec:dsmlsol}

To align with the proof of the DSML method, $\hat{\varphi}_v$ and $\hat{\psi}_v$ should be independently trained, which is similarly required in the standard DML \citep{chernozhukov2018double}. To this end, we always divide a dataset into two disjoint parts: one for training model \texttt{Y} and the other for training model \texttt{D}. At the same time, we make use of the $b$-fold cross-validation to select the optimal ML models and hyper-parameters in DSML. The detailed algorithm for DSML is presented in Algorithm~\ref{alg:DSML programming}.

\begin{algorithm}[H]
	\SetKwInOut{Input}{Input}
	\SetKwInOut{Output}{Output}
	\Input{$\{y_v^t\}_{vt}, \{d_v^t\}_{vt}, \{\mathbf{W}_v^t\}_{vt}$, candidates of ML models, ranges of hyper-parameters}
	\Output{$\{\hat{\theta}_v\}_v$}
\For{$v \in \mathcal{V}$}{    
    Construct $\mathbf{Y}_v^{t-I:t-1}$ with $y_v^{t-I}$, $y_v^{t-I+1}$, $\cdots$,  $y_v^{t-1}$, $\forall t$.
    
    Construct $\mathbf{D}_v^{t-I:t-1}$ with $d_v^{t-I}$, $d_v^{t-I+1}$, $\cdots$,  $d_v^{t-1}$, $\forall t$.
    
    Construct $\mathbf{Y}_{\mathcal{N}(v)}^{t-I:t-1}$ by averaging the speed of the neighboring regions $\mathcal{N}(v)$, $\forall t$.
    
    Combine $y_v^t$, $\mathbf{Y}_v^{t-I:t-1}$, $\mathbf{Y}_{\mathcal{N}(v)}^{t-I:t-1}$, $d_v^t$, $\mathbf{D}_v^{t-I:t-1}$, $\mathbf{W}_v^t$ for all $t$ to construct the entire dataset $\mathcal{D}$.
	
	Split the constructed dataset into $b$ sub-datasets randomly. We denote $\mathcal{D}_i$ as the $i$-th sub-dataset, and $\mathcal{D}_{-i} = \mathcal{D} / \mathcal{D}_i$, where $\mathcal{D}$ is the entire dataset, $i=0, \cdots, b-1$. 

	\For{$i=0; i<b; i++$}{
	    
	    Train Model $\mathtt{Y}$ by applying each of the candidate ML models with different hyper-parameter settings on the first half of $\mathcal{D}_{-i}$.
	    
        Train Model $\mathtt{D}$ by applying each of the candidate ML models with different hyper-parameter settings on the second half of $\mathcal{D}_{-i}$. 
	    
	    Select the optimal candidate ML model and hyper-parameter setting for Model $\mathtt{Y}$ and Model $\mathtt{D}$ respectively based on the performance on $\mathcal{D}_i$.
	    
	    Obtain the predicted values of $\hat{y}_v^t$ and $\hat{d}_v^t$ by running the trained Model $\mathtt{Y}$ and Model $\mathtt{D}$ on $\mathcal{D}_i$.
	    
        Calculate the residual $\hat{\epsilon}_v^t$ of the Model $\mathtt{Y}$ on $\mathcal{D}_i$.
        
        Calculate the residual $\hat{\xi}_v^t$ of the Model $\mathtt{D}$ on $\mathcal{D}_i$.
	}
	
	Merge residuals $\hat{\epsilon}_v^t$ and $\hat{\xi}_v^t$ from each sub-dataset $\mathcal{D}_i, \forall i$.
	
    Estimate $\hat{\theta}_v$ by OLS between $\hat{\epsilon}_v^t$ and $\hat{\xi}_v^t$.
}
Return $\{\hat{\theta}_v\}_v$.
	\caption{Solution algorithm to the DSML method.}
	\label{alg:DSML programming}
\end{algorithm}

In this paper, the candidate ML models include Gradient Boosting Regression, Random Forest Regression, and Ada Boosting Regression. The ranges of hyper-parameters are set based on the recommendation of \texttt{scikit-learn}.

\subsection{Solving the re-routing formulation}

As discussed above, Formulation~\eqref{eq:rerouting} is a non-linear program with high-dimensional decision variables on large-scale networks. To solve the formulation, we view $\tilde{m}_{rs}^t$ as an intermediate variable. With $\tilde{m}_{rs}^t$ known and fixed, Formulation~\eqref{eq:rerouting} reduces to a standard linear program. Additionally, $\tilde{m}_{rs}^t$ can be updated using the decision variables $(\tilde{f}_{rs}^t, \tilde{h}_{rsn}^t)$ with closed-form equations. Based on the above observations, we develop a solution algorithm to conduct the following two steps iteratively until convergence: 1) fix $\tilde{m}_{rs}^t$ and $\tilde{c}_{rsn}^t$, solve the simplified Formulation~\eqref{eq:rerouting} as a linear program to obtain $(\tilde{f}_{rs}^t, \tilde{h}_{rsn}^t)$; 2) use the solved $(\tilde{f}_{rs}^t, \tilde{h}_{rsn}^t)$ to update $\tilde{m}_{rs}^t$ based on Equation~\eqref{eq:psi}. Details of the algorithm are presented in Algorithm~\ref{alg:nonlinear_programming}. 

\begin{algorithm}[H]
	\SetKwInOut{Input}{Input}
	\SetKwInOut{Output}{Output}
	\Input{ $\{m_{rs}^t\}_{rst}$, $\{c_{rsn}^{t}\}_{rsnt}$, $\{q_{rs}^t\}_{rst}$, $\{d_s^t\}_{st}$, $\{y_s^t\}_{st}$, $\{\hat{\theta}_s\}_s$, $\{\mathcal{L}_{rs}\}_{s}$.} 
	\Output{ $\{\tilde{f}_{rs}^t\}_{rst}$ and $\{\tilde{h}_{rsn}^t\}_{rsnt}$.} 
\For {$t \in \mathbb{T}$}{	
	Initialize $\{\tilde{f}_{rs}^t\}_{rs}$ and $\{\tilde{h}_{rsn}^{t}\}_{rsn}$ such that the constraints of Formulation~\eqref{eq:rerouting} are satisfied. \\
	
	\While{changes of $\tilde{f}_{rs}^t$ and $\tilde{h}_{rsn}^t$ are within tolerances}{
	    
	    Update $\{\tilde{m}_{rs}^t\}_{rs}$ and $\{\tilde{c}_{rsn}^t\}_{rsn}$ based on Equation~\eqref{eq:psi}. \\
	    
	    Calculate and record the objective function $\sum_{r} \sum_{s} \tilde{f}_{rs}^t \tilde{m}_{rs}^t + \sum_{r} \sum_{s} \sum_{n} \tilde{h}_{rsn}^{t} \tilde{c}_{rsn}^{t}$. \\
	    
	    Solve Formulation~\eqref{eq:rerouting} as a linear program problem by fixing $\{\tilde{m}_{rs}^t\}_{rs}$ and $\{ \tilde{c}_{rsn}^{t} \}_{rsn}$.\\
	    
	   Obtain the solution results $\{\check{f}_{rs}^t\}_{rs}$ and $\{\check{h}_{rsn}^{t}\}_{rsn}$ after solving the above linear program.\\
	    
	   Update $\{\tilde{f}_{rs}^t\}_{rs}$ and $\{\tilde{h}_{rsn}^{t}\}_{rsn}$ by gradient descent with momentum, and the gradients are calculated as $\tilde{f}_{rs}^t - \check{f}_{rs}^t, \forall rs$ and $ \tilde{h}_{rsn}^{t} - \check{h}_{rsn}^{t}, \forall rsn$, respectively.  \\
    }
    }
    Return $\{\tilde{f}_{rs}^t\}_{rst}$ and $\{\tilde{h}_{rsn}^t\}_{rsnt}$.
    \caption{Solution algorithm to the re-routing formulation.}
    \label{alg:nonlinear_programming}
\end{algorithm}

We set the parameter for momentum to be $0.8$, and the tolerance is set to be $1e-3$ in terms of $\ell_2$-norm. To improve the running efficiency of Algorithm~\ref{alg:nonlinear_programming}, all the variables involved will be vectorized. For example, to efficiently evaluate Equation~\eqref{eq:psi}, we use an incidence matrix to replace the summations in Equation~\eqref{eq:tilde_o_s}, \eqref{eq:tilde_t_rs}, and \eqref{eq:w_as}. Matrix multiplications are much faster than loop-based summations on multi-core CPUs and GPUs, and hence the solution efficiency can be enhanced. For details of the vectorization procedures, readers are referred to \citet{ma2018estimating, ma2020estimating}.

\section{Numerical Experiments}
\label{sec:numberical experiment}

In this section, we examine the effectiveness of the DSML method and re-routing formulation in the Manhattan area. We will first present the estimation results obtained by the DSML method, followed by the optimization results in the re-routing formulation.

\subsection{Estimating the congestion effect of PUDOs}
\label{sec:Estimating the congestion effect of PUDOs}

In this section, numerical experiments of applying the DSML method in Manhattan are presented and discussed. We first describe the datasets used in the study, which include the NoPUDO, traffic speed, and precipitation. Subsequently, the estimation results in the Manhattan area are presented and discussed. Finally, we study the underlying estimation mechanism of the DSML method by analyzing its residuals of ML models.

\subsubsection{Data description}
\label{data_description}

\begin{figure}[H]
    \centering
    \includegraphics[width=0.6\linewidth]{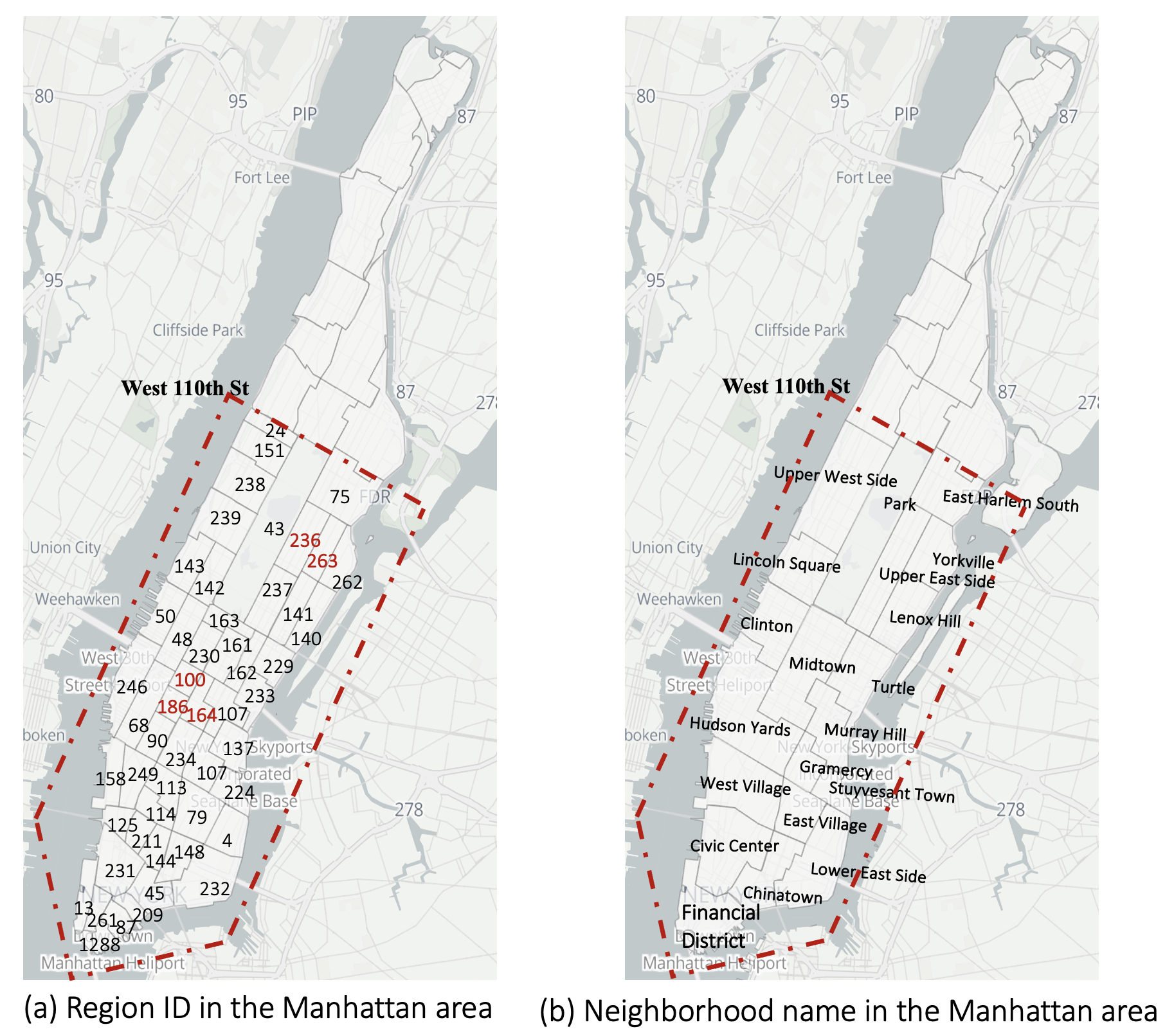}
    \caption{Map of the study area.}
    \label{fig:map}
\end{figure}

We fence 52 regions below West 110th Street in the Manhattan area to be our study area, as shown in Figure~\ref{fig:map}. Because travel demands in Manhatten are mainly concentrated in these fenced regions, estimating the congestion effect of PUDOs in these regions is increasingly important and lays a solid foundation for implementing the re-routing strategy to mitigate the congestion effect of PUDO~\citep{kim2020stepwise}.

This study refers to the NYC Taxi Zone system to divide regions in Manhattan. The division of taxi zones relies on the NYC Department of City Planning’s Neighborhood Tabulation Areas (NTAs). Therefore, the areas with approximate characteristics, for example, similar economic situations, population density, and travel demand, are categorized into a region. The NYC Taxi Zone has been widely utilized in region-based studies~\citep{beirigo2022learning, chung2023platform}. In this study, we apply the DSML method to estimate the region-based congestion effect based on the NYC Taxi Zone. If the region consists of similar areas, such as with similar road conditions, weather, and traffic regulations, it can mitigate the selection bias of manually composing sub-areas for a specific region. Besides, using the Taxi Zones also benefits controlling potential confounding factors, for example, uneven distribution of road conditions and different regulations of sub-areas within a region, when we estimate the congestion effect. Therefore, the adoption of the NYC Taxi Zone guarantees the stability of the estimation result by reducing selection bias in manually composing the region and controlling potential confounding factors.

We focus on the congested traffic conditions during afternoon peak hours from 16:00 to 20:00. Datasets used in this study include traffic speed, trip records, and precipitation from Feb 2019 to Jun 2020, as shown in Table~\ref{tb:datasets_information}. In this study, we use the time interval of 5 minutes as the unit of analysis, which is because the congestion effect estimation requires aligning the traffic speed and trip record datasets into the same time granularity. The finest time granularity of the traffic speed and trip records datasets is every 5 minutes and 1 second, respectively. Restricted by the different time granularities of both two datasets, the time interval, every 5 minutes, is the finest available unit of analysis in this study. If the time granularity of the datasets allows, higher-resolution analysis can be conducted, which will be left for future studies. Besides, we only include and consider the PUDOs generated by passengers of taxi cabs and ride-hailing vehicles, due to the data limitation. The developed framework can be extended trivially to incorporate the PUDOs from public transit and private cars if data allows. 

\begin{table}[H]
\caption{Data Description}
\label{tb:datasets_information}
\begin{tabular}{m{3cm} m{3cm} m{2.4cm} m{1.8cm} m{3.7cm}}
\toprule
Datasets & Time Range & Resolution & Quantity & Descriptions \\ [1.2ex] 
\hline
NYC traffic speed & Feb 2019 - Jun 2020 & every 5 mins & 404,351,029
 & Road segment, traffic speed, free flow speed, time stamp \\\hline
NYC trip records & Feb 2019 - Jun 2020 & every 5 mins & 18,157,071  & Pick-up region ID, drop-off region ID, time stamp \\\hline
NYC precipitation & Feb 2019 - Jun 2020 &every 1 hour& 11,987  & precipitation, time stamp  \\
\bottomrule
\end{tabular}
\end{table}

The detailed descriptions and data processing procedures for each dataset are as follows:

\begin{itemize}
     \item NYC speed data:
     The speed data contains several key fields including road ID, traffic speed, free flow speed, and timestamp. The traffic speed on each road is obtained based on probe vehicles, and the free flow speed is the average speed if there is no congestion or other adverse conditions on the road segment. To normalize the speed data for the DSML method, we calculate the relative speed $y_v^t$, as shown in Equation~\eqref{eq:12}. The relative speed of region $v$ is determined by the speed and free flow speed on each road and how many roads are in the current region.

\begin{equation}
\label{eq:12}
y_v^t  =  {\displaystyle \sum_{i \in v} \text{traffic speed}_i * \text{traffic flow}_i} / {\displaystyle \sum_{i \in v} \text{traffic flow}_i}
\end{equation}

The traffic flow in Equation~\eqref{eq:12} is determined by the free flow speed of the current road. $i$ represents the road belonging to the region $v$. As for the numerator, we first obtain the product of the speed and the corresponding traffic flow for each road segment $i$. We summarize all products of speed and traffic flow for each road $i$ in region $v$. As for the denominator, we calculate the sum of the traffic flow of each road $i$, which belongs to the region $v$. Equation~\eqref{eq:12} makes the traffic speed at the region level become normalized.

Note that the relative speed $y_v^t$ is only used for training the DSML method. When calculating the TTT, we will transform the relative speed back to the actual traffic speed.
     
     \item NYC trip records:
     Trip order information from the New York City Taxi \& Limousine Commission (NYC-TLC) contains timestamps, vehicle types, pick-up region ID, and drop-off region ID. These trip orders come from four types of vehicles: yellow taxis, green taxis, For-Hire Vehicles (FHV), and High Volume For-Hire Vehicles (HVFHV). The NoPUDO $d_v^t$ in the region $v$ every 5 minutes can be calculated as the number of orders starting or ending in the current region $v$ in the given time interval $t$.

     \item NYC precipitation:
     Iowa Environmental Mesonet monitors the precipitation information in the Manhattan area every hour, and we use the volume of rainfall as the indicator of weather, denoted as $\mathbf{W}_v^t$. 
     
   \end{itemize}

\subsubsection{Estimation results by the DSML method}
\label{sec:estimation result by DSML}

\begin{figure}[H]
    \centering
    \includegraphics[width=0.8\linewidth]{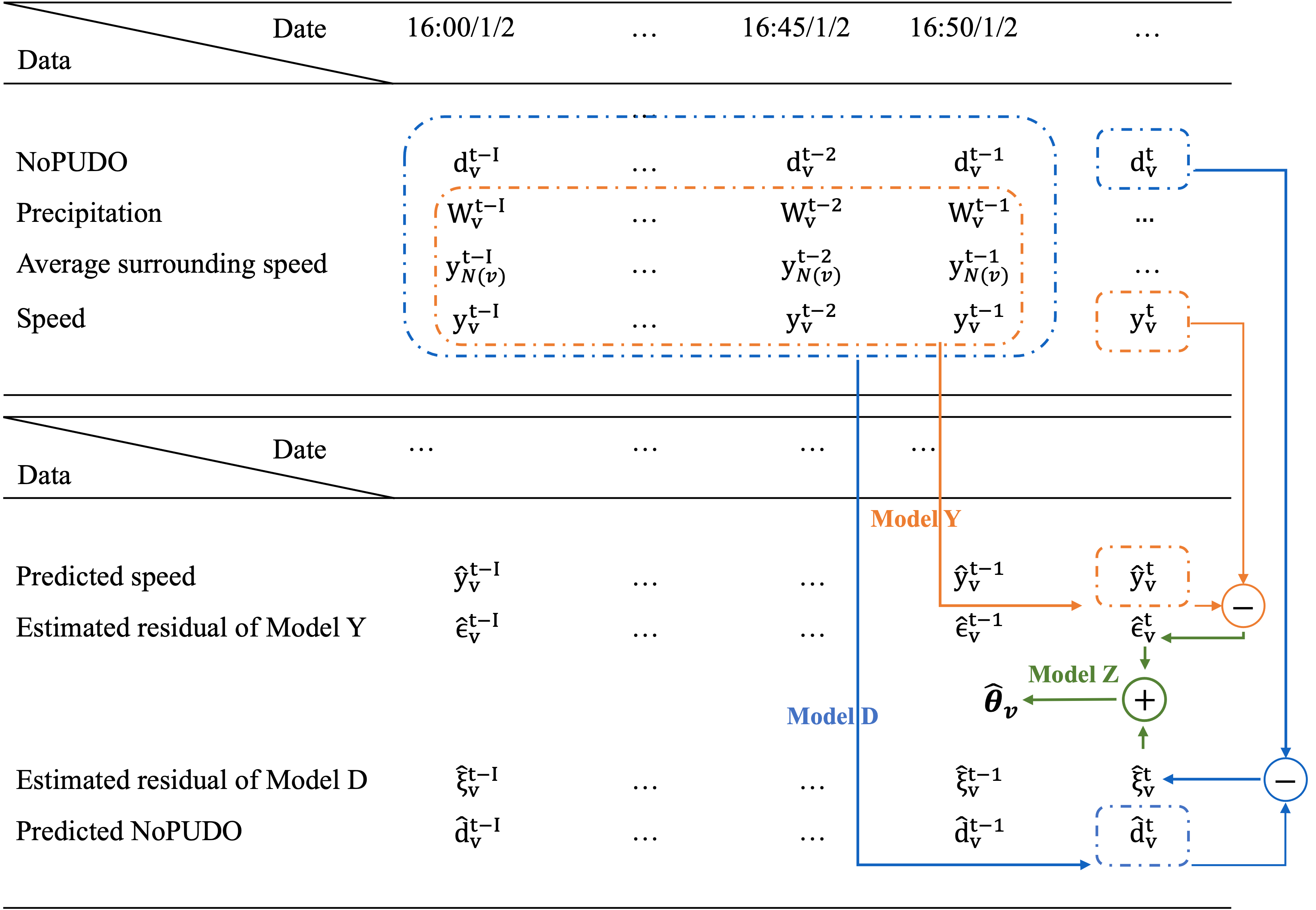}
    \caption{Variable relation of DSML for NYC data.}
    \label{fig:dsml_architecture}
\end{figure}

We implement and apply the DSML method in the Manhattan area to empirically demonstrate its effectiveness. An illustration of the variable relationship in the DSML method is shown in Figure~\ref{fig:dsml_architecture}. The upper table records the observed traffic data, and the lower table denotes the predicted values and residuals obtained from Model $\mathtt{Y}$ and Model $\mathtt{D}$. In our experiments, we set $I=10$, {\em i.e.}, the historical data ranges from time $t-10$ to $t-1$. The residuals $\epsilon_v^t$, $\xi_v^t$ are obtained from the differences of prediction and true values of traffic speed and NoPUDO in Model $\mathtt{Y}$ and Model $\mathtt{D}$ separately. We regard $\epsilon_v^t$ as the dependent variable, $\xi_v^t$ as the independent variable into a linear regression in Model $\mathtt{Z}$, and the congestion effect $\hat{\theta}_v$ can be estimated based on Algorithm~\ref{alg:DSML programming}. 

After implementing and running Algorithm~\ref{alg:DSML programming}, we found that the average congestion effect $\theta_v$ is $-0.0370$ on weekdays and $-0.0454$ on weekends for all region $v$ in the Manhattan area. It means that when there are an additional 100 unit PUDOs happening in a single region, the average traffic speed in that region will decrease by $3.70$ mph on weekdays and $4.54$ mph on weekends. Additionally, all the p-values for the estimated $\hat{\theta}_v$ in the DSML method are below 0.01, which indicates the estimated $\hat{\theta}_v$ is highly significant. Besides, the value of $\hat{\theta}_v$ is negative, meaning that NoPUDO has a negative effect on traffic speed. Furthermore, the $\hat{\theta}_v$ is varied with different regions depending on unique attributes and properties in each region.

\begin{figure}[h]
    \centering
    \includegraphics[width=0.8\linewidth]{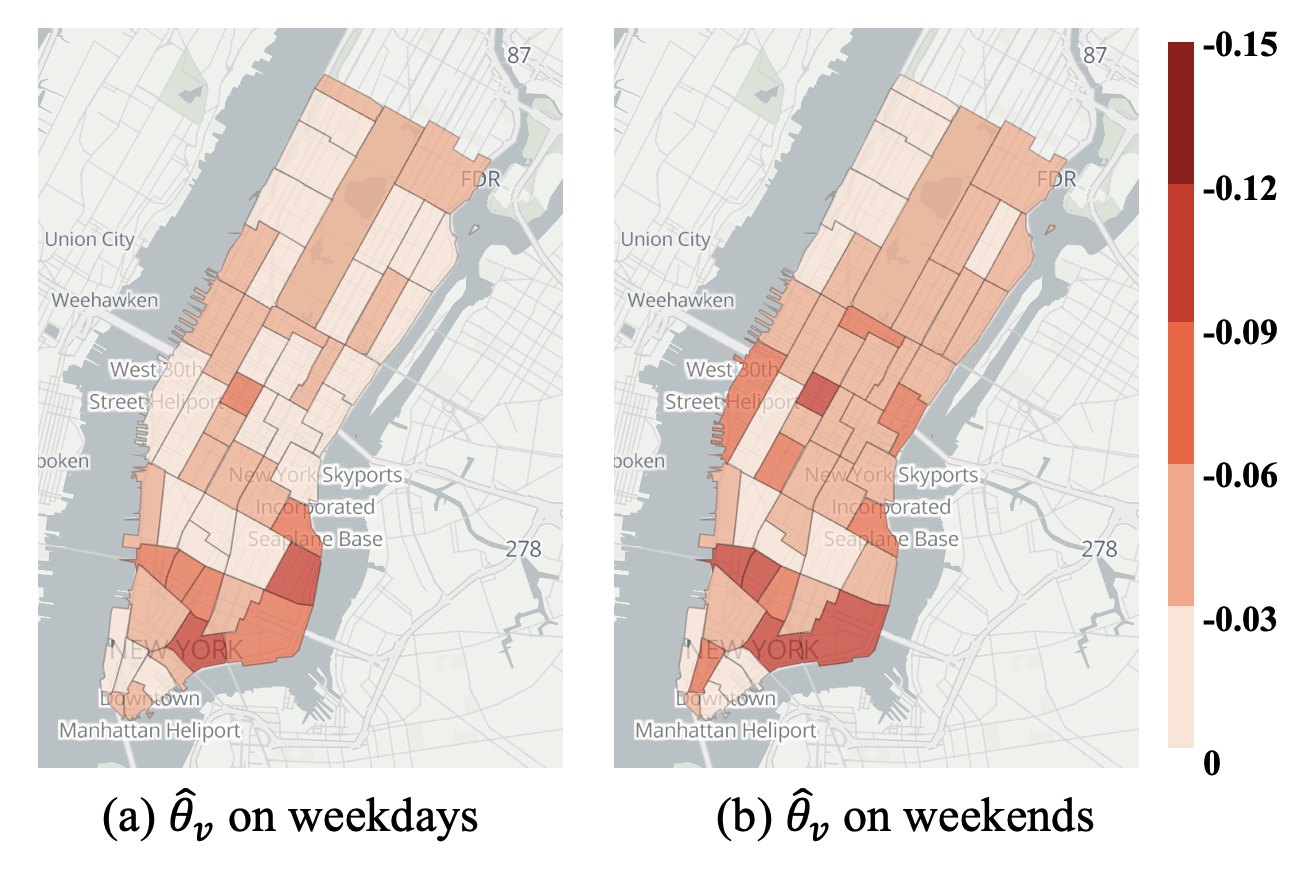}
    \caption{Overview of $\hat{\theta}$ learned from the DSML method and attractions in the Manhattan area.}
    \label{fig:Estimation_DSML}
\end{figure}

We visualize the spatial distribution of the estimated $\hat{\theta}_v$ on weekdays and weekends in the Manhattan map as shown in Figure~\ref{fig:Estimation_DSML}, respectively. 
In Figure~\ref{fig:Estimation_DSML} (a) and (b), deeper color indicates larger numerical values of $|\theta_v|$ and more severe congestion effects caused by PUDOs.



\begin{figure}[h]
    \centering
    \includegraphics[width=14cm]{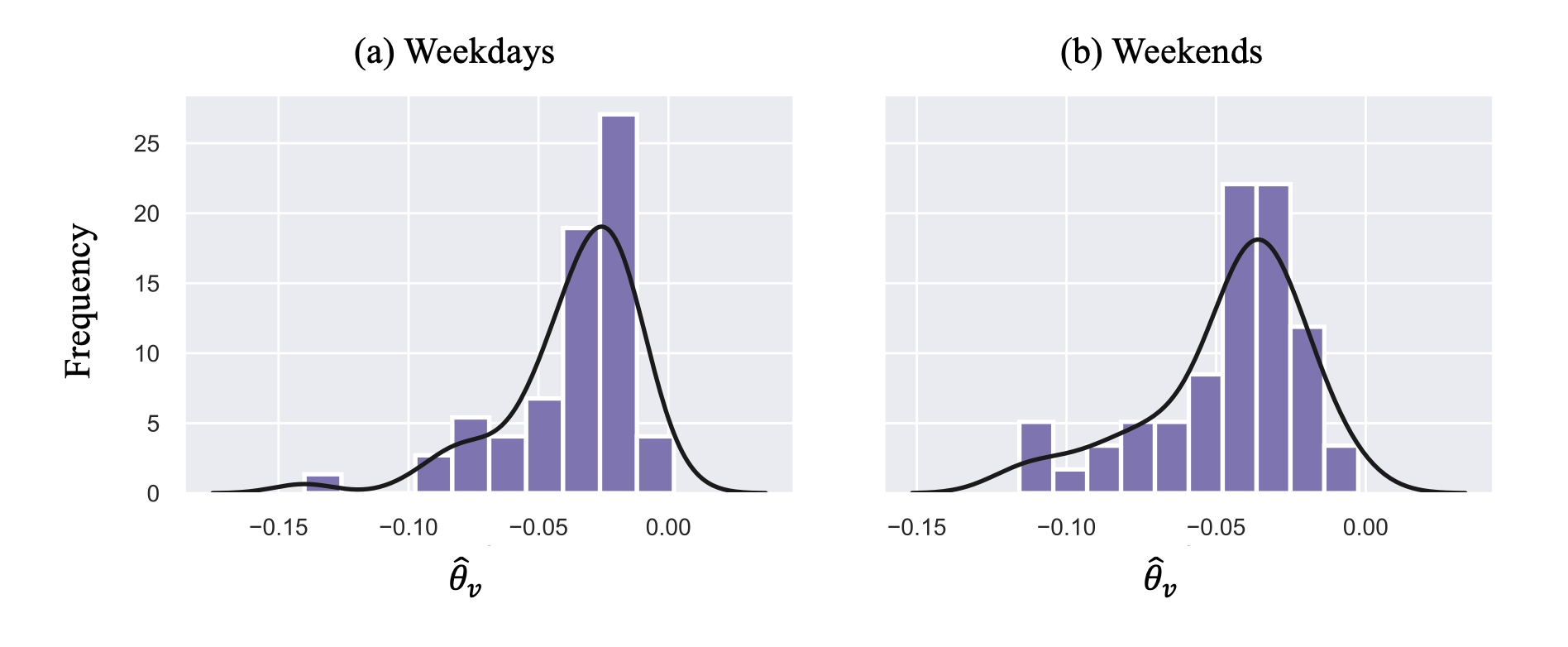}
    \caption{Histogram of $\hat{\theta}_v$ on weekdays and weekends.}
    \label{fig:density_DSML}
\end{figure}

The distributions of $\hat{\theta_v}$ on weekdays and weekends also vary significantly, as shown in Figure~\ref{fig:Estimation_DSML}(a) and (b). The congestion effect of PUDOs is more severe around points of interest (POIs) ({\em e.g.}, Times Square, Chinatown, and Brooklyn Bridge) on weekends than on weekdays, which is probably attributed to the frequent activities around sightseeing attractions during weekends. We further present the histogram of the estimated $\hat{\theta}_v$ for weekdays and weekends in Figure~\ref{fig:density_DSML}. One can see that $\hat{\theta}_v$ on weekends is more probable to be below $-0.10$, and the mode of $\hat{\theta}_v$ on weekends is smaller than that on weekdays.

To justify the robustness of the DSML method, we conduct two sensitivity analyses to examine the effects of ML methods selection and region sizes on estimation results in  Appendix~\ref{Sensitivity analysis regarding the selection of ML models} and Appendix~\ref{sec:sen_ana_region_size}, respectively. Besides, we compare the advanced DSML method with traditional estimation methods, such as DML and Linear Regression (LR) to illustrate its validity in the Appendix~\ref{Comparison among DSML, DML, and LR}. We also implement correlation analysis between the estimated congestion effect and the actual congestion map to illustrate the effectiveness of the DSML method in the Appendix~\ref{sec:Con_Corr}.

\subsubsection{Analyzing residuals of the DSML method}
\label{sec:residual}

We select two areas to validate the estimation results from the DSML method in detail. The first area is located in Midtown, which consists of Region 100 and 186; the second area is around Central Park, which consists of Region 236 and 263. The four regions will also be studied as cases in the re-routing formulation. 

\begin{figure}[h]
    \centering
    \includegraphics[width=0.6\linewidth]{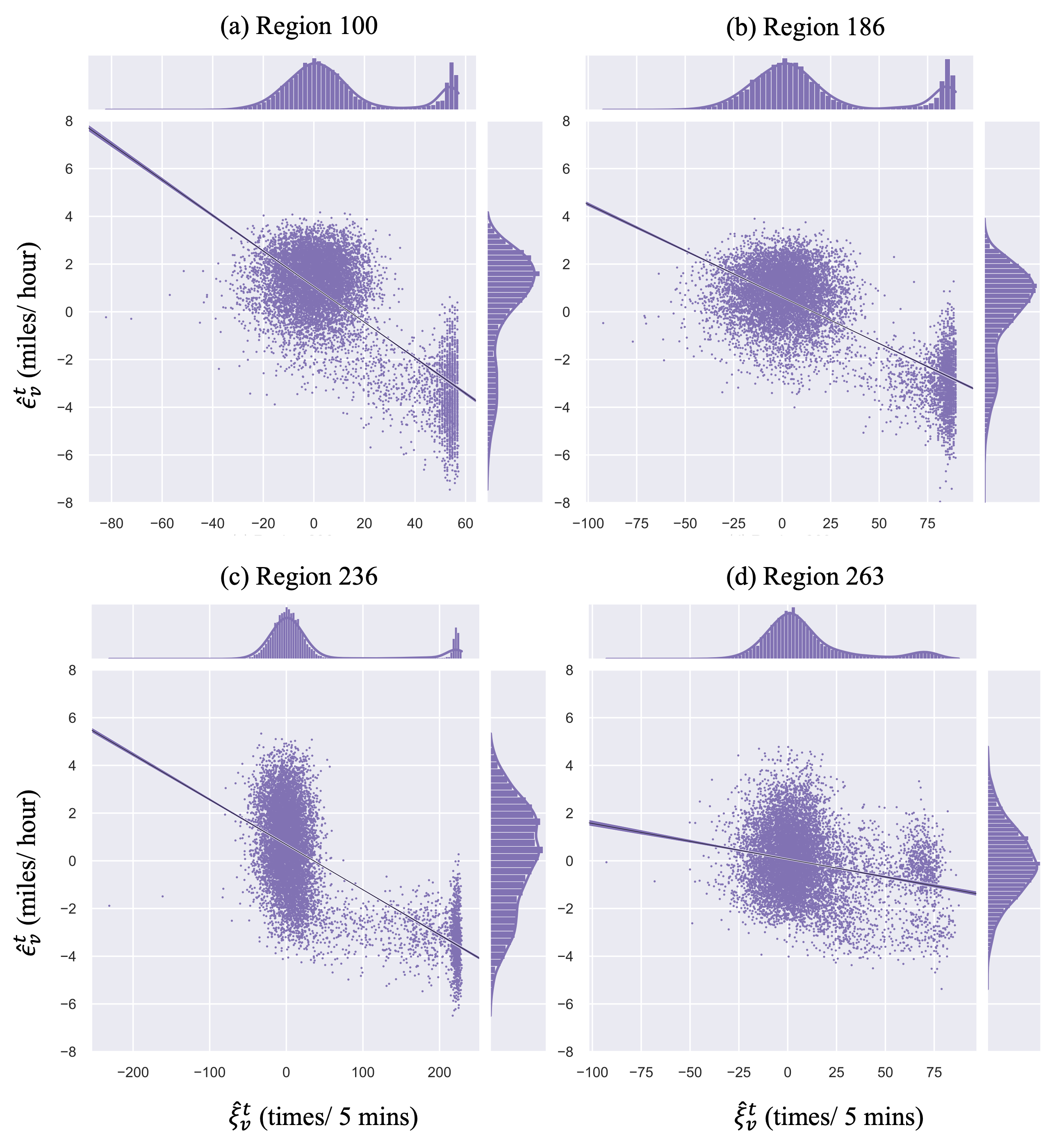}
    \caption{The fitted linear line in Model $\mathtt{Z}$.}
    \label{fig:residual_analysis}
\end{figure}

The fitted linear regression in Model $\mathtt{Z}$ on weekends for the four regions are separately shown in Figure~\ref{fig:residual_analysis}. The residuals of both Model $\mathtt{Y}$ and Model $\mathtt{D}$ are centered at the origin, which indicates that both above two models achieve a good fitting in prediction. More importantly, the two residuals $\hat{\epsilon}_v^t$ and $\hat{\xi}_v^t$ are negatively correlated, and this suggests a negative value of $\theta_v$. Specifically, it shows that the NoPUDO causes a negative congestion effect on traffic speed. Indeed, the slope of the fitted line is $\hat{\theta}_v$, and the t-test can be conducted to evaluate the significance of the estimated $\hat{\theta}_v$.

Additionally, we would like to clarify the potential reasons that might result in underestimation or overestimation of the DSML method. First, the linear assumption in Assumption~\ref{ap:01} might also lead to biased estimation in Model $\mathtt{Z}$. Especially, when we solve actual-world questions, the linear assumption might be violated. Using Figure~\ref{fig:residual_analysis}(a) as an example, we have  $\mathbb{E}\hat{\epsilon}_v^t < 0$ and $\mathbb{E}\hat{\xi}_v^t > 0$, which would lead to an underestimation of $\theta_v$ due to the mis-specificaiton of Assumption~\ref{ap:01} or violation of Assumption~\ref{as:noise}. Secondly, the performance of the machine learning models used in Model $\mathtt{Y}$ and Model $\mathtt{D}$ determines the distribution of residuals. If the machine learning models cannot fit the features and dependent variables well, it will lead to an inaccurate calculation of residuals. The regression in Model $\mathtt{Z}$ between two inaccurate residuals will result in the biased estimation of the congestion effect. However, quantifying the level of biasedness would be a challenging task for future studies.



\subsection{Solving the re-routing formulation}

In this section, we solve the re-routing formulation for some busy regions in the Manhattan area.

\subsubsection{Settings}

Four regions near Midtown and Central Park are selected as study areas, which are also used in Section~\ref{sec:residual}. We consider all the trips to these four regions. We consider the study period from 1st Jul 2019 to 30th Sep 2019, and time intervals 16:00--17:00, 17:00--18:00, 18:00--19:00, and 19:00--20:00 during the afternoon peaks are considered separately. The total traffic flow $q_{rs}^t$ in the Formulation~\eqref{eq:rerouting} can be calculated by summing the number of trip records whose pick-up region ID is region $r$ and drop-off region ID is region $s$ given the time interval. The total number of vehicles on the roads is set to be $\lambda$ times of the trip orders in the NYC datasets. We set $\beta = 0$, $l_v$ is calculated as the average travel distance with each region, $k$ is set as 3.5 mph, and $\hat{\theta}_v$ are estimated by DSML in Section~\ref{sec:Estimating the congestion effect of PUDOs}.

We calculate the improvement rate before and after re-routing based on Equation~\eqref{eq:13}.
\begin{equation}
\label{eq:13}
\text{improvement rate} = \frac{\text{TTT before re-routing} -\text{TTT after re-routing}}{\text{TTT before re-routing}} \times 100\%
\end{equation}

To evaluate the TTT after re-routing, we follow the steps in \citet{ma2020measuring}. We assume the hypothetical traffic conditions (in terms of travel time) after re-routing are calculated based on the changes of NoPUDO in each region, as presented in Equation~\eqref{eq:tilde_t_rs}. Only weekdays are considered as the results on weekends are similar.

\subsubsection{TTT after re-routing}

We run Algorithm~\ref{alg:nonlinear_programming} with $\lambda = 15$, and statistics for TTT are shown in Table~\ref{tb:weekdays_vehicle_upperbound}.

\begin{longtable}{p{2cm}p{4.5cm}p{4.5cm}p{3cm}}
\caption{TTT and improvement rates after re-routing on weekdays. (Mean$\pm$Std, $\lambda$ = 15)}
\label{tb:weekdays_vehicle_upperbound}

\endfirsthead
\endhead
    \hline
     & Before re-routing & After re-routing & Improvement rate\\
     & ($\times 10^{3}$ hours) & ($\times 10^{3}$ hours) & ($\%$)\\ 
    \hline
    \multicolumn{4}{c}{Midtown ($\gamma = 2.3$)}\\
    \hline
    \multicolumn{4}{c}{}\\
    Average &   4.41 $\pm$     0.71 & 4.30 $\pm$ 0.65  &    2.44 $\pm$    1.55\\
    16:00--17:00&   4.60 $\pm$  0.63 &  4.45 $\pm$  0.58 &     3.01 $\pm$    2.06\\
    17:00--18:00&   4.74 $\pm$  0.75 &  4.60 $\pm$  0.69  &    2.86  $\pm$   2.42\\
    18:00--19:00&   4.50 $\pm$  0.80 &  4.38 $\pm$  0.75  &     2.47  $\pm$   2.22\\
    19:00--20:00&   3.81 $\pm$  0.84 &  3.76 $\pm$  0.81  &    1.20  $\pm$    1.15\\
    \multicolumn{4}{c}{}\\
    \hline
    \multicolumn{4}{c}{Central Park ($\gamma = 1.6$)}\\
    \hline
    \multicolumn{4}{c}{}\\
    
    Average &   3.63 $\pm$ 0.75 &  3.54 $\pm$ 0.70  &    2.12 $\pm$    1.61\\
    16:00--17:00&   2.94 $\pm$  0.56 &  2.84 $\pm$  0.51  &    2.98 $\pm$    1.85\\
    17:00--18:00&   3.75 $\pm$  0.78 &  3.64 $\pm$  0.72 &    2.58 $\pm$    1.75\\
    18:00--19:00&   4.23 $\pm$  0.90 &  4.13 $\pm$  0.84  &    2.25 $\pm$    2.01\\
    19:00--20:00&   3.57 $\pm$  0.82 &   3.54 $\pm$  0.80  &    0.76 $\pm$    1.09\\
    \multicolumn{4}{c}{}\\
    \hline
\end{longtable}

We note that the mean and standard deviation in Table~\ref{tb:weekdays_vehicle_upperbound} are calculated based on the TTT of each day. One can see that the average improvement rate is $2.44 \%$ in Midtown and $2.12 \%$ in Central Park on weekdays. The improvements are more significant during 16:00--19:00 for both areas. The standard deviation is roughly half the mean, indicating the high randomness of network conditions. Overall, re-routing traffic flow with PUDO has great potential in reducing the total travel time for both areas and across all time periods. To provide a robust justification for the effectiveness of the re-routing strategy, we also conduct a sensitivity analysis of the re-routing formulation in Appendix~\ref{Sensitivity analysis of re-routing formulation}.

\section{Conclusion and Discussion}
\label{sec:conclusion}

This paper first time adopts the causal inference method to estimate the congestion effect of PUDOs with observational traffic data, and the estimated congestion effect can be further used to mitigate the congestion induced by PUDOs. To this end, the causal relationship between NoPUDO and traffic speed is identified through a causal graph, and the novel DSML method is developed to estimate the congestion effect of PUDOs based on the causal graph. Theoretical guarantees regarding the estimation results of DSML are also provided. To reduce the network-wide travel time, a re-routing formulation is developed and the corresponding solution algorithm is proposed. 

Experiments with real-world data in the Manhattan area demonstrate the effectiveness of the developed DSML method, and the estimation results align well with the actual traffic situations. On average, 100 additional units of the PUDO will decrease traffic speed by 3.70 mph on weekdays and 4.54 mph on weekends. The re-routing formulation also demonstrates great potential in reducing the total travel time. The improvement rate regarding the total travel time can reach 2.44\% in Midtown and 2.12\% in Central Park during weekdays.

As for future research directions, it holds great practical significance to incorporate different road attributes and properties in estimating the congestion effect of NoPUDO. For example, PUDOs are more likely to cause more severe congestion on a one-way and one-lane road with narrow curb space, while the congestion effect on large curb space might be negligible. This study mainly focuses on estimating the congestion effect of PUDOs based on regional levels. The region-based trip records data prevents this study from further validating the learned congestion effect on finer granular units, for example, at the street or road segment level. If there is available finer granular data about NoPUDO and traffic speed at street levels, we can reapply the DSML method to obtain a more accurate estimation of congestion effect based on the street level and the estimation results can be utilized for the downstream tasks, for example, designing curbside pricing and optimizing curb space. 
In addition, restricted by the time granularity of the data, we estimate congestion effects using the NoPUDO and traffic speed within the same time interval. If both second-based NoPUDO and traffic speed data are available, it would be interesting to conduct event-triggered causal estimation by modeling each sequence of PUDO and the corresponding changes in traffic speed. 
Furthermore, it would be essential to identify and distinguish the congestion effects of PUDOs from various vehicle types, and the re-routing formulation can also be tailored and customized for different vehicle types. For instance, compared to a single-rider vehicle, PUDOs from ride-sharing vehicles might be more likely to cause more congestion due to the longer-lasting duration of pick-up or drop-off. Developing and promoting personalized re-routing strategies for different types of vehicles could further reduce the total travel time. Additionally, as drop-offs usually take less time than pick-ups, we might consider modeling PU and DO separately in the DSML method to obtain a more accurate congestion effect of PU and DO, and designing the different re-routing strategies for PU and DO separately. Lastly, from the perspective of mathematically designing the DSML method, we will examine how to further relax the basic linear assumption to improve estimation accuracy and investigate how the different distribution assumptions of the error terms affect the estimation result in future studies.



\section*{Supplementary Materials}
\label{sec:supplementary materials}

The DSML method is implemented and the re-routing problem is solved in Python and open-sourced on GitHub (\url{https://github.com/LexieLiu01/DSML}).

\ACKNOWLEDGMENT{%
The work described in this paper was supported by the National Natural Science Foundation of China (No. 52102385),  grants from the Research Grants Council of the Hong Kong Special Administrative Region, China (Project No. PolyU/25209221 \& PolyU/15206322), a grant from the Otto Poon Charitable Foundation Smart Cities Research Institute (SCRI) at the Hong Kong Polytechnic University (Project No. P0043552). The second author was supported by a National Science Foundation grant CMMI-1931827. 
The contents of this paper reflect the views of the authors, who are responsible for the facts and the accuracy of the information presented herein. The authors thank the reviewer panel for thoughtful comments that substantially improved this study.
}

\bibliographystyle{informs2014trsc}
\bibliography{ref}

\begin{thebibliography}{80}
\providecommand{\natexlab}[1]{#1}
\providecommand{\url}[1]{\texttt{#1}}
\providecommand{\urlprefix}{URL }

\bibitem[{Abrevaya, Hsu, \protect\BIBand{} Lieli(2015)}]{abrevaya2015estimating}
Abrevaya J, Hsu YC, Lieli RP, 2015 \emph{Estimating conditional average treatment effects}. \emph{Journal of Business \& Economic Statistics} 33(4):485--505.

\bibitem[{Agarwal, Mani, \protect\BIBand{} Telang(2023)}]{agarwal2023impact}
Agarwal S, Mani D, Telang R, 2023 \emph{The impact of ride-hailing services on congestion: Evidence from indian cities}. \emph{Manufacturing \& Service Operations Management} 25(3):862--883.

\bibitem[{Ahmed \protect\BIBand{} Ghasemzadeh(2018)}]{ahmed2018impacts}
Ahmed MM, Ghasemzadeh A, 2018 \emph{The impacts of heavy rain on speed and headway behaviors: an investigation using the shrp2 naturalistic driving study data}. \emph{Transportation research part C: emerging technologies} 91:371--384.

\bibitem[{Amemiya \protect\BIBand{} Fuller(1967)}]{amemiya1967comparative}
Amemiya T, Fuller WA, 1967 \emph{A comparative study of alternative estimators in a distributed lag model}. \emph{Econometrica, Journal of the Econometric Society} 509--529.

\bibitem[{Angrist \protect\BIBand{} Pischke(2009)}]{angrist2009mostly}
Angrist JD, Pischke JS, 2009 \emph{Mostly harmless econometrics: An empiricist's companion} (Princeton university press).

\bibitem[{Anurag et~al.(2019)Anurag, Kalin, Mohammadreza, Pragun, \protect\BIBand{} Arun}]{Anurag2019curb}
Anurag K, Kalin P, Mohammadreza K, Pragun V, Arun K, 2019 \emph{Mind the curb: Findings from commercial vehicle curb usage in california}. \emph{Transportation Research Board (TRB) 98th Annual Meeting} .

\bibitem[{Arnott \protect\BIBand{} Rowse(2013)}]{arnott2013curbside}
Arnott R, Rowse J, 2013 \emph{Curbside parking time limits}. \emph{Transportation Research Part A: Policy and Practice} 55:89--110.

\bibitem[{Babar \protect\BIBand{} Burtch(2020)}]{babar2020examining}
Babar Y, Burtch G, 2020 \emph{Examining the heterogeneous impact of ride-hailing services on public transit use}. \emph{Information Systems Research} 31(3):820--834.

\bibitem[{Beirigo, Schulte, \protect\BIBand{} Negenborn(2022)}]{beirigo2022learning}
Beirigo BA, Schulte F, Negenborn RR, 2022 \emph{A learning-based optimization approach for autonomous ridesharing platforms with service-level contracts and on-demand hiring of idle vehicles}. \emph{Transportation Science} 56(3):677--703.

\bibitem[{Beojone \protect\BIBand{} Geroliminis(2021)}]{beojone2021inefficiency}
Beojone CV, Geroliminis N, 2021 \emph{On the inefficiency of ride-sourcing services towards urban congestion}. \emph{Transportation research part C: emerging technologies} 124:102890.

\bibitem[{Burtch, Carnahan, \protect\BIBand{} Greenwood(2018)}]{burtch2018can}
Burtch G, Carnahan S, Greenwood BN, 2018 \emph{Can you gig it? an empirical examination of the gig economy and entrepreneurial activity}. \emph{Management Science} 64(12):5497--5520.

\bibitem[{Butrina et~al.(2020)Butrina, Le~Vine, Henao, Sperling, \protect\BIBand{} Young}]{butrina2020municipal}
Butrina P, Le~Vine S, Henao A, Sperling J, Young SE, 2020 \emph{Municipal adaptation to changing curbside demands: Exploratory findings from semi-structured interviews with ten us cities}. \emph{Transport Policy} 92:1--7.

\bibitem[{Castiglione et~al.(2016)Castiglione, Chang, Cooper, Hobson, Logan, Young, Charlton, Wilson, Mislove, Chen et~al.}]{castiglione2016tncs}
Castiglione J, Chang T, Cooper D, Hobson J, Logan W, Young E, Charlton B, Wilson C, Mislove A, Chen L, et~al., 2016 \emph{Tncs today: a profile of san francisco transportation network company activity}. \emph{San Francisco County Transportation Authority (June 2016)} .

\bibitem[{Castiglione et~al.(2018)Castiglione, Cooper, Sana, Tischler, Chang, Erhardt, Roy, Chen, \protect\BIBand{} Mucci}]{castiglione2018tncs}
Castiglione J, Cooper D, Sana B, Tischler D, Chang T, Erhardt GD, Roy S, Chen M, Mucci A, 2018 \emph{Tncs \& congestion} .

\bibitem[{Chai \protect\BIBand{} Rodier(2020)}]{chai2020automated}
Chai H, Rodier C, 2020 \emph{Automated vehicles and central business district parking: The effects of drop-off-travel on traffic flow and vehicle emissions [supporting dataset]} .

\bibitem[{Chernozhukov et~al.(2017)Chernozhukov, Chetverikov, Demirer, Duflo, Hansen, \protect\BIBand{} Newey}]{chernozhukov2017double}
Chernozhukov V, Chetverikov D, Demirer M, Duflo E, Hansen C, Newey W, 2017 \emph{Double/debiased/neyman machine learning of treatment effects}. \emph{American Economic Review} 107(5):261--265.

\bibitem[{Chernozhukov et~al.(2018)Chernozhukov, Chetverikov, Demirer, Duflo, Hansen, Newey, \protect\BIBand{} Robins}]{chernozhukov2018double}
Chernozhukov V, Chetverikov D, Demirer M, Duflo E, Hansen C, Newey W, Robins J, 2018 \emph{Double/debiased machine learning for treatment and structural parameters}.

\bibitem[{Chung, Zhou, \protect\BIBand{} Ethiraj(2023)}]{chung2023platform}
Chung HD, Zhou YM, Ethiraj S, 2023 \emph{Platform governance in the presence of within-complementor interdependencies: evidence from the rideshare industry}. \emph{Management Science} .

\bibitem[{Covert \protect\BIBand{} Sweeney(2023)}]{covert2023relinquishing}
Covert TR, Sweeney RL, 2023 \emph{Relinquishing riches: Auctions versus informal negotiations in texas oil and gas leasing}. \emph{American Economic Review} 113(3):628--663.

\bibitem[{Dong et~al.(2022)Dong, Luo, Xu, Yin, \protect\BIBand{} Wang}]{dong2022strategic}
Dong T, Luo Q, Xu Z, Yin Y, Wang J, 2022 \emph{Strategic driver repositioning in ride-hailing networks with dual sourcing}. \emph{Available at SSRN} .

\bibitem[{Dube et~al.(2020)Dube, Jacobs, Naidu, \protect\BIBand{} Suri}]{dube2020monopsony}
Dube A, Jacobs J, Naidu S, Suri S, 2020 \emph{Monopsony in online labor markets}. \emph{American Economic Review: Insights} 2(1):33--46.

\bibitem[{Durand \protect\BIBand{} Vaara(2009)}]{durand2009causation}
Durand R, Vaara E, 2009 \emph{Causation, counterfactuals, and competitive advantage}. \emph{Strategic Management Journal} 30(12):1245--1264.

\bibitem[{Ellickson, Kar, \protect\BIBand{} Reeder~III(2023)}]{ellickson2023estimating}
Ellickson PB, Kar W, Reeder~III JC, 2023 \emph{Estimating marketing component effects: Double machine learning from targeted digital promotions}. \emph{Marketing Science} 42(4):704--728.

\bibitem[{Erhardt et~al.(2019)Erhardt, Roy, Cooper, Sana, Chen, \protect\BIBand{} Castiglione}]{erhardt2019transportation}
Erhardt GD, Roy S, Cooper D, Sana B, Chen M, Castiglione J, 2019 \emph{Do transportation network companies decrease or increase congestion?} \emph{Science advances} 5(5):eaau2670.

\bibitem[{Fiebig \protect\BIBand{} Bartels(1996)}]{fiebig1996frisch}
Fiebig DG, Bartels R, 1996 \emph{The frisch-waugh theorem and generalized least squares}. \emph{Econometric Reviews} 15(4):431--443.

\bibitem[{Golias \protect\BIBand{} Karlaftis(2001)}]{golias2001taxi}
Golias I, Karlaftis M, 2001 \emph{The taxi market in athens, greece, and its impacts on urban traffic}. \emph{Transportation Quarterly} 55(1).

\bibitem[{Goodchild et~al.(2019)Goodchild, MacKenzie, Ranjbari, Machado, \protect\BIBand{} Chiara}]{uwstudy}
Goodchild A, MacKenzie D, Ranjbari A, Machado J, Chiara GD, 2019 \emph{Curb allocation change project}.

\bibitem[{Gordon, Moakler, \protect\BIBand{} Zettelmeyer(2023)}]{gordon2023close}
Gordon BR, Moakler R, Zettelmeyer F, 2023 \emph{Close enough? a large-scale exploration of non-experimental approaches to advertising measurement}. \emph{Marketing Science} 42(4):768--793.

\bibitem[{Greene(2003)}]{greene2003nonspherical}
Greene W, 2003 \emph{Nonspherical disturbances—the generalized regression model}. \emph{Greene WH. Econometric Analysis. 5th edition. Upper Saddle River, NJ: Prentice Hall} 191--214.

\bibitem[{Greenwood, Wattal et~al.(2017)}]{greenwood2017show}
Greenwood BN, Wattal S, et~al., 2017 \emph{Show me the way to go home: An empirical investigation of ride-sharing and alcohol related motor vehicle fatalities.} \emph{MIS Q.} 41(1):163--187.

\bibitem[{Han et~al.(2005)Han, Chin, Franzese, \protect\BIBand{} Hwang}]{han2005estimating}
Han LD, Chin SM, Franzese O, Hwang H, 2005 \emph{Estimating the impact of pickup-and delivery-related illegal parking activities on traffic}. \emph{Transportation Research Record} 1906(1):49--55.

\bibitem[{Hastie et~al.(2009)Hastie, Tibshirani, Friedman, \protect\BIBand{} Friedman}]{hastie2009elements}
Hastie T, Tibshirani R, Friedman JH, Friedman JH, 2009 \emph{The elements of statistical learning: data mining, inference, and prediction}, volume~2 (Springer).

\bibitem[{He et~al.(2016)He, Yan, Liu, \protect\BIBand{} Ma}]{he2016traffic}
He F, Yan X, Liu Y, Ma L, 2016 \emph{A traffic congestion assessment method for urban road networks based on speed performance index}. \emph{Procedia engineering} 137:425--433.

\bibitem[{Imbens \protect\BIBand{} Rubin(2015)}]{imbens2015causal}
Imbens GW, Rubin DB, 2015 \emph{Causal inference in statistics, social, and biomedical sciences} (Cambridge University Press).

\bibitem[{Iqbal(2019)}]{Iqbal2019uber}
Iqbal M, 2019 \emph{Uber revenue and usage statistics}. \urlprefix\url{https://www.businessofapps.com/data/uber-statistics/}.

\bibitem[{Jaller et~al.(2021)Jaller, Rodier, Zhang, Lin, \protect\BIBand{} Lewis}]{jaller2021fighting}
Jaller M, Rodier C, Zhang M, Lin H, Lewis K, 2021 \emph{Fighting for curb space: Parking, ride-hailing, urban freight deliveries, and other users} .

\bibitem[{Kallus(2023)}]{kallus2023treatment}
Kallus N, 2023 \emph{Treatment effect risk: Bounds and inference}. \emph{Management Science} .

\bibitem[{Kim et~al.(2020)Kim, Sharda, Zhou, \protect\BIBand{} Pendyala}]{kim2020stepwise}
Kim T, Sharda S, Zhou X, Pendyala RM, 2020 \emph{A stepwise interpretable machine learning framework using linear regression (lr) and long short-term memory (lstm): City-wide demand-side prediction of yellow taxi and for-hire vehicle (fhv) service}. \emph{Transportation Research Part C: Emerging Technologies} 120:102786.

\bibitem[{K{\"u}nzel et~al.(2019)K{\"u}nzel, Sekhon, Bickel, \protect\BIBand{} Yu}]{kunzel2019metalearners}
K{\"u}nzel SR, Sekhon JS, Bickel PJ, Yu B, 2019 \emph{Metalearners for estimating heterogeneous treatment effects using machine learning}. \emph{Proceedings of the national academy of sciences} 116(10):4156--4165.

\bibitem[{Lewis(1973)}]{lewis1973counterfactuals}
Lewis D, 1973 \emph{Counterfactuals. harvard university press. cambridge, ma}.

\bibitem[{Lewis(1986)}]{lewis1986philosophical}
Lewis D, 1986 \emph{Philosophical papers, volume ii oxford university press}.

\bibitem[{Lewis(2004)}]{lewis2004causation}
Lewis D, 2004 \emph{Causation as influence}. \emph{The Journal of Philosophy} 97(4):182--197.

\bibitem[{Liu, Ma, \protect\BIBand{} Qian(2023)}]{liu2023optimal}
Liu J, Ma W, Qian S, 2023 \emph{Optimal curbside pricing for managing ride-hailing pick-ups and drop-offs}. \emph{Transportation Research Part C: Emerging Technologies} 146:103960.

\bibitem[{Liu, Zhang, \protect\BIBand{} Yang(2021)}]{liu2021modeling}
Liu W, Zhang F, Yang H, 2021 \emph{Modeling and managing the joint equilibrium of destination and parking choices under hybrid supply of curbside and shared parking}. \emph{Transportation Research Part C: Emerging Technologies} 130:103301.

\bibitem[{Lovell(2008)}]{lovell2008simple}
Lovell MC, 2008 \emph{A simple proof of the fwl theorem}. \emph{The Journal of Economic Education} 39(1):88--91.

\bibitem[{Lu(2019)}]{Ryland2019curb}
Lu R, 2019 \emph{Pushed from the curb: Optimizing curb space for use for ride-sourcing vehicles}. \emph{Transportation Research Board (TRB) 98th Annual Meeting} .

\bibitem[{Ma, Pi, \protect\BIBand{} Qian(2020)}]{ma2020estimating}
Ma W, Pi X, Qian S, 2020 \emph{Estimating multi-class dynamic origin-destination demand through a forward-backward algorithm on computational graphs}. \emph{Transportation Research Part C: Emerging Technologies} 119:102747.

\bibitem[{Ma \protect\BIBand{} Qian(2020)}]{ma2020measuring}
Ma W, Qian S, 2020 \emph{Measuring and reducing the disequilibrium levels of dynamic networks with ride-sourcing vehicle data}. \emph{Transportation Research Part C: Emerging Technologies} 110:222--246.

\bibitem[{Ma \protect\BIBand{} Qian(2018)}]{ma2018estimating}
Ma W, Qian ZS, 2018 \emph{Estimating multi-year 24/7 origin-destination demand using high-granular multi-source traffic data}. \emph{Transportation Research Part C: Emerging Technologies} 96:96--121.

\bibitem[{Manzoor et~al.(2023)Manzoor, Chen, Lee, \protect\BIBand{} Smith}]{manzoor2023influence}
Manzoor E, Chen GH, Lee D, Smith MD, 2023 \emph{Influence via ethos: On the persuasive power of reputation in deliberation online}. \emph{Management Science} .

\bibitem[{McCormack et~al.(2019)McCormack, Goodchild, Sheth, Hurwitz, Jashami, \protect\BIBand{} Cobb}]{mccormack2019developing}
McCormack E, Goodchild A, Sheth M, Hurwitz D, Jashami H, Cobb DP, 2019 \emph{Developing design guidelines for commercial vehicle envelopes on urban streets} .

\bibitem[{Mitman et~al.(2018)Mitman, Davis, Armet, \protect\BIBand{} Knopf}]{ITE2018Curbside}
Mitman MF, Davis S, Armet IB, Knopf E, 2018 \emph{Curbside management practitioners guide}. Technical report.

\bibitem[{Morgan \protect\BIBand{} Winship(2015)}]{morgan2015counterfactuals}
Morgan SL, Winship C, 2015 \emph{Counterfactuals and causal inference} (Cambridge University Press).

\bibitem[{Oprescu, Syrgkanis, \protect\BIBand{} Wu(2019)}]{oprescu2019orthogonal}
Oprescu M, Syrgkanis V, Wu ZS, 2019 \emph{Orthogonal random forest for causal inference}. \emph{International Conference on Machine Learning}, 4932--4941 (PMLR).

\bibitem[{Park, Tafti, \protect\BIBand{} Shmueli(2023)}]{park2023transporting}
Park S, Tafti A, Shmueli G, 2023 \emph{Transporting causal effects across populations using structural causal modeling: An illustration to work-from-home productivity}. \emph{Information Systems Research} .

\bibitem[{Pearl(2009)}]{pearl2009causality}
Pearl J, 2009 \emph{Causality} (Cambridge university press).

\bibitem[{Pearl(2019)}]{pearl2019seven}
Pearl J, 2019 \emph{The seven tools of causal inference, with reflections on machine learning}. \emph{Communications of the ACM} 62(3):54--60.

\bibitem[{Pearl \protect\BIBand{} Bareinboim(2022)}]{pearl2022external}
Pearl J, Bareinboim E, 2022 \emph{External validity: From do-calculus to transportability across populations}. \emph{Probabilistic and causal inference: The works of Judea Pearl}, 451--482.

\bibitem[{Pearl et~al.(2000)}]{pearl2000models}
Pearl J, et~al., 2000 \emph{Models, reasoning and inference}. \emph{Cambridge, UK: CambridgeUniversityPress} 19(2).

\bibitem[{Rahaim(2019)}]{SFP2019}
Rahaim J, 2019 \emph{Transportation impact analysis guidelines.} Technical report, San Francisco Planning Department.

\bibitem[{Retallack \protect\BIBand{} Ostendorf(2019)}]{retallack2019current}
Retallack AE, Ostendorf B, 2019 \emph{Current understanding of the effects of congestion on traffic accidents}. \emph{International journal of environmental research and public health} 16(18):3400.

\bibitem[{Rong, Cheng, \protect\BIBand{} Wang(2017)}]{rong2017taxi}
Rong L, Cheng H, Wang J, 2017 \emph{Taxi call prediction for online taxicab platforms}. \emph{Asia-Pacific Web (APWeb) and Web-Age Information Management (WAIM) Joint Conference on Web and Big Data}, 214--224 (Springer).

\bibitem[{Schaller et~al.(2011)Schaller, Maguire, Stein, Ng, \protect\BIBand{} Blakeley}]{schaller2011parking}
Schaller B, Maguire T, Stein D, Ng W, Blakeley M, 2011 \emph{Parking pricing and curbside management in new york city}. Technical report.

\bibitem[{Smith et~al.(2019)Smith, Wylie, Salzberg, Womeldorff, Rubendall, \protect\BIBand{} Ballus-Armet}]{smith2019data}
Smith A, Wylie A, Salzberg A, Womeldorff E, Rubendall G, Ballus-Armet I, 2019 \emph{A data driven approach to understanding and planning for curb space utility}. Technical report.

\bibitem[{Spirtes(2010)}]{spirtes2010introduction}
Spirtes P, 2010 \emph{Introduction to causal inference.} \emph{Journal of Machine Learning Research} 11(5).

\bibitem[{Tafti \protect\BIBand{} Shmueli(2020)}]{tafti2020beyond}
Tafti A, Shmueli G, 2020 \emph{Beyond overall treatment effects: Leveraging covariates in randomized experiments guided by causal structure}. \emph{Information Systems Research} 31(4):1183--1199.

\bibitem[{Tan \protect\BIBand{} Zhang(2021)}]{tan2021good}
Tan W, Zhang J, 2021 \emph{Good days, bad days: Stock market fluctuation and taxi tipping decisions}. \emph{Management Science} 67(6):3965--3984.

\bibitem[{Tirachini(2020)}]{tirachini2020ride}
Tirachini A, 2020 \emph{Ride-hailing, travel behaviour and sustainable mobility: an international review}. \emph{Transportation} 47(4):2011--2047.

\bibitem[{Wager \protect\BIBand{} Athey(2018)}]{wager2018estimation}
Wager S, Athey S, 2018 \emph{Estimation and inference of heterogeneous treatment effects using random forests}. \emph{Journal of the American Statistical Association} 113(523):1228--1242.

\bibitem[{Wijayaratna(2015)}]{wijayaratna2015impacts}
Wijayaratna S, 2015 \emph{Impacts of on-street parking on road capacity}. \emph{Australasian Transport Research Forum}, 1--15.

\bibitem[{Xu, Ghose, \protect\BIBand{} Xiao(2023)}]{xu2023mobile}
Xu Y, Ghose A, Xiao B, 2023 \emph{Mobile payment adoption: An empirical investigation of alipay}. \emph{Information Systems Research} .

\bibitem[{Xu et~al.(2021)Xu, Chen, Yin, \protect\BIBand{} Ye}]{xu2021equilibrium}
Xu Z, Chen Z, Yin Y, Ye J, 2021 \emph{Equilibrium analysis of urban traffic networks with ride-sourcing services}. \emph{Transportation science} 55(6):1260--1279.

\bibitem[{Xu, Yin, \protect\BIBand{} Zha(2017)}]{xu2017optimal}
Xu Z, Yin Y, Zha L, 2017 \emph{Optimal parking provision for ride-sourcing services}. \emph{Transportation Research Part B: Methodological} 105:559--578.

\bibitem[{Yao et~al.(2021)Yao, Chu, Li, Li, Gao, \protect\BIBand{} Zhang}]{yao2021survey}
Yao L, Chu Z, Li S, Li Y, Gao J, Zhang A, 2021 \emph{A survey on causal inference}. \emph{ACM Transactions on Knowledge Discovery from Data (TKDD)} 15(5):1--46.

\bibitem[{Yao et~al.(2019)Yao, Li, Li, Xue, Gao, \protect\BIBand{} Zhang}]{yao2019estimation}
Yao L, Li S, Li Y, Xue H, Gao J, Zhang A, 2019 \emph{On the estimation of treatment effect with text covariates}. \emph{International Joint Conference on Artificial Intelligence}.

\bibitem[{Yuan, Knoop, \protect\BIBand{} Hoogendoorn(2015)}]{yuan2015capacity}
Yuan K, Knoop VL, Hoogendoorn SP, 2015 \emph{Capacity drop: Relationship between speed in congestion and the queue discharge rate}. \emph{Transportation Research Record} 2491(1):72--80.

\bibitem[{Zalewski, Buckley, \protect\BIBand{} Weinberger(2012)}]{zalewski2012regulating}
Zalewski AJ, Buckley SM, Weinberger RR, 2012 \emph{Regulating curb space: developing a framework to understand and improve curbside management}. Technical report.

\bibitem[{Zhang et~al.(2023)Zhang, Mittal, Djavadian, Twumasi-Boakye, \protect\BIBand{} Nie}]{zhang2023ride}
Zhang K, Mittal A, Djavadian S, Twumasi-Boakye R, Nie YM, 2023 \emph{Ride-hail vehicle routing (river) as a congestion game}. \emph{Transportation Research Part B: Methodological} 177:102819.

\bibitem[{Zhang \protect\BIBand{} Nie(2022)}]{zhang2022mitigating}
Zhang K, Nie YM, 2022 \emph{Mitigating traffic congestion induced by transportation network companies: A policy analysis}. \emph{Transportation Research Part A: Policy and Practice} 159:96--118.

\bibitem[{Zhang, Li, \protect\BIBand{} Qian(2023)}]{zhang2023ridesharing}
Zhang Y, Li B, Qian S, 2023 \emph{Ridesharing and digital resilience for urban anomalies: Evidence from the new york city taxi market}. \emph{Information Systems Research} .

\end{thebibliography}


%
%
%
\clearpage
\setcounter{page}{1}

\begin{APPENDICES}

\section{Nomenclature}
\label{sec:notations}

The list of notations used in this paper is shown in Table ~\ref{tb:dsmldf}.

\begin{longtable}{m{1.5cm} m{13cm}}
\caption{List of notations.}
\label{tb:dsmldf}
\endfirsthead
\endhead

    \hline
    Notations & Description \\
    \hline
    \multicolumn{2}{c}{\textbf{Regions Related Variables}}\\
    \hline
    \multicolumn{2}{c}{}\\

    $\mathcal{V}$ & Set of regions.\\
    
    $v, n$ & An index of a region in $\mathcal{V}$. \\
    
    $\mathcal{R}$ & Set of origin regions. \\
    
    $r$ & An index of an origin region in $\mathcal{R}$. \\
    
    $\mathcal{S}$ & Set of destination regions.\\
    
    $s$ & An index of a destination region in $\mathcal{S}$. \\
    
    $\mathcal{N}(s)$ & Set of neighboring regions for region $s$.\\
	
    \multicolumn{2}{c}{}\\

    \hline
    \multicolumn{2}{c}{\textbf{Observed Variables}} \\
    \hline
    \multicolumn{2}{c}{}\\

	$y_{v}^t$ & Traffic speed in the region $v$ in the time interval $t$.\\
	
	$\mathbf{Y}_v^{t-I:t-1}$ & Vector of speed during the time intervals $t-I, \cdots, t-1$ in the region $v$. $I$ is a constant that determines the length of historical data. \\

    $\mathbf{Y}_{\mathcal{N}(v)}^{t-I:t-1}$ & Vector of average speed of all regions $n \in \mathcal{N}(v)$ during the time intervals $t-I, \cdots, t-1$. \\

    $d_{v}^t$ & NoPUDO in the region $v$ in the time interval $t$. \\

    $\mathbf{D}_v^{t-I:t-1}$ & Vector of the NoPUDO in region $v$ during the historical time intervals $t-I ... t-1$. \\

    $\mathbf{W}_{v}^t$ & External control variables in region $v$ in the time interval $t$. \\

    $\theta_v$ & Congestion effect of PUDOs in region $v$. One additional PUDO will make speed $y_v^t$ increase by $\theta_v$ in region $v$. \\
	
    \multicolumn{2}{c}{}\\
    \hline

	\multicolumn{2}{c}{\textbf{Functions and Residuals of DSML}}\\
    \hline
    \multicolumn{2}{c}{}\\

	$\varphi_v$ & A function used to predict $y_v^t$ without consideration of the congestion effect of PUDOs. \\
	
	$e_v^t$ & The residual of $\varphi_v$ and $\theta_v d_v^t$ when predicting $y_v^t$.\\
	
	
	$\psi_v$ & A function used to predict $d_v^t$. \\
	
	$\xi_v^t$ & The residual of $\psi_v$ when predicting $d_v^t$. \\
	
	\multicolumn{2}{c}{}\\
    \hline

	\multicolumn{2}{c}{\textbf{Estimated Variables}}\\
    \hline
    \multicolumn{2}{c}{}\\
    $\hat{\varphi}_v$ & Model \texttt{Y}.\\
    
    $\hat{\psi}_v$ & Model \texttt{D}.\\
    
	$\hat{y}_{v}^t$ & Prediction of the speed $y_{v}^t$ in region $v$ and time interval $t$.\\

    $\hat{d}_{v}^t$ & Prediction of  the NoPUDO $d_{v}^t$ in  region $v$ and time interval $t$. \\
    
    $\hat{e}_v^t$ & Estimation of the residual of the linear regression in Equation~\eqref{eq:lr}.\\
    
    $\hat{\epsilon}_v^t$ & Estimation of the residual of Model $\mathtt{Y}$, which is obtained by subtracting the prediction $\hat{y_{v}^t}$ and the true value of $y_{v}^t$.\\
    
    $\hat{\xi}_v^t$ & Estimation of the residual of Model $\mathtt{D}$, which is obtained by the subtracting the prediction $\hat{d}_v^t$ and the true value of $d_v^t$. \\

    $\hat{\theta}_v$ & Estimation of $\theta_v$. \\
    
    \multicolumn{2}{c}{}\\
    \hline

	\multicolumn{2}{c}{\textbf{Network Flow Related Variables}}\\
    \hline
    \multicolumn{2}{c}{}\\
    
	$q_{rs}^{t}$ & Total traffic flow from region $r$ to region $s$ before re-routing in the time interval $t$.\\
    
    $\tilde{f}_{rs}^{t}$ & Traffic flow that stays on the original routes from origin region $r$ to destination region $s$ in the time interval $t$.\\
    
    $\tilde{h}_{rsn}^{t} $ & Traffic flow that departs from region $r$ to one temporary destination $n$, $n \in \mathcal{N}(s)$ by vehicles, and from $n$ to the final destination $s$ by walking.\\
    
    \multicolumn{2}{c}{}\\
    \hline
	\multicolumn{2}{c}{\textbf{Iterated Variables in Re-routing}}\\
    \hline
    \multicolumn{2}{c}{}\\

    $\tilde{d}_s^t$ & The number of drop-off in region $s$ after re-routing in the time interval $t$. \\
    
    $\Delta_s^t$ & The change of the NoPUDO in region $s$ before and after re-routing in the time interval $t$. \\
    
    $\tilde{y}_s^t$ & Updated traffic speed each re-routing.\\
    
    ${m}_{rs}^t$ & Travel time from region $r$ to region $s$ before re-routing in the time interval $t$.\\
    
    $\tilde{m}_{rs}^t$ & Travel time from region $r$ to region $s$  after re-routing in the time interval $t$.\\
    
    $\tilde{c}_{rsn}^t$ & Travel time for traffic flow depart from region $r$ to region $n$ by vehicles, and from region $n$ to region $s$ by walking after re-routing in the time interval $t$.\\
    
    \multicolumn{2}{c}{}\\
    \hline
    
    \multicolumn{2}{c}{\textbf{Constant Variables}}\\
    \hline
    \multicolumn{2}{c}{}\\
    $k$ & Average walking speed.\\
    
    ${u}_{ns}$ & Walking time cost from region $n$ to region $s$.\\
    
    ${\mathcal{L}_{rs}}$ & Set of regions in the shortest path from origin $r$ to destination $s$, indexed by $v$.\\
    
    $l_v$ & The average travel distance in region $v$. \\
    
    \multicolumn{2}{c}{}\\
    \hline
    
\end{longtable}

\section{Conditional Average Treatment Effects of DSML (CATE)}
\label{subappendix:CATE}

To capture and estimate the nonlinear congestion effect of NoPUDO under different traffic conditions, we design and develop the conditional average treatment effects estimation in the DSML method. We extend the DSML method by adding conditional variables to obtain the dynamic congestion effect estimation under different traffic conditions. The CATE affords us to examine and explore the nonlinear relationship between the NoPUDO and traffic speed given different traffic conditions over time. We propose the Assumption~\ref{ap:nonlinear2} for estimating the CATE of the DSML method.

\begin{assumption}[Non-Linear effects]
\label{ap:nonlinear2}
For a specific region $v$, given fixed $\mathbf{Y}_v^{t-I: t-1}, \mathbf{Y}_{\mathcal{N}(v)}^{t-I:t-1}, \mathbf{W}_{v}^t$, the congestion effect $\theta_v^t$ is defined in Equation~\eqref{eq:theta_CATE}.
\begin{equation}
\label{eq:theta_CATE}
y_v^t |_{\texttt{do}(d_v^t = d_1)} - y_v^t |_{\texttt{do}(d_v^t = d_2)} = \theta_v^t\left(\mathbf{Y}_v^{t}, \mathbf{Y}_{\mathcal{N}(v)}^{t}, \mathbf{W}_{v}^t \right)  \cdot (d_1 - d_2) 
\end{equation}
where $\texttt{do}(\cdot)$ is the do-operation defined in \citet{pearl2009causality}, and $d_1$ and $d_2$ are two arbitrarily positive integers representing the NoPUDO. The $\theta_v^t\left(\mathbf{Y}_v^{t}, \mathbf{Y}_{\mathcal{N}(v)}^{t}, \mathbf{W}_{v}^t \right)$ indicates that the congestion effect of NoPUDO at time interval $t$ and region $v$ is affected by the traffic dynamics and other external attributes.
\end{assumption}

Mathematically, the CATE of the DSML method also contains three submodels: Model $\mathtt{Y}$, Model $\mathtt{D}$, and Model $\mathtt{Z}$. The first two sub-models, Model $\mathtt{Y}$ and Model $\mathtt{D}$ in the CATE are the same as that of ATE, as shown in Equation~\eqref{eq:model y} and Equation~\eqref{eq:model d}. The Model $\mathtt{Z}$ of CATE in the DSML method can refer to Equation~\eqref{eq:CATE_modelZ}. 

\begin{equation}
\label{eq:CATE_modelZ}
   \hat{\epsilon}_v^t =  \mathcal{X}^T \beta_v^t \hat{\xi}_v^t + \hat{e}_v^t
\end{equation}
where $\hat{e}_v^t$ represents the random error of the linear regression model.

First, we obtain the residual $\hat{\epsilon}_v^t$ from Model $\mathtt{Y}$ as shown in Equation~\eqref{eq:residual of model y} and residual $\hat{\xi}_v^t$ from Model $\mathtt{D}$ as shown in Equation~\eqref{eq:residual of model d}. As for the Model $\mathtt{Z}$ for the CATE, it is a linear regression between the $\hat{\epsilon}_v^t$ and $\hat{\xi}_v^t$ given the conditions $\mathcal{X}$. The $\mathcal{X}$ represents the time-varying traffic conditions. In Equation~\eqref{eq:CATE_modelZ}, the value of $\hat{\epsilon}_v^t$, $\hat{\xi}_v^t$, $\mathcal{X}$ is obtained from the Model $\mathtt{Y}$, Model $\mathtt{D}$, and $\left(\mathbf{Y}_v^{t}, \mathbf{Y}_{\mathcal{N}(v)}^{t}, \mathbf{W}_{v}^t \right)$ respectively. The Model $\mathtt{Z}$ shown in Equation~\eqref{eq:CATE_modelZ} is to learn the coefficient $\beta_v^t$. After obtaining the value of the coefficient $\beta_v^t$, we plug it into Equation~\eqref{eq:CATE_theta} to calculate the estimated $\hat{\theta}_v^t$:

\begin{equation}
\label{eq:CATE_theta}
   \hat{\theta}_v^t =  \mathcal{X}^T \beta_v^t.
\end{equation}

We further extend the numerical experiments to estimate the CATE for $\hat{\theta_v^t}$, and the estimated congestion effect is shown in Figure~\ref{fig:CATE_case}. We show the results for region 100, 186, 236, and 263, the same region example as shown in Figure~\ref{fig:residual_analysis}. The CATE value of $\hat{\theta_v^t}$ is plotted from 16:00 to 20:00 across the whole study period. The x-axis is the time point and the y-axis represents the average value of $\hat{\theta_v^t}$. The purple line represents the mean of the $\hat{\theta}_v^t$ at these time points with the deviation. The four subfigures indicate that the average value of $\hat{\theta_v^t}$ is negative, which meets our assumption and matches with the ATE results. The fluctuation also indicates the dynamic congestion effect of NoPUDO under different traffic conditions.

\begin{figure}[h]
    \centering
    \includegraphics[width=0.8\textwidth]{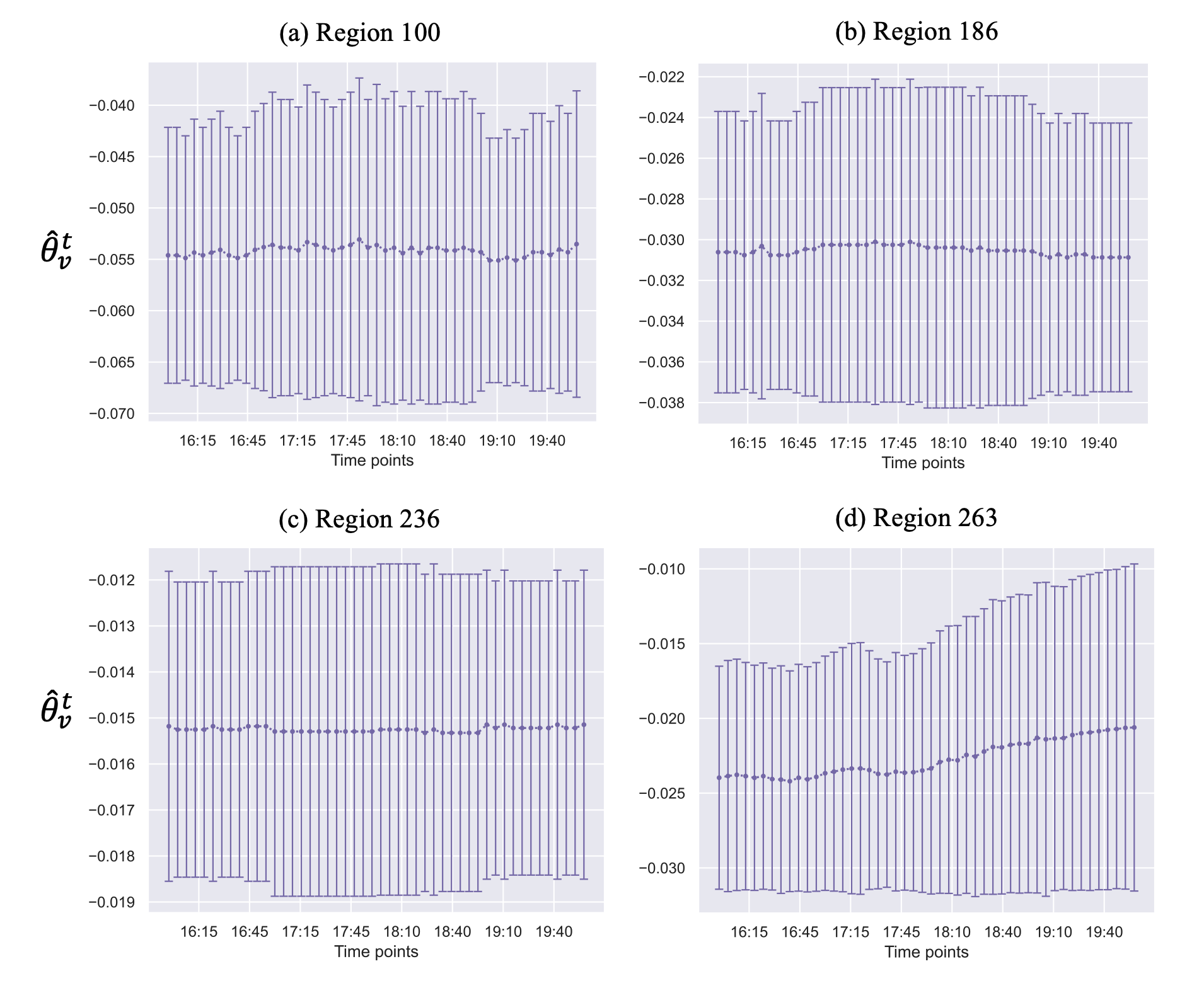}
    \caption{CATE results $\hat{\theta_v^t}$ on weekdays.}
    \label{fig:CATE_case}
\end{figure}

\section{Robustness check of the contemporaneous time interval index setting}
\label{sec:distributed_lag_model}

In the structural equation model shown in Equation~\eqref{eq:y}, we use the same time interval index to capture the congestion effect between NoPUDO and traffic speed. To provide robust justifications for the use of the contemporaneous time interval index given the coarse time granularity of the current datasets, we quantify the time series correlation between traffic speed and NoPUDO over time by a distributed lag model, as shown in Equation~\eqref{eq:dis_lag_model}~\citep{amemiya1967comparative}.

\begin{equation}
\label{eq:dis_lag_model}
   {y}^t_v = {\iota}_{0} {d}^t_v + {\iota}_{1} {d}^{t-1}_v + \cdots +{\iota}_{I} {d}^{t-I}_v +{\kappa}^t_v , \forall v,t,
\end{equation}
where 
$\kappa^t_v$ is the error term. The coefficients ${\iota}_{0}$, ${\iota}_{1}$ ... and ${\iota}_{I}$ quantify the correlation between traffic speed and NoPUDO with respect to different time intervals.

In the experimental setting, we set $I = 10$, and feed the whole dataset for different $v,t$ into the above equation to capture the overall trend of the effect of NoPUDO on traffic speed over time. We separately estimate the coefficients based on datasets of weekdays and weekends. The estimation result is shown in the Table~\ref{tb:distributed_lag_model}. First, the coefficients present a negative effect of NoPUDO on traffic speed over time. Second, the coefficient of $d_v^t$ is statistically significant and the most negative compared to those at other time intervals, such as $d_v^{t-1}$, $d_v^{t-2}$... and $d_v^{t-10}$. 
Therefore, given the current data granularity based on 5 minutes, the empirical result validates that the congestion effect requires us to make an estimation based on traffic speed and NoPUDO with the same time interval in this study.

\begin{ThreePartTable}
\begin{TableNotes}
\item Note. Standard errors are in parentheses.
\item *p $\textless$ 0.1, **p $\textless$ 0.05, ***p $\textless$ 0.01.\\
\end{TableNotes}
\begin{longtable}[c]{@{} l *{2}{l}}

\caption{Estimation result of time series correlation by the distributed lag model.} 
\label{tb:distributed_lag_model} \\
\toprule
{\bf Variables} & \multicolumn{1}{c@{}}{\bf Weekdays}&\multicolumn{1}{c@{}}{\bf Weekends}\\

\midrule
\endfirsthead

\caption*{Estimation result of time series correlation by the distributed lag model (continued)}\\
\toprule
Variables & \multicolumn{1}{c@{}}{Weekdays}&\multicolumn{1}{c@{}}{Weekends}\\
\midrule
\endhead

\bottomrule
\endfoot

\bottomrule
\insertTableNotes
\endlastfoot
NoPUDO at time interval $t$ & -0.024*** & -0.024*** \\
    & (0.001) & (0.001)\\
NoPUDO at time interval $t-1$ & -0.012*** & -0.012*** \\
    & (0.001)  & (0.001)\\
NoPUDO at time interval $t-2$ & -0.005*** & -0.006*** \\
    & (0.001)  & (0.001)\\
NoPUDO at time interval $t-3$ & -0.001* & -0.004*** \\
    & (0.001)  & (0.001)\\
NoPUDO at time interval $t-4$ & 0.001 & -0.002** \\
    & (0.001)  & (0.001)\\
NoPUDO at time interval $t-5$ & 0.002*** & -0.001 \\
    & (0.001)  & (0.001)\\
NoPUDO at time interval $t-6$ & 0.002*** & -0.001 \\
    & (0.001)  & (0.001)\\
NoPUDO at time interval $t-7$ & 0.001 & -0.002** \\
    & (0.001)  & (0.001)\\
NoPUDO at time interval $t-8$ & -0.002*** & -0.004*** \\
    & (0.001)  & (0.001)\\
NoPUDO at time interval $t-9$ & -0.007*** & -0.007*** \\
    & (0.001)  & (0.001)\\
NoPUDO at time interval $t-10$ & -0.017*** & -0.015*** \\
    & (0.001)  & (0.001)\\
Intercept &  18.622***  & 20.450***  \\
    & (0.010)  & (0.014)\\
    \hline
\end{longtable}
\end{ThreePartTable}

\section{Property of $\hat{\theta}_v$}
\label{Proof of Proposition}
In this section, we first prove Proposition~\ref{prop:linear} for the case of linear models, then Proposition~\ref{prop:nonlinear} is proved for the generalized cases. 
\subsection{Proof of Proposition~\ref{prop:linear}}
\label{ap:linear}
Based on the settings presented in Proposition~\ref{prop:linear}, we prove $\hat{\theta}_v$ is an unbiased estimator of $\theta_v$. To demonstrate the essential idea, we first use linear models for $\varphi_v$, as shown in Equation~\eqref{eq:model_FWL}. 

\begin{equation}
\label{eq:model_FWL}
{y}_{v}^{t} = \theta_v {d}_{v}^{t} + \mathbf{A}^T \mathbf{Y}_v^{t-I:t-1} + \mathbf{B}^T \mathbf{Y}_{\mathcal{N}(v)}^{t-I:t-1} + e_{v}^{t}
\end{equation}
where we assume $\mathbf{A}, \mathbf{B}, \mathbf{Y}_v^{t-I:t-1}, \mathbf{Y}_{\mathcal{N}(v)}^{t-I:t-1}$ are flattened vectors, and both $\mathbf{A}$ and $\mathbf{B}$ are parameters of $\varphi_v$.

Following the steps in DSML, we build additional regression models for $y_v^t$ and $d_v^t$, as presented in Equation~\eqref{eq:model_AUX_y} and \eqref{eq:model_AUX_d}. 

\begin{eqnarray}
{y}_{v}^{t} &=& \mathbf{A_y}^T  \mathbf{Y}_v^{t-I:t-1} + \mathbf{B_y}^T \mathbf{Y}_{\mathcal{N}(v)}^{t-I:t-1}  + \hat{\varepsilon}_{v}^{t}\label{eq:model_AUX_y}
\\
{d}_{v}^{t} &=& \mathbf{A_d}^T  \mathbf{Y}_v^{t-I:t-1} + \mathbf{B_d}^T \mathbf{Y}_{\mathcal{N}(v)}^{t-I:t-1} + \mathbf{C_d}^T \mathbf{D}_v^{t-I:t-1} + \hat{\xi}_v^t\label{eq:model_AUX_d}
\end{eqnarray}
where $\mathbf{A_y}$, $\mathbf{B_y}$, $\mathbf{A_d}$, $\mathbf{B_d}$ and $\mathbf{C_d}$ are vectors of coefficients. $(\mathbf{A_y}, \mathbf{B_y})$ are the parameters for $\hat{\varphi}_v$, and $(\mathbf{A_d}, \mathbf{B_d}, \mathbf{C_d})$ are the parameters for $\hat{\psi}_v$.

We consider an alternative least-squares regression question:

\begin{equation}
\label{eq:model_LSR_y}
\hat{\varepsilon}_{v}^{t} = \hat{\theta}_v \hat{\xi}_v^t + \hat{e}_{v}^{t}
\end{equation}

To analyze the property of $\hat{\theta}_v$, we derive $\hat{\varepsilon}_{v}^{t}$ by substituting Equation~\eqref{eq:model_FWL} into Equation~\eqref{eq:model_AUX_y}, as shown in Equation~\eqref{eq:model_FWL_proof_1}. 
\begin{equation}
\label{eq:model_FWL_proof_1}
\hat{\varepsilon}_{v}^{t} = \theta_v d_v^t + (\mathbf{A} - \mathbf{A_y})^T \mathbf{Y}_v^{t-I:t-1} + (\mathbf{B} - \mathbf{B_y})^T \mathbf{Y}_{\mathcal{N}(v)}^{t-I:t-1}  + e_v^t
\end{equation}

Then we plug the variable $d_v^t$ in the Equation~\eqref{eq:model_AUX_d} into Equation~\eqref{eq:model_FWL_proof_1}. Eventually, we can formulate the $\hat{\varepsilon}_{v}^{t}$ in the Equation~\eqref{eq:model_FWL_proof_2}.

\begin{equation}
\label{eq:model_FWL_proof_2}
\hat{\varepsilon}_{v}^{t}  = \theta_v \hat{\xi}_v^{t} + (\theta_v \mathbf{A_d} + \mathbf{A} - \mathbf{A_y})^T \mathbf{Y}_v^{t-I:t-1} + (\theta_v \mathbf{B_d} + \mathbf{B} - \mathbf{B_y})^T \mathbf{Y}_{\mathcal{N}(v)}^{t-I:t-1} + (\theta_v \mathbf{C_d})^T \mathbf{D}_v^{t-I:t-1} + e_v^t
\end{equation}

As $\hat{\varepsilon}_{v}^{t}$ is the residual from the linear regression in Equation~\eqref{eq:model_AUX_y}, it is not correlated with $\mathbf{Y}_v^{t-I:t-1}$ or $\mathbf{Y}_{\mathcal{N}(v)}^{t-I:t-1}$ given both variables are the attributes of the linear regression. Additionally, $\hat{\varepsilon}_{v}^{t}$ is not correlated with $\mathbf{D}_v^{t-I:t-1}$ due to the causal graph in Figure~\ref{fig:dsml_casual_graph}. Therefore, we have the coefficients $\theta_v \mathbf{A_d} + \mathbf{A} - \mathbf{A_y}$, $\theta_v \mathbf{B_d} + \mathbf{B} - \mathbf{B_y}$, and $\theta_v \mathbf{C_d}$ equal to zero in Equation~\eqref{eq:model_FWL_proof_2}. Consequently, we have Equation~\eqref{eq:model_FWL_result} holds.

\begin{equation}
\label{eq:model_FWL_result}
\hat{\varepsilon}_{v}^{t}  = \theta_v \hat{\xi}_v^{t} + e_v^t
\end{equation}

By comparing Equation~\eqref{eq:model_LSR_y} and Equation~\eqref{eq:model_FWL_result}, we have Equation~\eqref{eq:results} holds.
\begin{equation}
\label{eq:results}
\begin{array}{lll}
\hat{\theta}_v &=& \theta_v\\
\hat{e}_{v}^{t} &=& e_{v}^{t}
\end{array}
\end{equation}

The above proof is extended from the Frisch-Waugh-Lovell (FWL) theorem~\citep{fiebig1996frisch, lovell2008simple}, and we show $\theta_v \mathbf{C_d} = 0$ based on the specific problem setting for the causal graph in this study.

\subsection{Proof of Proposition~\ref{prop:nonlinear}}
\label{ap:nonlinear}

To prove Proposition~\ref{prop:nonlinear}, we rely on Theorem 3.1 in \citet{chernozhukov2018double}. To this end, we verify that both Assumption 3.1 and 3.2 in \citet{chernozhukov2018double} hold. 
For region $v$, we set $\eta_v = (\varphi_v, \psi_v)$, and the inputs for both functions are omitted. Then the Neyman score function can be defined in Equation~\eqref{eq:neyman}
\begin{equation}
\label{eq:neyman}
\omega(\theta_v, \eta_v) = \left(y_v^t - \theta_v d_v^t - \varphi_v\right) \left( d_v^t  - \psi_v\right)
\end{equation}

We note that $\omega(\theta_v, \eta)$ is insensitive to the small change of either $\varphi_v$ or $\theta_v$, as presented in Equation~\eqref{eq:insen}.

\begin{equation}
\label{eq:insen}
    \partial_{\eta_v} \mathbb{E}\omega(\theta_v, \eta_v)  [\eta_v - \eta_v^0] = 0
\end{equation}

Then $\omega(\theta_v, \eta)$ is Neyman orthogonal, which satisfies Assumption 3.1. Additionally, Assumption 3.2 is satisfied because Equation~\eqref{eq:conv2} holds. Given that the data splitting technique presented in Section~\ref{sec:dsmlsol} is adopted to train $\varphi_v$ and $\psi_v$ separately, then based on Theorem 3.1 in \citet{chernozhukov2018double}, Proposition~\ref{prop:nonlinear} is proved.

\section{Causality analysis of DSML}
\label{Causality analysis of DSML}

The central idea of estimating causality is to construct a counterfactual world that matches the factual world~\citep{lewis1973counterfactuals, lewis1986philosophical, lewis2004causation,
durand2009causation}, and the difference in the dependent variable between the counterfactual and factual world is the targeted causality. In general, there are two primary approaches to constructing the counterfactual world, including experimental and non-experimental approaches. RCT randomly separates participants into control and treatment groups and only conducts interventions in the treatment group. Regarding the treatment group, the status of the control group is considered as the counterfactual world. However, such controlled experiments are not always feasible for complex real-world transportation scenarios \citep{ greene2003nonspherical, angrist2009mostly, imbens2015causal, gordon2023close}. The second approach is to leverage historical data to make a prediction of the dependent variable and construct the counterfactual world based on the predicted results. The DSML method, which falls under the second approach, formulates the counterfactual world from a data-driven standpoint. In our context, the predicted numerical value of NoPUDO and traffic speed can be regarded as the results from the counterfactual world. We unfold the congestion effect estimation using the second approach due to the unavailable experiment setting and infeasible treatment manipulation. Specifically, it is challenging to conduct a field experiment by setting up the control and treatment groups due to the inevitably disruptive nature of the real traffic situation. Furthermore, the treatment in our setting is a continuous number of pick-ups and drop-offs. It is demanding to manipulate the treatment, NoPUDO, in the intricate and dynamic real-world road networks.

We consistently follow a well-established procedure of causality inference to conduct the congestion effect estimation in the study. \cite{park2023transporting} propose that causal estimation can be derived from the following steps: (1) identifying related variables; (2) clarifying causal relationships; (3) constructing causal diagrams; (4) determining the possibility of transporting historical data from other existing causal scenarios; (5) formulating structural equation based on causal diagrams; and (6) making a final causality estimation. In the same vein, we first identify the research gap through an exhaustive literature review, focusing on investigating the congestion effect of NoPUDO on traffic speed. Second, we justify and conclude the causal relationship between NoPUDO and traffic speed based on empirical evidence, as shown in Figure \ref{fig:illustration of PUDO}. Third, we extrapolate and conceptualize the causal relationship between NoPUDO and traffic speed in the causal diagram as presented in Figure \ref{fig:dsml_casual_graph}. This causal diagram integrates the transportation domain knowledge to demonstrate the intertwined temporal and spatial relationship among involved variables. Fourth, we skip the fourth step mentioned above because the congestion effect has not been extensively investigated in pioneering literature and there are few existing similar causal scenarios. Fifth, we mathematically formulate the causal graph and prove rigorously these assumptions and equations in Section~\ref{sec:structural equation models}. Last, we conduct the congestion estimation using the observational data of NoPUDO and traffic speed in Section~\ref{sec:numberical experiment}.

The DSML method can guarantee that our estimates are of causality rather than correlation by dual channels: causal graph drawing and mathematical formulas validation. In this study, we first utilize the causal graph to delineate the causal relationship among the variables and apply the machine learning method to estimate the congestion effect based on the above causal graph. The causality analysis of the DSML method first needs to trace back to the well-built causal graph, as shown in Figure \ref{fig:dsml_casual_graph}. An essential principle for ensuring that the estimation represents causality rather than correlation is that we should eliminate the additional influence of confounders on the estimation. The causal graph conveys the causal assumption and empowers us to identify and control confounding variables in the estimation \citep{manzoor2023influence}.

Specifically, in Model $\mathtt{Y}$, the confounders are the historical traffic speed record at the current region $\mathbf{Y}_v^{t-I: t-1}$, the historical traffic speed record at the neighboring regions $\mathbf{Y}_{\mathcal{N}(v)}^{t-I:t-1}$, and external control variables $\mathbf{W}_v^t$. In a similar vein, in the Model $\mathtt{D}$, the confounders are the historical NoPUDO in the current region $\mathbf{D}_v^{t-I: t-1}$, the historical traffic speed record in the current region $\mathbf{Y}_v^{t-I: t-1}$, the historical traffic speed record in the neighboring region $\mathbf{Y}_{\mathcal{N}(v)}^{t-I:t-1}$, and external control variables $\mathbf{W}_{v}^t$. Essentially, the central idea is to capture and model the trend of the time series, and hence the historical records of speed and NoPUDO are the confounders. Furthermore, the congestion effect is estimated separately and independently for each region, which also has controlled implicit confounders induced by the region heterogeneity. Comprehensively controlling confounders in the DSML method can effectively ensure our estimates of causality.

Besides, from the perspective of structural causal modeling, we have rigorously proved that if the above Equations~\eqref{eq:y},~\eqref{eq:d},~\eqref{eq:noise1},~\eqref{eq:noise2},~\eqref{eq:noise12} and~\eqref{eq:noise22} hold, it can guarantee that we can accurately and unbiasedly estimate the causal effects of NoPUDO on traffic speed. The machine learning-based DSML method is different from the traditional econometric methods, such as Difference-in-Difference (DID) which mainly emphasizes the difference in the value of the dependent variable before and after adding treatment. In the DSML method, we first formulate and capture the correlation between the traffic speed and NoPUDO in Equation~\eqref{eq:y}. To distill the pure causality, we included and formulated the confounders in Model $\mathtt{Y}$ and Model $\mathtt{D}$, as shown in Equation~\eqref{eq:model y} and~\eqref{eq:model d}. The non-linear relationship between the confounding variables and traffic speed/NoPUDO is obtained by parameter learning based on machine learning methods. The residuals represent the remaining parts of traffic speed and NoPUDO after excluding the disturbance of confounders \citep{xu2023mobile}. In the Model $\mathtt{Z}$, the regression between the above two residuals captures and quantifies the causality. The double-layer design of the DSML method can elegantly address the issue of confounding correlation between independent variables and covariates to obtain the causality, which remains challenging in one single regression as shown in Equation~\eqref{eq:y}~\citep{dube2020monopsony}.

The DSML method mainly differs from the traditional econometric models in partially linear assumption and means of parameters learning. First, traditional econometric models, such as ordinary least squares (OLS), assume everything linear, while the DSML method emphasizes the partially linear assumption, and models the non-linear relationship between covariates and the independent variable by machine learning models~\citep{chernozhukov2018double}. The partially linear assumption is more adaptive and applicable to real-world scenarios and can cater to the emerging complex estimation demands. Second, the DSML method adopts machine learning methods to formulate the non-linear relationship in Model $\mathtt{Y}$ and $\mathtt{D}$, which delve deep into data and enable greater flexibility in estimation. In the traditional econometric methods, there are pre-defined mathematical formulas defined before we feed the data set to estimate the targeted coefficient. In the DSML method, two functions, $\hat{\varphi}_v$ and $\hat{\psi}_v$, are fully trained based on the data, and the parameters are learned by the machine learning models. The DSML method guarantees an unbiased estimation by implementing Neyman-orthogonal scores and cross-validation to overcome the biases caused by using machine learning methods. We will further discuss the implementation details of the DSML method in Section~\ref{sec:dsmlsol}.

To have an intuitive understanding of the causality in the DSML method, we use an example of a factual world and a counterfactual world to unfold and illustrate the underlying mechanism~\citep{morgan2015counterfactuals}. Specifically, the treatment in this study is the amount of the PUDO. The factual world is that we increase or decrease the number of PUDO, and mainly then observe the change of the actual traffic speed. The counterfactual world is if the number of PUDO did not increase or decrease, we can observe the change in the actual traffic speed accordingly. An example of the factual and counterfactual world is shown in Table~\ref{tb:counterfactual world}. Given a region, we identify the world at the time interval $t-1$ as the time period before applying the treatment. The world at the time interval $t$ is regarded as the time period after receiving the treatment. As shown in Table~\ref{tb:counterfactual world}, the NoPUDO increases by 5. The additional 5 units of PUDO is the treatment, and the actual treatment is continuous. As for the factual world in the time interval $t-1$ and $t$, we can observe the difference in the traffic speed in the current region. As for the counterfactual world, there is no difference in NoPUDO for both the before and after time periods. Similarly, we can observe the numerical value of the traffic speed. The difference in traffic speed in the time interval $t-1$ and $t$ is due to the change in the time series. In this case, the targeted causality is the difference between two differences: one is the disparity of traffic speed at the ``Before" and ``After" time periods in the factual world; another is the difference of traffic speed at the ``Before" and ``After" time periods in the counterfactual world.

\begin{ThreePartTable}
\begin{longtable}[c]{@{} l *{4}{l} @{}}

\caption{An example of factual and counterfactual world } \label{tb:counterfactual world} \\
\toprule
World (Group) & \multicolumn{2}{c@{}}{Before: at time interval $t-1$}&\multicolumn{2}{c@{}}{After: at time interval $t$}\\
\cmidrule(l){2-3}
\cmidrule(l){4-5}
& Traffic speed (mph) & NoPUDO & Traffic speed (mph) & NoPUDO \\

\midrule
\endfirsthead

\caption*{An example of factual and counterfactual world (continued)}\\
\toprule
World (Group) & \multicolumn{2}{c@{}}{Before: at time interval $t-1$}&\multicolumn{2}{c@{}}{After: at time interval $t$}\\
\cmidrule(l){2-3}
\cmidrule(l){4-5}
& Traffic speed (mph) & NoPUDO & Traffic speed (mph) & NoPUDO \\
\midrule
\endhead

\bottomrule

\endlastfoot

        Factual world (Treated) &           15 &    25 &           14 &    30 \\
        Counterfactual world (Controlled) &      15 &    25 &      14.5   &    25 \\

    \hline
\end{longtable}
\end{ThreePartTable}

\section{Proof of Proposition~\ref{prop:decompose}}
\label{sec:proofde}
The total travel time (TTT) before the re-routing can be calculated as $\sum_{r,s \in \mathcal{R}}q_{rs}^t  m_{rs}^t$, and the TTT after re-routing is represented as the objective function in Formulation~\eqref{eq:rerouting}. Therefore, the change of TTT ($\Delta TTT $) can be written in Equation~\eqref{eq:decom}.
\begin{equation}
\label{eq:decom}
\begin{array}{cllllll}
\Delta TTT &=& \left( \sum_{r \in \mathcal{R}} \sum_{s \in \mathcal{R}} \tilde{f}_{rs}^t  \tilde{m}_{rs}^t + \sum_{r,s \in \mathcal{R}} \sum_{n \in \mathcal{N}(s)} \tilde{h}_{rsn}^t  \tilde{c}_{rsn}^t \right) - \sum_{r,s \in \mathcal{R}}q_{rs}^t m_{rs}^t \\
&=&\sum_{r,s \in \mathcal{R}}  \left(   \tilde{f}_{rs}^t \tilde{m}_{rs}^t  + \sum_{n \in \mathcal{N}(s)} \tilde{h}_{rsn}^t  \tilde{c}_{rsn}^t - q_{rs}^t m_{rs}^t \right)  \\
&=&\sum_{r,s \in \mathcal{R}}  \left(  \tilde{f}_{rs}^t m_{rs}^t -  \tilde{f}_{rs}^t m_{rs}^t- \tilde{f}_{rs}^t \tilde{m}_{rs}^t +\sum_{n \in \mathcal{N}(s)} \tilde{h}_{rsn}^t  \tilde{c}_{rsn}^t  - q_{rs}^t m_{rs}^t\right) \\
&=&\sum_{r,s \in \mathcal{R}}  \left(   \tilde{f}_{rs}^t m_{rs}^t -  \tilde{f}_{rs}^t m_{rs}^t+ \tilde{f}_{rs}^t \tilde{m}_{rs}^t +  \sum_{n \in \mathcal{N}(s)} \tilde{h}_{rsn}^t  \left( \tilde{m}_{rn}^t + w_{ns} \right) - q_{rs}^t m_{rs}^t \right) \\
&=&\sum_{r,s \in \mathcal{R}} \left(   \tilde{f}_{rs}^t m_{rs}^t -  \tilde{f}_{rs}^t m_{rs}^t+ \tilde{f}_{rs}^t \tilde{m}_{rs}^t - q_{rs}^t m_{rs}^t\right)  \\
&&+ \sum_{r,s \in \mathcal{R}} \sum_{n \in \mathcal{N}(s)} \tilde{h}_{rsn}^t  \left( \tilde{m}_{rn}^t - m_{rn}^t + m_{rn}^t + w_{ns} - m_{ns}^t + m_{ns}^t \right) \\
&=&\sum_{r,s \in \mathcal{R}} \left(   \tilde{f}_{rs}^t m_{rs}^t + \sum_{n \in \mathcal{N}(s)}\tilde{h}_{rsn}^t \left(m_{rn}^t  + m_{ns}^t \right) - q_{rs}^t m_{rs}^t\right) + \sum_{r,s \in \mathcal{R}} \tilde{f}_{rs}^t\left(   \tilde{m}_{rs}^t - m_{rs}^t   \right) \\
&&+ \sum_{r,s \in \mathcal{R}} \sum_{n \in \mathcal{N}(s)} \tilde{h}_{rsn}^t  \left( \tilde{m}_{rn}^t - m_{rn}^t  + w_{ns} - m_{ns}^t \right) \\
&=&\sum_{r,s \in \mathcal{R}}  \left(   \tilde{f}_{rs}^t m_{rs}^t + \sum_{n \in \mathcal{N}(s)} \tilde{h}_{rsn}^t\left(m_{rn}^t  + m_{ns}^t \right) - q_{rs}^t m_{rs}^t\right)+ \sum_{r,s \in \mathcal{R}} \tilde{f}_{rs}^t\left(   \tilde{m}_{rs}^t -m_{rs}^t\right)    \\
&&+ \sum_{r,s \in \mathcal{R}}   \sum_{n \in \mathcal{N}(s)} \tilde{h}_{rsn}^t \left(\tilde{m}_{rn}^t  - m_{rn}^t + w_{ns} - m_{ns}^t \right)   \\
&=& \Delta_\text{Counterfactual} + \Delta_\text{PUDO, Remain} + \Delta_\text{PUDO, Detour} 
\end{array}
\end{equation}
The above decomposition completes the proof of Proposition~\ref{prop:decompose}.

\section{Sensitivity analysis regarding the selection of ML models}
\label{Sensitivity analysis regarding the selection of ML models}

We examine the robustness of different ML models used in Model $\mathtt{Y}$ and Model $\mathtt{D}$. In Algorithm~\ref{alg:DSML programming}, the optimal ML model is selected from Gradient Boosting, Random Forest, and Ada Boosting Regression using cross-validation. In this section, we specify the ML model used in Model $\mathtt{Y}$ and Model $\mathtt{D}$ and evaluate how the estimation results are different from the original ones. In general, we believe a smaller difference indicates a more robust DSML method in terms of the choice of ML models.

To this end, we run the DSML method by fixing Model $\mathtt{Y}$ and Model $\mathtt{D}$ to be either Gradient Boosting Regression, or Random Forest Regression, or Ada Boosting Regression. Then we compare the difference between the newly estimated and the original $\hat{\theta}_v$ through Pearson correlation coefficients, and the results are presented in Table~\ref{tb:CRL_DSML}. One can see that the correlation coefficients for Gradient Boosting, Random Forest, and Ada Boosting Regression are 0.99, 0.94, and 0.83, respectively. All the correlation coefficients are high, indicating that the proposed DSML method is robust to the choice of the ML models for Model $\mathtt{Y}$ and Model $\mathtt{D}$.

\begin{table*}
\begin{floatrow}
\capbtabbox
{
    \begin{tabular}{c c c c }
    \toprule
    ML models & GB & RF & Ada \\
    \hline
    Correlation coefficient & 0.99  & 0.94 & 0.83 \\
    \hline
    \end{tabular}
}
{
\caption{Sensitivity analysis of ML models used in the DSML method.}
\label{tb:CRL_DSML}
}
\capbtabbox
{
    \begin{tabular}{l l l }
    \toprule
    Models & DML & LR \\
    \hline
    Correlation coefficient & -0.14  & 0.27 \\
    \hline
    \end{tabular}
}
{
\caption{Correlation analysis for DSML vs. DML and DSML vs. LR.}
\label{tb:CRL_comparison}
}
\end{floatrow}
\end{table*}

\section{Sensitivity analysis regarding the region size}
\label{sec:sen_ana_region_size}

In this section, we conduct the sensitivity analysis to examine the effect of region size on the estimation result in the DSML method. The definition of the region in this paper follows the NYC Department of City Planning's Neighboring Tabulation Areas. The above estimation result is analyzed and obtained with a single region as a unit. How the region size affects the estimation result in the DSML method is not well studied and examined. Conducting the sensitivity analysis of region size lays a solid foundation for us to extend and promote the DSML method into other scenarios and enhance its generalization. In this sensitivity analysis, we first merge the neighboring regions into an aggregate region, which has a larger zone size than a single region. Subsequently, the DSML method can be re-applied to the aggregate region to estimate the congestion effect of NoPUDO for the entire aggregate region. 

We take regions 48, 68, and 100 as a case study to examine how the region size affects the estimation result. Region 100, belonging to the Midtown, has been studied in the residual analysis in Figure~\ref{fig:residual_analysis}. As shown in Figure~\ref{fig:map}, region 48 belongs to the Clinton area, region 68 is within the Hudson Yards area, and both regions are adjacent to region 100. We run the DSML method by viewing the two or three regions as a whole region with a large size. The estimation result of these aggregate regions can justify the effectiveness of the DSML method in different region sizes.

As for the regions 48, 68, and 100, there are four types of combinations, denoted as case 1 (aggregating regions 48 and 68), case 2 (aggregating regions 48 and 100), case 3 (aggregating regions 68 and 100), and case 4 (aggregating regions 48, 68, and 100). The estimation result of the above four cases on weekdays and weekends is shown in Table~\ref{tb:Case 1 Estimation Results of Aggregated Regions by DSML}. As for the four cases, the numerical value of the estimated congestion effect on the aggregated regions is within the range of that of its composing regions. The estimated effects of the aggregate region are generally the average of the effects of those disaggregated regions. The variation in the estimation result of the above cases indicates that the region size is a critical factor affecting the estimation result.

The effect of the region size on the congestion effect estimation can be disentangled from the perspective of the underlying mathematical mechanism and calculation principle. In this paper, we adopt the NYC Taxi zone system to divide regions in Manhattan and conduct the congestion effect estimation based on the region level. Specifically, when we preprocess data to obtain the average traffic speed, $y_v^t$ and NoPUDO $d_v^t$, we firstly fence each region following the NYC Taxi zone system and calculate the average traffic speed and count the total number of PUDO in the current region. The numerical value of these variables will be used to estimate the congestion effect at the region level in the DSML method. If different analysis units are used, such as the street or city level, the data generation for the above variables would also be different. The estimation result on the smaller analysis unit would be more precise due to finer-grained data collection and effectively controlled variables regarding each smaller area, but the fine-grained estimation might suffer the issue of limited trip record data. An intuitive example is that trip records are not evenly distributed in the streets. There are few trip records at the street level during non-peak hours. 

Generally, the sensitivity analysis regarding the region size demonstrates that the zone scale affects the congestion effect estimation. Theoretically, the DSML method can be adaptive to estimating the congestion effect for different analysis units, including the street level, the region level, the city level, and so on. Compared to the coarse-grained level, the fine-grained unit requires running the DSML method based on the data at a more disaggregate level. The estimation based on the fine-grained and small area implicitly controls other covariates, which guarantees a more precise estimation result. However the limited data raises the issue of data lacking for small areas. Trading off between analysis granularity and data size limitation should depend on specific research questions to be solved. Specifically, in our study, we aim at designing and taking the region-based re-routing strategy to mitigate the congestion effect, and hence the estimation of the congestion effect of NoPUDO should also be conducted based on the region level.

\begin{ThreePartTable}
\begin{TableNotes}
\item Note. Standard errors are in parentheses.
\item *p $\textless$ 0.1, **p $\textless$ 0.05, ***p $\textless$ 0.01.\\
\end{TableNotes}
\begin{longtable}[c]{@{} l *{2}{l}}

\caption{Estimation result of aggregate regions by DSML} 
\label{tb:Case 1 Estimation Results of Aggregated Regions by DSML} \\
\toprule
Regions ID & \multicolumn{1}{c@{}}{Weekdays}&\multicolumn{1}{c@{}}{Weekends}\\
& $\theta$ & $\theta$ \\

\midrule
\endfirsthead

\caption*{Estimation result by DSML, DML and LR (continued)}\\
\toprule
Region ID & \multicolumn{1}{c@{}}{Weekdays}&\multicolumn{1}{c@{}}{Weekends}\\
& $\theta$ & $\theta$ \\
\midrule
\endhead

\bottomrule
\endfoot

\bottomrule
\insertTableNotes
\endlastfoot

48 & -0.040*** & -0.036*** \\
    & (0.001) & (0.001)\\
68 & -0.025*** & -0.021*** \\
    & (0.001)  & (0.001)\\
100 & -0.075*** & -0.090*** \\
    & (0.001)  & (0.001)\\

Case 1 (48 \& 68) & -0.035*** & -0.032*** \\
    & (0.001)  & (0.001)\\

Case 2 (48 \& 100) & -0.052*** & -0.057*** \\
    & (0.001)  & (0.001)\\

Case 3 (68 \& 100) & -0.048*** & -0.050*** \\
    & (0.001)  & (0.001)\\

Case 4 (48 \& 68 \& 100) & -0.043*** & -0.051*** \\
    & (0.001)  & (0.001)\\
    \hline
\end{longtable}
\end{ThreePartTable}

\section{Comparison among DSML, DML, and LR}
\label{Comparison among DSML, DML, and LR}

We compare the developed DSML method with the standard DML and classical LR methods in terms of involved features, outcome variables, and methods, as shown in Table~\ref{tb:differences_models}. Both DSML and DML methods consist of three sub-models: the first and second sub-models use machine learning models to predict $y_v^t$ and $d_v^t$ separately and the third sub-model runs a linear regression between the above residuals from the first two sub-models to estimate the congestion effect. One of the biggest differences between DSML and DML methods lies in the involved features. Specifically, Model $\mathtt{Y}$ does not include $\mathbf{D}_v^{t-I:t-1}$ to predict traffic speed $y_v^t$.

The estimation result of $\hat{\theta}_v$ using the DML method on weekdays is shown in Figure~\ref{fig: Estimation_comparison} (a). The estimation results on weekends are similar and hence omitted. On average, $\hat{\theta}_v$ is $-0.008$ estimated by DML. The absolute numerical value of the estimated $\hat{\theta}_v$ by DML is generally smaller than that by DSML, and $\hat{\theta}_v$ for all regions is almost identically small. From the spatial distribution of the estimated results in Figure~\ref{fig: Estimation_comparison} (a), it is obvious that the DML method cannot accurately capture the heterogeneity of the congestion effects among different regions. The failure in this estimation of the DML method is attributed to that it additionally considers the non-existing relationship from $\mathbf{D}_{v}^{t-I:t-1}$ to $y_v^t$ based on the causal graph in Figure~\ref{fig:dsml_casual_graph}.

We also compare the LR and DSML methods from multiple aspects of the theoretical underpinnings and estimation results. LR emphasizes modeling the linear relationship between the dependent variable $y_v^t$ and independent variable $d_v^t$ directly in the linear regression, and the DSML method first focuses on using advanced machine learning methods to make predictions separately and obtain residuals from the first two sub-models before the final linear regression. The adoption of machine learning methods in the DSML method significantly enhances flexibility and improves the effectiveness of the estimation, but it might also induce biased estimation because of regularization and overfitting \citep{chernozhukov2018double,gordon2023close}. As we present in Section \ref{sec:dsmlsol} regarding the underlying solution algorithm, the DSML method overcomes the bias by using Neyman-orthogonal moments/scores in the regularization and eliminating the bias from overfitting through implementing the cross-validation. 


\begin{table}
\caption{Different features and outcome variables of DSML, DML, and LR.}
\label{tb:differences_models}
\begin{tabular}{l l l l }
\toprule
Models & Features & Outcome Variable  & Methods \\ [1.2ex] 
\hline
\multirow{3}{*}{DSML} & $\mathbf{Y}_{v}^{t-I:t-1}$, $\mathbf{Y}_{\mathcal{N}(v)}^{t-I:t-1}$, $\mathbf{W}_{v}^t$ & $y_{v}^{t}$  & ML models \\
& $\mathbf{D}_v^{t-I:t-1}$, $\mathbf{Y}_v^{t-I:t-1}$, $\mathbf{Y}_{\mathcal{N}(v)}^{t-I:t-1}$, $\mathbf{W}_{v}^t$ & $d_{v}^t$  & ML models \\

&$\hat{\xi}_{v}^t$ & $\hat{\epsilon}_{v}^t$ & linear regression \\

\hline
\multirow{3}{*}{DML} & $\mathbf{D}_{v}^{t-I:t-1}$, $\mathbf{Y}_{v}^{t-I:t-1}$, $\mathbf{Y}_{\mathcal{N}(v)}^{t-I:t-1}$, $\mathbf{W}_{v}^t$ & $y_{v}^t$ & ML models \\
& $\mathbf{D}_v^{t-I:t-1}$, $\mathbf{Y}_v^{t-I:t-1}$, $\mathbf{Y}_{\mathcal{N}(v)}^{t-I:t-1}$, $\mathbf{W}_v^t$ & $d_v^t$ & ML models \\
& $\xi_{v}^t$ & $\epsilon_{v}^t$ & linear regression \\
\hline
LR & $d_{v}^t$  & $y_{v}^t$ & linear regression \\
\bottomrule
\end{tabular}
\end{table}



Traditional LR, while appealing for its formulation understandability and coefficient interpretability, will face challenges when addressing the high-dimensional variables and separating the correlation between the covariates and independent variables. First, intricate transportation scenarios require us to consider and include many control variables representing the intertwined spatial-temporal relationship in the congestion effect estimation. It triggers highly complex computations to estimate the coefficient in the presence of high-dimensional nuisance parameters. The actual computation violates the traditional assumption (e.g. Donsker properties) that limits the complexity of the parameter space in the classical semi-parametric setting \citep{chernozhukov2018double}. Secondly, the intuition of using two separate structural equations in the DSML method is to address the endogeneity issues raised by the correlation between the covariates $\mathbf{Y}_{v}^{t-I:t-1}$, $\mathbf{Y}_{\mathcal{N}(v)}^{t-I:t-1}$ and the independent variable $d_v^t$ \citep{xu2023mobile}. One contained in $\mathbf{Y}_{v}^{t-I:t-1}$ or $\mathbf{Y}_{\mathcal{N}(v)}^{t-I:t-1}$ could be correlated with the independent variable $d_v^t$ in Equation \eqref{eq:y}. The residuals of Model $\mathtt{Y}$ and Model $\mathtt{D}$ are the remaining parts that have not been explained by the covariables. Leveraging the residuals to run the final regression can mitigate the disturbance from covariates, resulting in extracting and distilling the pure causal effect.

The estimation result of $\hat{\theta}_v$ using the LR method on weekdays is shown in Figure~\ref{fig: Estimation_comparison} (b). The average numerical value of $\hat{\theta}_v$ estimated by LR is $-0.055$. Based on the spatial distribution of the estimated congestion effect, LR overlooks the intertwined spatio-temporal relationship between $y_v^t$ and $d_v^t$, and the estimated $\hat{\theta}_v$ is smaller (the absolute value is larger) than that estimated from DSML, which is consistent with Example~\ref{ex:over}. In conclusion, compared to the DSML method, LR is appealing for its explainability due to its understandable operation mechanism. However, it cannot be applied in the high-dimensional parameters due to the existing inherent assumption. A conflict arises between the practical requirements to address practical issues and the constraints of the traditional LR method.

To further distinguish the difference in estimation results obtained from DSML, DML, and LR, we conduct the correlation analysis and t-test to delve deeper into our comparison and analysis. First, the correlation coefficients between DML and DSML, LR and DSML are shown in Table~\ref{tb:CRL_comparison}. The correlation coefficients can offer an overview regarding whether the spatial distribution of the estimated congestion effect is correlated. The small numerical value of the coefficients in Table~\ref{tb:CRL_comparison} indicates that the estimated $\hat{\theta}_v$ by DSML is completely different from that estimated by DML or LR. 
Additionally, we also conduct the t-test and compare the significance level of the estimated coefficients,
and the results show that some $\hat{\theta}_v$ estimated by the DML method are not statistically significant, which might be due to the superfluous consideration of the confounding factors $\mathbf{D}_{v}^{t-I:t-1}$ in Model $\mathtt{Y}$. 
Although the significance levels of the estimated coefficient from LR are almost statistically significant, the estimated $\hat{\theta}_v$ not only contains the causality but also includes a correlation between NoPUDO and traffic speed, based on our discussion in Example~\ref{ex:over}.


\begin{figure}[H]
    \centering
    \includegraphics[width=0.8\linewidth]{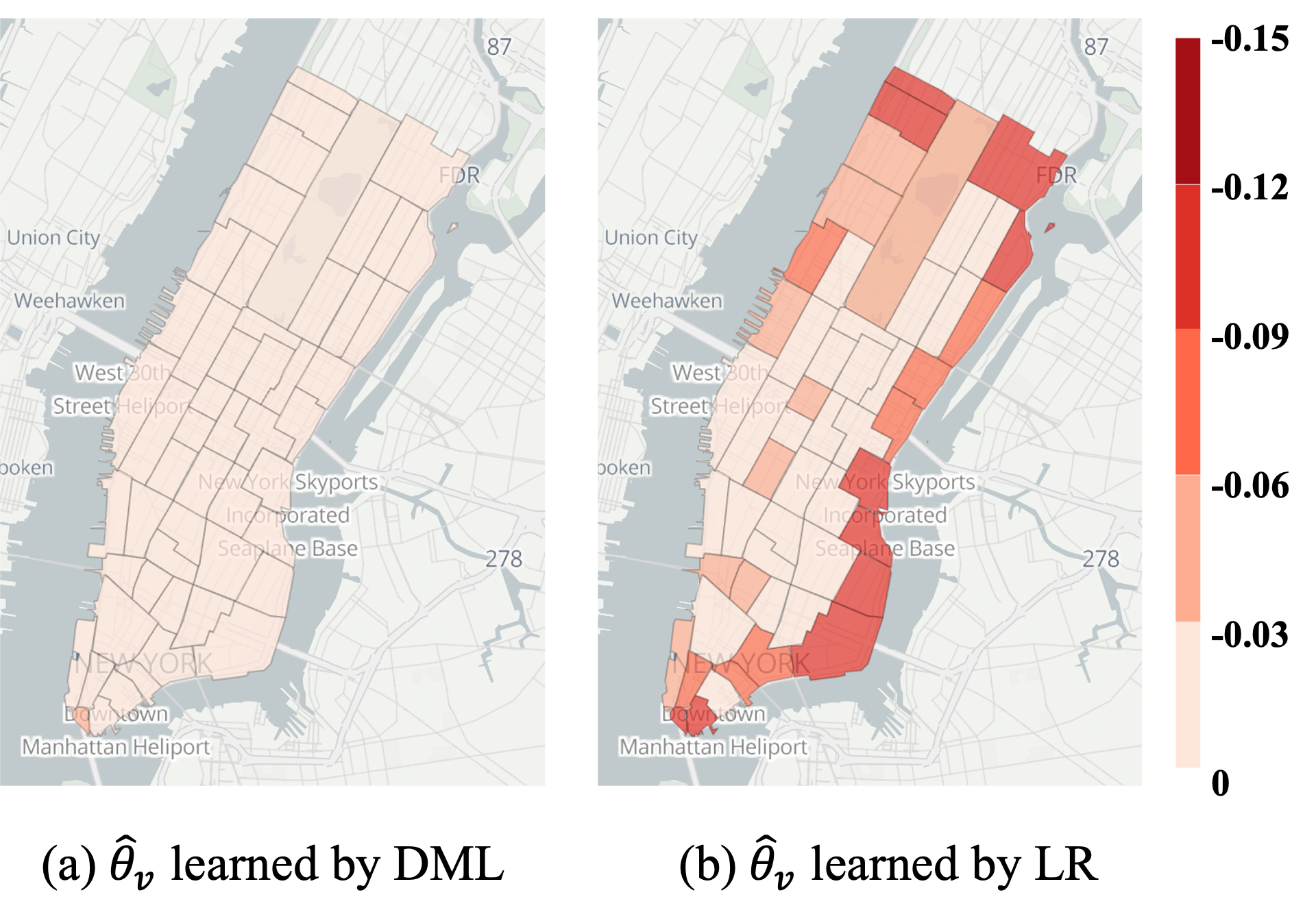}
    \caption{Comparison of estimated $\hat{\theta}_v$ by DML and LR on weekdays.}
    \label{fig: Estimation_comparison}
\end{figure}

\section{Congestion map and correlation analysis}
\label{sec:Con_Corr}

To reveal the congestion level of each region, we propose a traffic indicator, the speed index that calculates the ratio of the traffic speed and the free flow speed. The necessity for such a traffic indicator is because the same value of speed in different regions represents different congestion levels. The speed index can then reflect the relative congestion levels among different regions. The formula of the speed index is shown in Equation~\eqref{eq:speed_index}. The spatial distribution of the speed index of each region on weekdays and weekends is shown in Figure~\ref{fig:Traffic_index}.

\begin{equation}
\label{eq:speed_index}
{\text{speed index of a region}} = \frac{\text{relative speed in the region}} {\text{free flow speed in the region}}
\end{equation}

\begin{figure}[h]
    \centering
    \includegraphics[width=0.7\linewidth]{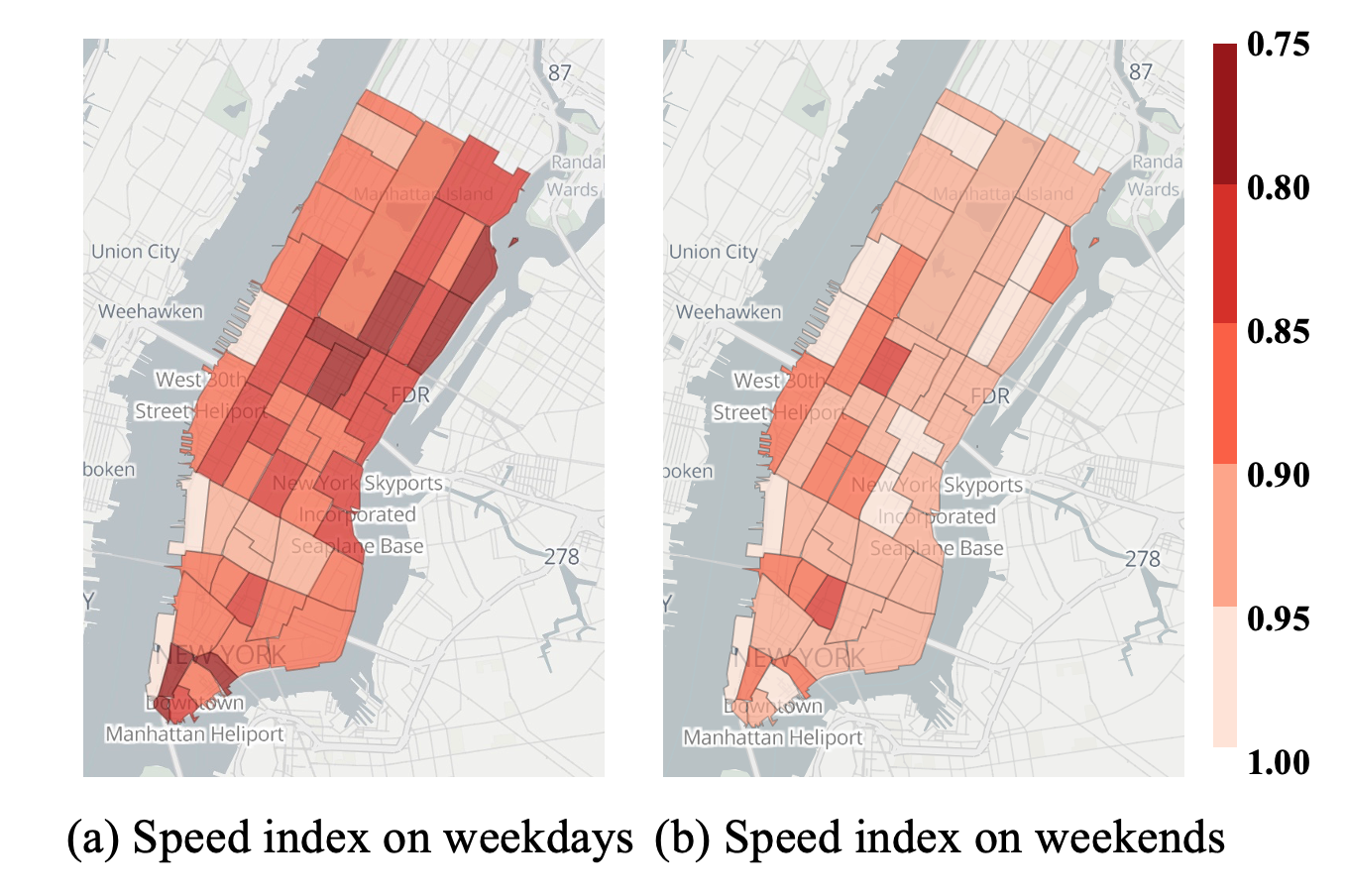}
    \caption{Congestion map in the Manhattan area.}
    \label{fig:Traffic_index}
\end{figure}

As shown in Figure~\ref{fig:Traffic_index}, the region marked with a redder color has a smaller value of the speed index, and it represents that the region is more congested. The congestion effect quantifies the negative effect of NoPUDO on traffic speed, and the speed index reflects the traffic situation of the transportation system. The correlation analysis between the speed index and the estimated congestion effect is shown in Table~\ref{tb:correlation_speed_index_congestion_effect}. The Pearson correlation coefficient between congestion effect and speed index on weekdays is $0.067$, and that value on weekends is $0.403$. This indicator measures the overall spatial correlation between the congestion effect and the speed index. The Pearson correlation coefficient shows that the speed index has a higher correlation with the estimated congestion effect on weekends rather than on weekdays.

\begin{table}[h]
\caption{Correlation analysis between speed index and estimated congestion effect.}
\label{tb:correlation_speed_index_congestion_effect}

\begin{tabular}{c c c c }
\toprule
Speed index & Speed index on weekdays & Speed index on weekends \\
\hline
Correlation coefficient & 0.067  & 0.403 \\
\hline

\end{tabular}
\end{table}

\section{Sensitivity analysis of re-routing formulation}
\label{Sensitivity analysis of re-routing formulation}

To evaluate the sensitivity with respect to demand level, we first perturb $\lambda$ to be 5, 10, 20, 25, and 30 and evaluate the according improvement rate. Note that $\lambda$ indicates the level of total traffic demands, and higher $\lambda$ represents more traffic demand. The mean and standard deviation of the improvement rates on different $\lambda$ for the Midtown and Central Park are shown in Figure~\ref{fig:Optimization_MidTown_weekdays_amount} and Figure~\ref{fig:Optimization_Central_Park_weekdays_amount}, respectively.

\begin{figure}[H]
    \centering
    \includegraphics[width=0.8\textwidth]{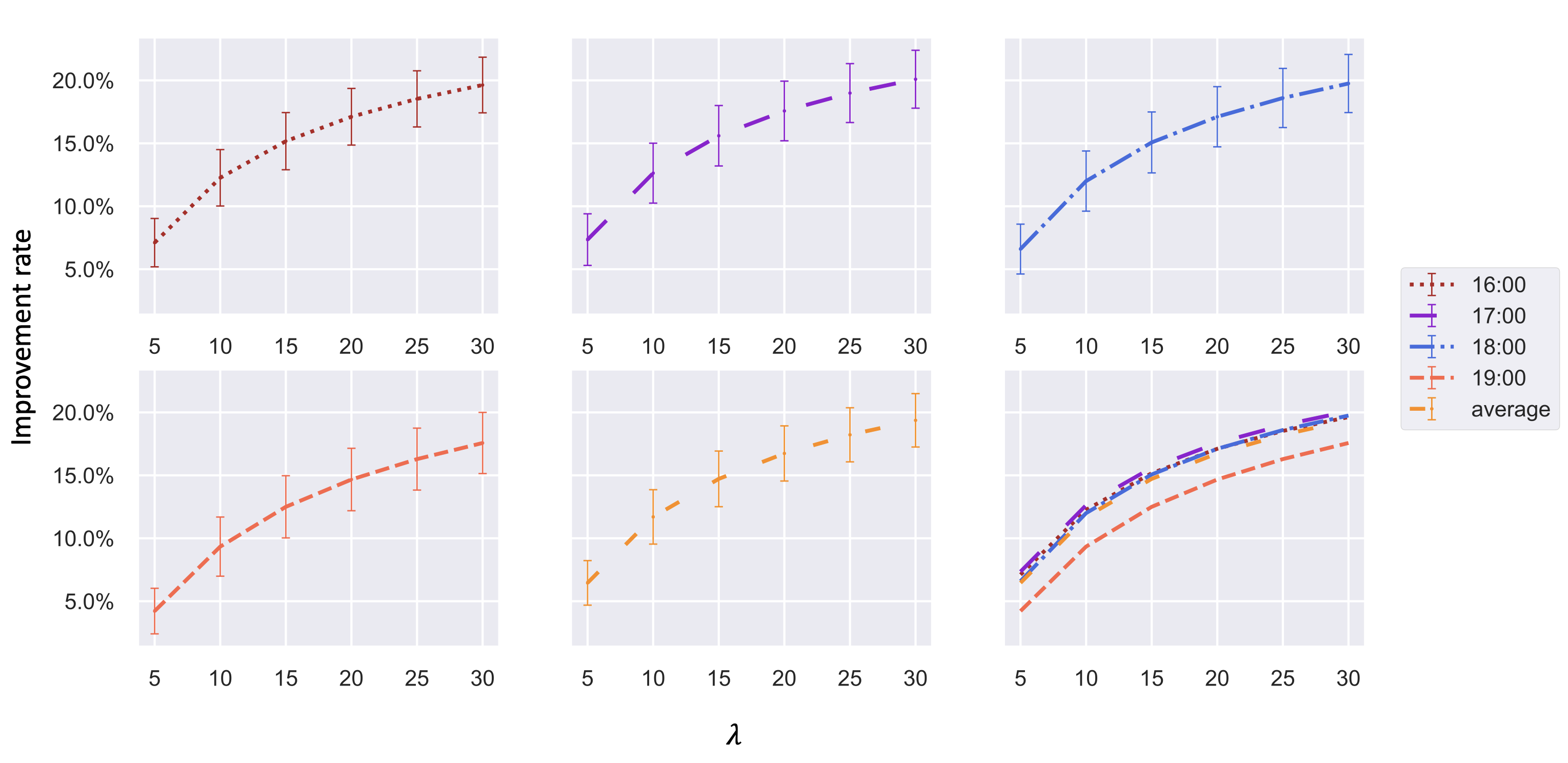}
    \caption{Improvement rates on different $\lambda$ in Midtown (error bar represents the standard deviation).}
    \label{fig:Optimization_MidTown_weekdays_amount}
\end{figure}

\begin{figure}[H]
    \centering
    \includegraphics[width=0.8\textwidth]{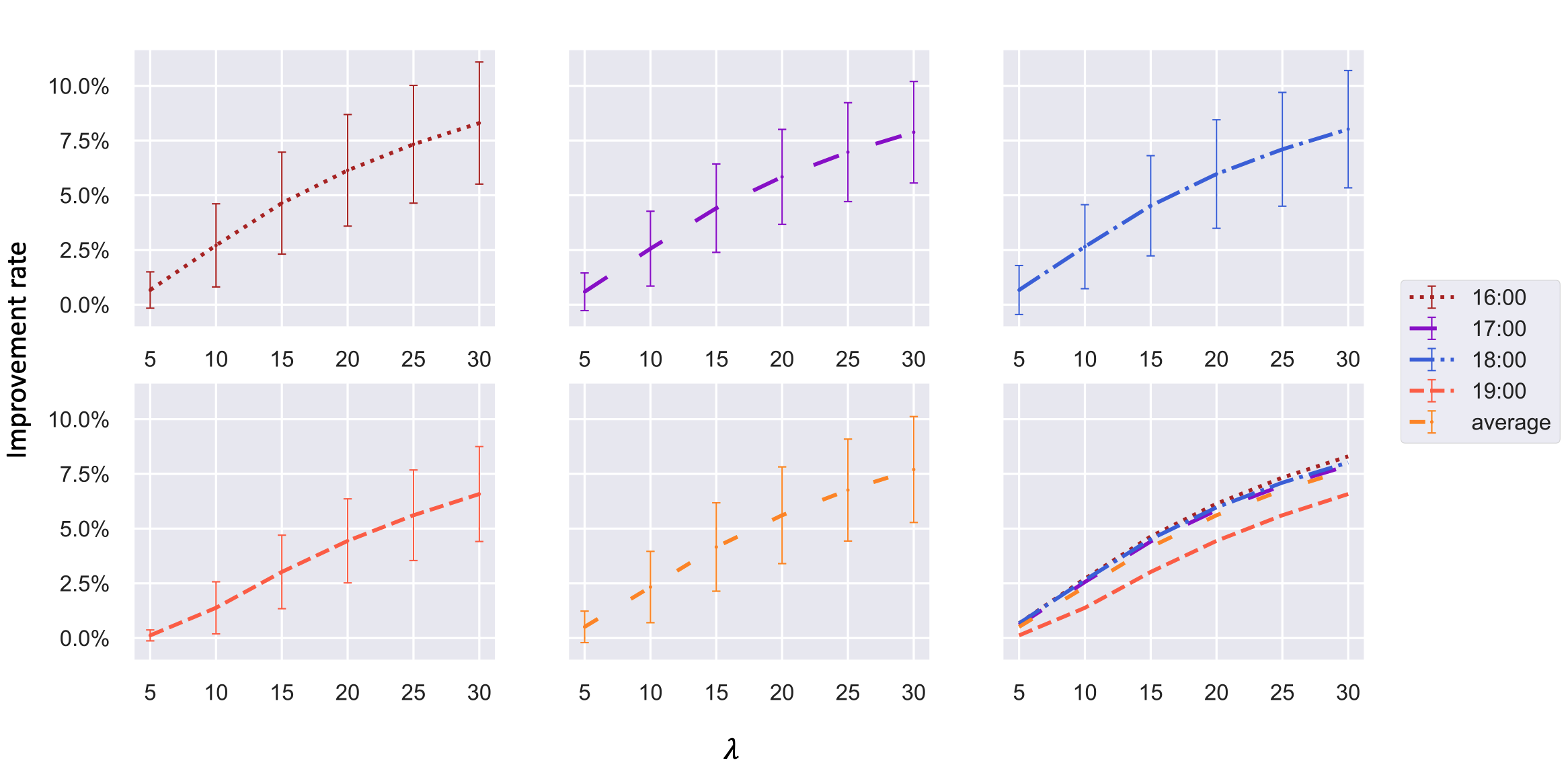}
    \caption{Improvement rates on different $\lambda$ in Central Park (error bar represents the standard deviation).}
    \label{fig:Optimization_Central_Park_weekdays_amount}

\end{figure}

In general, higher traffic demands encourage a larger improvement rate for both areas. Re-routing traffic flow with PUDOs turns out to be a promising and robust tool for system optimal under different demands levels. Additionally, an interesting finding is that the standard deviation of the improvement rate is also increasing. This suggests that when the demand increases, network conditions become more random, and the TTT improvement becomes more stochastic. 

Secondly, we vary $\gamma$ from 2.1 to 2.5 for Midtown, and from 1.4 to 1.8 for Central Park, to examine the sensitivity regarding the limitation of NoPUDO changes. The resulted improvement rate curves are shown in Figure~\ref{fig:Optimization_MidTown_weekdays_maximum} and Figure~\ref{fig:Optimization_Central_Park_weekdays_maximum}.

\begin{figure}[H]
    \centering
    \includegraphics[width=0.8\textwidth]{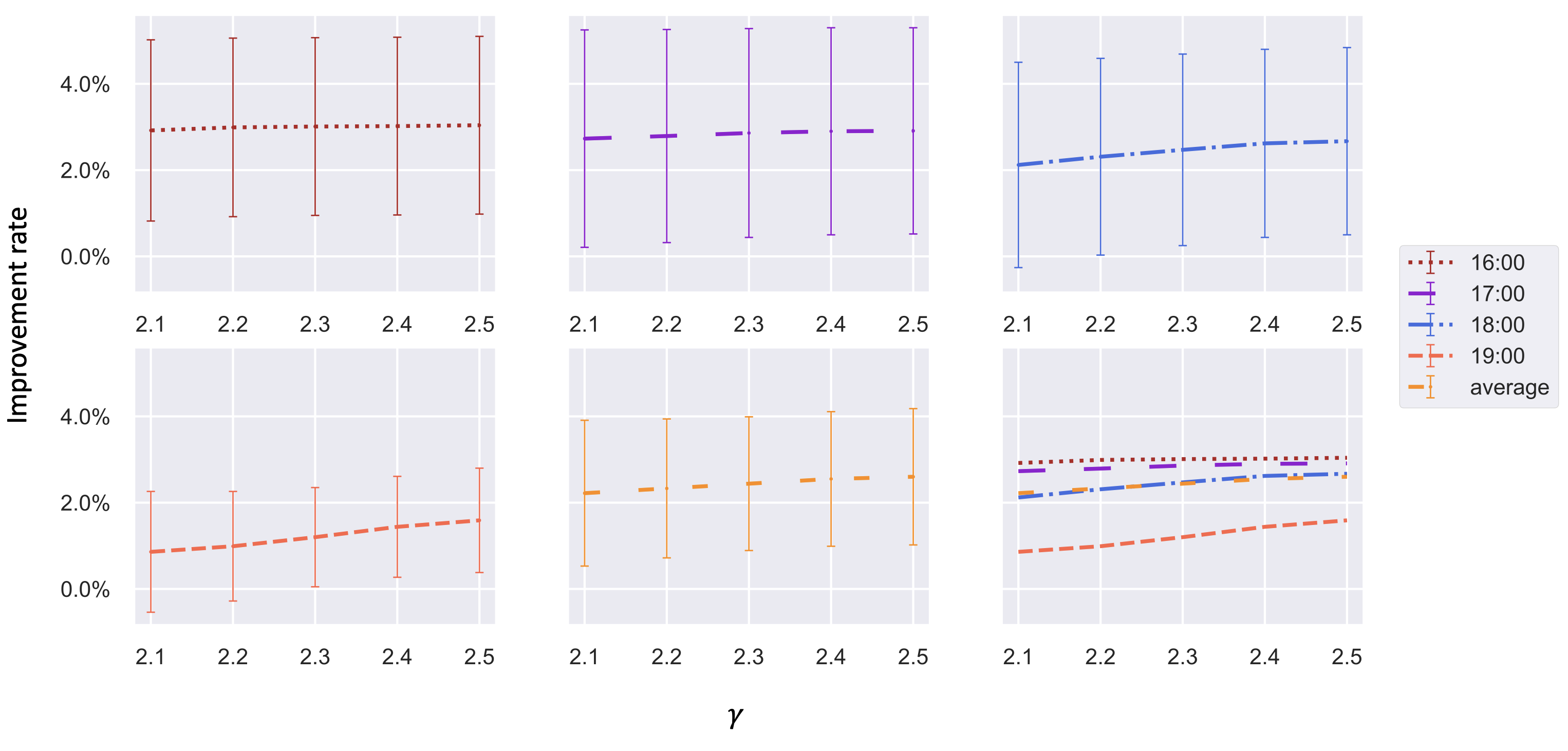}
    \caption{Improvement rates on different $\gamma$ in Midtown (error bar represents the standard deviation).}
    \label{fig:Optimization_MidTown_weekdays_maximum}
\end{figure}

\begin{figure}[H]
    \centering
    \includegraphics[width=0.8\textwidth]{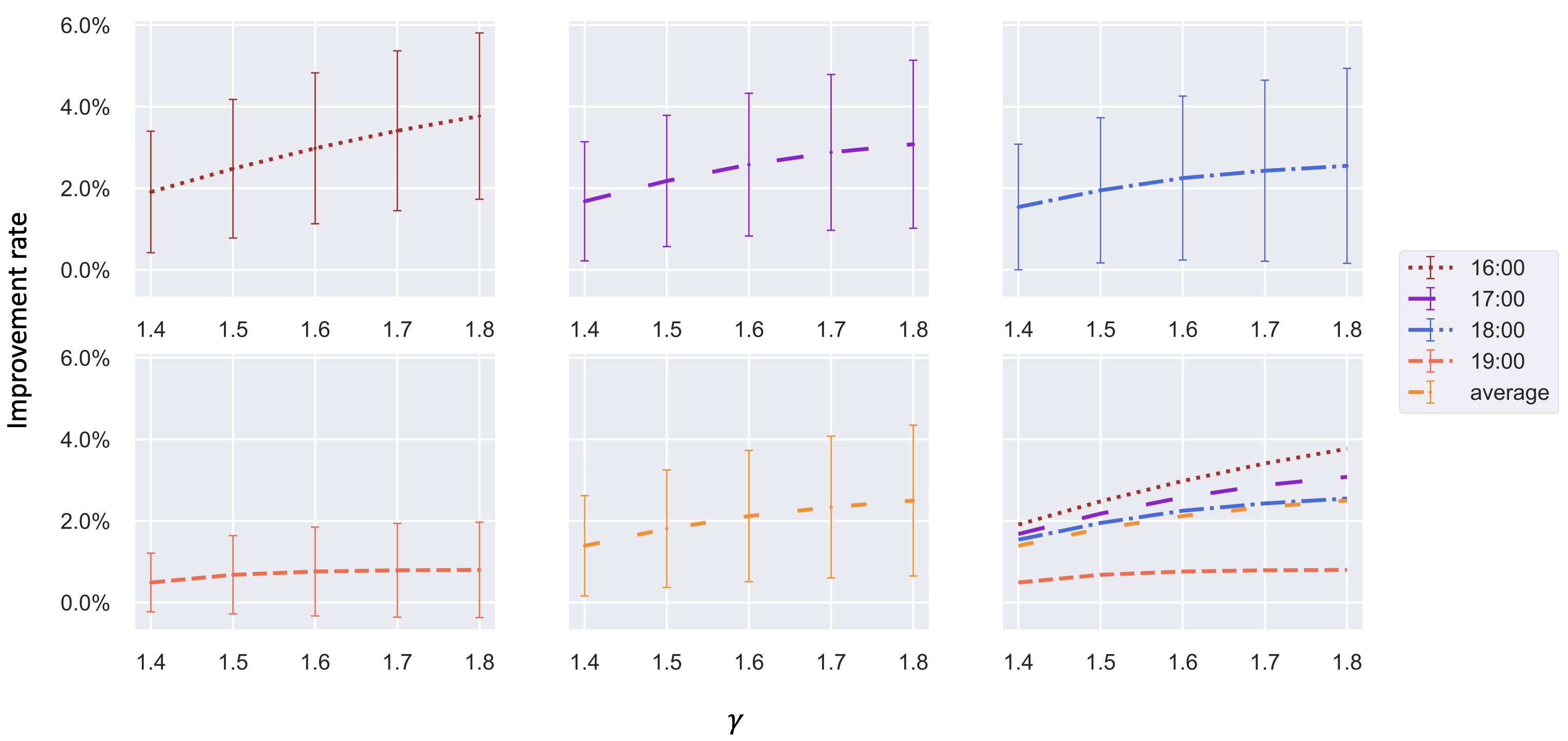}
    \caption{Improvement rates on different $\gamma$ in Central Park (error bar represents the standard deviation).}
    \label{fig:Optimization_Central_Park_weekdays_maximum}
\end{figure}

The improvement rate increases when $\gamma$ increases, and the reason is straightforward: increasing $\gamma$ will relax the limitation on the changes of NoPUDO in each region, and hence the search space for the re-routing formulation becomes larger. Another noteworthy point is that the standard deviation of the improvement rates remains the same when $\gamma$ changes in Midtown, while the standard deviation increases with respect to $\gamma$ in Central Park. This might be because of the unique characteristics and demand levels in each region.

\end{APPENDICES}

\clearpage




\end{document}